\documentclass{article}

\newif\iftr
\newif\ifconf
\newif\ifbd
\newif\ifplan

\trtrue
\conffalse
\plantrue
\bdfalse

\PassOptionsToPackage{numbers, sort&compress}{natbib}
\ifconf
\usepackage{neurips_2025}
\fi

\iftr
\usepackage[preprint]{neurips_2025}
\fi

\usepackage[utf8]{inputenc} % allow utf-8 input
\usepackage[T1]{fontenc}    % use 8-bit T1 fonts
\usepackage[hidelinks]{hyperref}       % hyperlinks
\usepackage{url}            % simple URL typesetting
\usepackage{booktabs}       % professional-quality tables
\usepackage{amsfonts}       % blackboard math symbols
\usepackage{nicefrac}       % compact symbols for 1/2, etc.
\usepackage{microtype}      % microtypography
\usepackage{xcolor}         % colors
\usepackage[most]{tcolorbox} % prompt box

\newif\ifcnf   % Submission to a conf or journal, with space contraints
\newif\ifsq     % Squeeze space?
\newif\ifsqCAP
\newif\ifsqVS
\newif\ifsqEN
\newif\ifsqTIT

\newcommand{\ignore}[1]{}

\sqtrue
\sqCAPfalse
\sqENtrue
\sqVStrue
\iftr
\sqfalse
\fi
\usepackage{etex}
\usepackage{balance}
\usepackage{epstopdf}
\usepackage{placeins}

\usepackage{graphicx}
\usepackage{float}
\usepackage{multirow}
\usepackage{rotating}
\usepackage{makecell}
\usepackage{tabulary}
\usepackage{parcolumns}
\usepackage{tikz}
\usetikzlibrary{tikzmark}

\usepackage{xpatch}
\expandafter\xpatchcmd
\csname pgfk@/tikz/every picture/.@cmd\endcsname
{\thepage}{\arabic{page}}{}{}

\tikzstyle{comment} = [draw, fill=blue!70, text=white, text width=3cm, minimum height=1cm, rounded corners, align=left, font=\scriptsize]
\tikzstyle{background_alg} = [draw, fill=blue!20, opacity=0.4, inner sep=4pt, rounded corners=2pt]

\usetikzlibrary{shapes}
\usetikzlibrary{plotmarks}
\usetikzlibrary{calc, fit}

\usepackage{enumitem}

\usepackage{amsthm}
\usepackage{amsmath,amssymb,amsfonts}
\usepackage{mathtools,mathrsfs}

\usepackage{soul}
\usepackage{fontawesome}
\usepackage{pifont}
\usepackage{textcomp}
\usepackage{booktabs}
\usepackage{url}
\usepackage{pbox}
\usepackage[normalem]{ulem}
\usepackage[10pt]{moresize}
\newcommand{\noAnswer}{\textcolor{black}{\faQuestionCircle}}

\ifsqCAP
\usepackage[font={normalfont, scriptsize}]{caption}
\usepackage[font={normalfont, scriptsize}]{subcaption}
\else
\fi

\newcommand{\vspaceSQ}[1]{\ifsqVS\vspace{#1}\fi}
\newcommand{\enlargeSQ}[1]{\ifsqEN\enlargethispage{\baselineskip}\fi}

\ifsqTIT
\usepackage[compact]{titlesec}
\titlespacing*{\section}{0pt}{3pt}{-1pt}
\titlespacing*{\subsection}{0pt}{0pt}{-3pt}
\titlespacing*{\subsubsection}{0pt}{2pt}{1pt}
\fi

\usepackage{xcolor}
\definecolor{darkgrey}{RGB}{70,70,70}
\definecolor{lightgrey}{RGB}{200,200,200}
\definecolor{lyellow}{RGB}{255,255,100}
\definecolor{llyellow}{RGB}{250,250,180}
\definecolor{lgreen}{RGB}{144,238,144}
\definecolor{raphael_comments}{RGB}{13, 145, 24}

\usepackage[customcolors]{hf-tikz}
\hfsetbordercolor{white}
\hfsetfillcolor{vlgray}

\definecolor{vlgray}{rgb}{0.77 0.77 0.77}
\definecolor{ablack}{rgb}{0.2 0.2 0.2}
\definecolor{vllgray}{rgb}{0.9 0.9 0.9}
\definecolor{bblue}{rgb}{0.7 0.7 0.99}

\usepackage{colortbl}

\usepackage{inconsolata}
\usepackage{listings}

\ifsq
\else
\fi

\newcommand{\maciej}[1]{\textcolor{blue}{[Maciej: #1]}}

\newcommand{\lorenzo}[1]{\textcolor{teal}{[Lorenzo: #1]}}

\definecolor{hlL}{rgb}{0.8 0.8 0.99}

\newcounter{highlight}

\newcounter{hlLR}

\newcounter{hlLIR}

\newcounter{hlLIIR}

\newcounter{Ahighlight}

\renewcommand{\epsilon}{\ensuremath\varepsilon}

\renewcommand{\phi}{\ensuremath{\varphi}}

\newcommand{\ione}{\raisebox{-0.125em}{\includegraphics[height=0.75em]{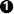}}}
\newcommand{\itwo}{\raisebox{-0.125em}{\includegraphics[height=0.75em]{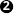}}}
\newcommand{\ithree}{\raisebox{-0.125em}{\includegraphics[height=0.75em]{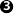}}}
\newcommand{\ifour}{\raisebox{-0.125em}{\includegraphics[height=0.75em]{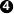}}}
\newcommand{\iquery}{\raisebox{-0.125em}{\includegraphics[height=0.75em]{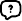}}}
\newcommand{\ireply}{\raisebox{-0.125em}{\includegraphics[height=0.75em]{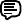}}}
\newcommand{\itruth}{\raisebox{-0.125em}{\includegraphics[height=0.75em]{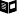}}}
\newcommand{\illm}{\raisebox{-0.125em}{\includegraphics[height=0.75em]{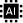}}}

\if 0

\usepackage[linesnumbered,ruled]{algorithm2e}
\usepackage{multicol}
\SetKwComment{Comm}{$\triangleright$\ }{}
\SetAlFnt{\scriptsize}
\SetAlCapFnt{\scriptsize}
\SetAlCapNameFnt{\scriptsize}
\SetKwInOut{Input}{Input}
\SetKwInOut{Output}{Output}

\makeatletter
\NewDocumentCommand{\LeftComment}{s m}{%
\Statex \IfBooleanF{#1}{\hspace*{\ALG@thistlm}}\(\triangleright\) #2}
\makeatother

\fi

\renewcommand{\maciej}[1]{}
\renewcommand{\lorenzo}[1]{}

\usepackage{fontawesome}
\usepackage{pifont}
\usepackage{textcomp}
\usepackage{xcolor}
\usepackage{soul}
\usepackage{colortbl}
\usepackage{subcaption}
\usepackage{graphicx}
\usepackage{tabularx}
\usepackage{wrapfig}

\usepackage{placeins} % use of FloatBarrier for the appendix

\newcommand{\faY}[0]{\faBatteryFull}
\newcommand{\faH}[0]{\faBatteryHalf}
\newcommand{\faN}[0]{\faTimes}
\newcommand{\faU}[0]{\noAnswer}

\newcommand{\name}[0]{\textsc{CheckEmbed}}
\newcommand{\nameS}[0]{\name\ }
\newcommand{\nameA}[0]{\textsc{CE}}
\newcommand{\nameAS}[0]{\nameA\ }

\ifconf
\usepackage[font={normalfont, scriptsize}]{caption}
\usepackage[font={normalfont, scriptsize}]{subcaption}
\fi

\title{\name: Effective Verification of LLM Solutions to Open-Ended Tasks}
\author{%
  Maciej Besta\thanks{corresponding author} \\
  ETH Zurich \\
  \And
  Lorenzo Paleari \\
  ETH Zurich \\
  \And
  Marcin Copik \\
  ETH Zurich \\
  \And
  Robert Gerstenberger \\
  ETH Zurich \\
  \And
  Ales Kubicek \\
  ETH Zurich \\
  \And
  Piotr Nyczyk \\
  Cledar \\
  \And
  Patrick Iff \\
  ETH Zurich \\
  \And
  Eric Schreiber \\
  ETH Zurich \\
  \And
  Tanja Srindran \\
  ETH Zurich \\
  \And
  Tomasz Lehmann \\
  Cledar \\
  Warsaw University of Technology \\
  \And
  Hubert Niewiadomski \\
  Cledar \\ IDEAS Research Institute \\
  \And
  Torsten Hoefler \\
  ETH Zurich \\
}

\begin{document}

\maketitle

\begin{abstract}
Large Language Models (LLMs) are transforming a wide range of domains, yet verifying their outputs remains a significant challenge, especially for complex open-ended tasks such as consolidation, summarization, and knowledge extraction. To address this, we introduce \nameS (\nameA): a simple, scalable, and accurate verification method. \nameAS reduces each LLM answer to a single embedding vector using powerful modern embedding LLM models like SFR-Embedding-Mistral. Prior methods such as BERTScore and SelfCheckGPT relied on weaker encoders like BERT, forcing them to operate at token or sentence granularity. In contrast, \nameAS performs fast, semantically rich comparisons directly at the whole-answer level, overcoming key limitations in both accuracy and scalability. We conduct a comprehensive design and time complexity analysis across 13 verification baselines, including classical text scorers (e.g., BLEU), stability-based methods (e.g., SelfCheckGPT), and generative evaluators (e.g., LLM-as-a-Judge), which highlights the effectiveness, efficiency, versatility, and simplicity of \nameA. Empirical results show that \nameAS reliably detects hallucinations in both closed and open-ended tasks. We further present evidence that \nameAS generalizes beyond text to other modalities such as vision, establishing it as a practical and versatile verification framework.
\end{abstract}

\iftr
\begin{center}
\textbf{Website \& code:} {\url{https://github.com/spcl/CheckEmbed}}
\end{center}
\fi

% keywords: Autonomous Agents, Large Language Models, LLMs

\section{Introduction}
\label{sec:intro}

Large Language Models (LLMs)~\cite{zhao2023survey, minaee2024large} have unlocked impressive capabilities across a wide range of domains. Yet, the reliability of LLM-generated content remains an open challenge, particularly the detection of hallucinations~\cite{petroni2019language, huang2023survey, zhang2023sirens} and the broader task of verifying correctness in free-form answers~\cite{chang2023survey, rawte2023survey, zhao2023survey, minaee2024large, wang2024usercentric, radharapu2023aart}. This problem becomes especially pronounced in complex open-ended tasks such as summarization, explanation, or definition extraction from legal documents, where outputs are long and highly unstructured. Here, hallucinations go beyond incorrect atomic facts: they may involve plausible-sounding but subtly wrong reasoning, invented definitions, or misleading paraphrases of the input. These hallucinations are hard to isolate and diagnose, as they may span sentences or entire output sections while maintaining fluency and coherence—posing challenges that exceed the capabilities of token- or sentence-level comparisons used in many modern schemes~\cite{manakul2023selfcheckgpt}.

A wide array of LLM verification techniques has been proposed, falling into three major categories. First, \textit{traditional text quality scorers} such as BLEU~\cite{10.3115/1073083.1073135}, ROUGE~\cite{lin2004rouge}, BLEURT~\cite{sellam2020bleurt}, BARTScore~\cite{yuan2021bartscore}, UniEval~\cite{zhong2022towards}, and COMET~\cite{rei2020comet} evaluate attributes like fluency, coherence, adequacy, and factuality, typically by comparing generated outputs to reference answers or inputs. Second, \textit{LLM-based evaluators (``LLM-as-a-Judge'')} such as G-Eval~\cite{liu2023geval} and others~\cite{gu2024survey, li2024generation, zheng2023judging} use autoregressive language models to assess generated content, either by framing verification as a question answering (QA) task or by generating scalar judgments on output quality. Third, recent \textit{stability-based methods}, notably SelfCheckGPT~\cite{manakul2023selfcheckgpt}, repeatedly sample the same query to assess the consistency of LLM outputs. If repeated outputs agree semantically, they are considered more likely to be correct. Each of these classes offers unique benefits but also suffers from limitations in complex open-ended tasks.

\ifconf
\begin{table}[t]
\vspaceSQ{-1em}
\centering
\footnotesize
%\small
\scriptsize
%\ssmall
\ifsq
\setlength{\tabcolsep}{1.5pt}
\renewcommand{\arraystretch}{0.6}
\else
\setlength{\tabcolsep}{3.25pt}
\fi
\caption{\textbf{\underline{The advantages of \nameAS over other verification schemes}}.
\textbf{No additional training (\underline{N-T?})}: Does the method require no training beyond the LLM's pretraining and fine-tuning?
%
%\textbf{Low inference cost (\underline{L-I?})}: Does the method avoid costly operations during inference?
%
\textbf{No ground truth needed (\underline{N-G?})}: Can the method operate without requiring any ground-truth labels or reference outputs?
\textbf{Supports open-ended task verification (\underline{S-O?})}: Is the method applicable to complex, unstructured tasks like summarization or term extraction?
\textbf{Supports fine-grained hallucination detection (\underline{S-H?})}: Can the method detect hallucinations at the level of individual facts?
\textbf{Model-agnostic (\underline{Me-A?})}: Can the method be applied to any LLM (e.g., black-box or white-box), without requiring model-specific instrumentation?
\textbf{Modality-agnostic (\underline{Ma-A?})}: Can the method be adapted to other modalities, such as Vision Models? % or Graph Foundation Models?
\textbf{Negligible runtime overhead (\underline{N-O?})}: Does the method incur minimal latency and resource consumption, making it suitable for high-throughput or real-time use?
\textbf{No preprocessing (\underline{N-P?})}: Can the method be applied directly to raw model outputs, without any intermediate structuring or cleanup?
\textbf{Interpretability of verdict (\underline{I-V?})}: Does the scheme produce interpretable outputs (e.g., visualizable heatmaps, confidence scores, rationales), or just scalar scores?
\textbf{Supports symmetric answer comparison (\underline{S-S?})}: Can the method symmetrically compare two answers (i.e., A vs B) for relative quality or semantic overlap?
\textbf{Robustness to surface form variation (\underline{R-S?})}: Can the scheme recognize semantic equivalence when surface forms differ (e.g., paraphrasing, reordering, synonym substitution)?
\textbf{Supports zero-shot settings (\underline{S-Z?})}: Can the method be used without task-specific fine-tuning or labeled data?
\textbf{No bias toward generator (\underline{N-B?})}: Does the method show a preference for outputs produced by the same model used for verification?
\textbf{Deployment simplicity (\underline{D-S?})}: How easily can the scheme be deployed in production (e.g., minimal dependencies, no pipeline engineering)?
``\faY'': full support, ``\faH'': partial support, ``\faN'': no support, ``\faU'': unknown. \maciej{Add Mind~\citep{su2024unsupervised}, HELM?}
}
\label{tab:qualitative}
\begin{tabular}{@{}lllllllllllllll@{}}
\toprule
\textbf{Scheme} & \textbf{N-T?} & \textbf{N-G?} & \textbf{S-O?} & \textbf{S-H?} & \textbf{Me-A?} & \textbf{Ma-A?} & \textbf{N-O?} & \textbf{N-P?} & \textbf{I-V?} & \textbf{S-S?} & \textbf{R-S?} & \textbf{S-Z?} & \textbf{N-B?} & \textbf{D-S?} \\
\midrule
\multicolumn{15}{c}{\textbf{Text--quality scorers}} \\
%
% \midrule
%
BLEU~\cite{10.3115/1073083.1073135} & \faY  & \faN & \faN & \faN & \faY & \faN & \faY & \faH & \faN & \faN & \faN & n/a & n/a & \faY\\
ROUGE~\cite{lin2004rouge} & \faY  & \faN & \faN & \faN & \faY & \faN & \faY & \faY & \faN & \faH & \faN & n/a & n/a & \faY \\
BARTScore~\cite{yuan2021bartscore} & \faH &  \faY & \faY & \faH & \faY & \faN & \faU & \faY & \faY & \faH & \faY & \faH & \faU & \faH \\
UniEval~\cite{zhong2022towards} & \faN &  \faY & \faY & \faU & \faY & \faN & \faU & \faY & \faY & \faU & \faH & \faY & \faU & \faH \\
BLEURT~\cite{sellam2020bleurt} & \faN &  \faN & \faH & \faU & \faY & \faN & \faU & \faY & \faN & \faN & \faY & \faH & \faU & \faN \\
PRISM~\cite{thompson2020automatic} & \faN & \faN & \faN & \faN & \faY & \faN & \faU & \faY & \faN & \faY & \faH & \faY & \faU & \faN \\
\midrule
\multicolumn{15}{c}{\textbf{Stability--based verification}} \\
%
%\midrule
%
SelfCheckGPT (BERT)~\cite{manakul2023selfcheckgpt} & \faY &  \faY & \faH & \faH & \faY & \faN & \faH & \faY & \faN & \faY & \faY & \faY & \faU & \faH \\
SelfCheckGPT (NLI)~\cite{manakul2023selfcheckgpt} & \faN & \faY & \faH & \faH & \faY & \faN & \faN & \faY & \faN & \faN & \faY & \faY & \faU & \faN \\
HaloCheck~\cite{elaraby2023halo} & \faY & \faY & \faH & \faY & \faY & \faN & \faY & \faY & \faN & \faY & \faY & \faY & \faY & \faH \\
\midrule
\multicolumn{15}{c}{\textbf{Embedding--based similarity}} \\
%
% \midrule
%
BERTScore~\cite{zhang2020bertscore} & \faY & \faN & \faH & \faN & \faY & \faH & \faY & \faY & \faY & \faY & \faY & \faY & \faU & \faH \\
SentenceBERT~\cite{reimers2019sentence} & \faN & \faN & \faU & \faU & \faN & \faH & \faY & \faY & \faN & \faY & \faY & \faH & \faU & \faN \\
\midrule
\multicolumn{15}{c}{\textbf{LLM-as-a-Judge}} \\
%
% \midrule
%
G-Eval~\cite{liu2023geval} & \faY & \faY & \faY & \faH & \faH & \faH & \faY & \faY & \faH & \faU & \faY & \faY & \faN & \faY \\
GPTScore~\cite{fu2024gptscore} & \faY & \faY & \faY & \faH & \faY & \faN & \faU & \faY & \faY & \faU & \faY & \faY & \faU & \faY \\
%
%\midrule
%
%\multicolumn{16}{c}{\textbf{Trained verifiers}} \\
%
% \midrule
%
%Reward models~\cite{??, ??, ??} \\
%GenRM~\cite{??} \\
%
% \midrule
%
% \multicolumn{16}{c}{\textbf{Tool-augmented verification}} \\
%
\midrule
\makecell[l]{\textbf{\nameAS [This work]}} & \faY & \faY & \faY & \faY & \faY & \faY & \faY & \faY & \faY & \faY & \faY & \faY & \faY & \faY \\
\bottomrule
\end{tabular}
\vspaceSQ{-2em}
\end{table}

\fi

Many traditional scorers like BLEU and ROUGE are sensitive to surface-level lexical overlap and fail in cases of valid paraphrases~\cite{sai2022survey, sulem2018bleu}. BLEURT and COMET address some of these issues via pretrained encoders, but still require references and additional training. BARTScore and UniEval introduce learned evaluation via sequence-to-sequence models, necessitating training on task-specific data, and may still fail in tasks with diverse valid outputs. LLM-based judges like G-Eval offer greater flexibility but entail the cost and brittleness of prompting, may themselves hallucinate, and suffer from generator bias~\cite{liu2023geval}. SelfCheckGPT avoids references and generative evaluation, but it addresses closed questions only. Moreover, it fundamentally operates at token or sentence granularity, which scales poorly with the answer length.

To address the above issues, we introduce \nameS (\nameA), a design based on a very simple yet powerful idea: \textit{combine the stability-based verification with whole-answer embedding similarity} (\textbf{contribution~1}). Methods such as SelfCheckGPT or BARTScore intentionally utilize token-level and sentence-level information to overcome the limitations of whole-answer embedding similarity, but they have traditionally relied on embedding models of low capabilities. Here, \nameAS builds on a core observation: {modern} embedding models (e.g., \texttt{SFR-Embedding-Mistral}~\cite{SFRAIResearch2024} or \texttt{bge-large-en}~\cite{xiao2024cpack}) are capable of encoding not just lexical content but also rich, high-level meaning of full passages~\cite{lee2024nv}. Following this observation and the stability principle of SelfCheckGPT, we generate and embed multiple answers to a given query. We then assess alignment between multiple answers (or between an answer and a reference) using a simple metric such as cosine similarity. Thanks to harnessing modern embedding models, and unlike prior methods that operate at fine granularity, \nameAS can capture both high-level semantic consistency and subtle factual divergence, all while maintaining very high computational efficiency as it only computes one embedding per answer.

\iftr

\fi

We compare \nameAS quantitatively to 13 {schemes capable of LLM output verification} in Table~\ref{tab:qualitative} using 14 design dimensions explained in the table caption (\textbf{contribution~2}). \nameAS enhances the key strengths of each major method category while avoiding their downsides. Compared to traditional text scorers, it eliminates reference dependence, sensitivity to surface-level differences, and reliance on training. Compared to generative evaluators, it is orders of magnitude faster, cheaper, and more stable with respect to the prompt format. Compared to stability-based methods, it handles open-ended questions and it scales much better. On top of that, while \nameS primarily targets text, it can also be applied to other modalities, for example images (given appropriate embedding methods), which we also showcase in our empirical analysis.

We apply \nameS to complex, real-world document analysis tasks, such as extracting term-definition pairs and summarizing long documents (\textbf{contribution~3}). In all cases, \nameAS shows strong empirical advantages, supported by a time complexity analysis (\textbf{contribution~4}). It reliably produces high scores for semantically correct outputs, and low scores for hallucinated or unstable ones. Crucially, it does so without requiring references or expensive generative judges. We also demonstrate evidence that \nameAS can reliably verify tasks in other modalities, including vision (\textbf{contribution~5}). The result is a practical system that is accurate, fast, and applicable to domains beyond LLMs, offering reliable verification signals with minimal computational and implementation overhead.

%\vspaceSQ{-0.5em}
\section{The \nameS Design}
\label{sec:design}
%\vspaceSQ{-0.5em}

We now describe the \nameS design, which is summarized in Figure~\ref{fig:overview}. It is very simple, but as we show in the following evaluation sections, the simplicity is precisely what enables advantages in both effectiveness and efficiency.

First, a user sends a \textbf{question \iquery~to the LLM} \ione~with all the essential input data. 
\iftr
The pipeline enables batching these questions, i.e., it is possible to send multiple questions in the same pipeline and they pass through each of the next stages individually.
\fi
%
% The following two points are done in the same step, but the final Use Case is that the user has no ground truth and we are just checking the correctenss of the first reply (That apply also if we ask directly multiple sample, we want to verify correcteness of first one)
%
Next, the system \textbf{prompts the LLM \illm~several ($k$) times} \itwo~with the same question \iquery; $k$ is a user parameter. Each reply \ireply~has no prior knowledge of the previous answer guaranteeing that there is no bias. $k$ introduces a tradeoff: more responses (higher $k$) means more compute time and cost (more tokens used), but also a better check of correctness. However, as we show in Section~\ref{sec:eval}, \nameS enables high level of confidence in its verification outcome even when $k$ is low.
After that, each reply is embedded \ithree, using a pre-specified embedding model. The potential ground-truth answer \itruth~is also embedded.
The pairwise \textbf{similarity scores between embeddings} are computed \ifour~(e.g., with cosine similarity or Pearson correlation) and then summarized using a selected measure in order to provide a simple threshold number that can be used to drive decision making in practical deployments. The scores can also be grouped into a (symmetric) heatmap matrix, which facilitates explainability.

\iftr
\begin{figure}[ht]
  \centering
  \includegraphics[width=1.0\linewidth]{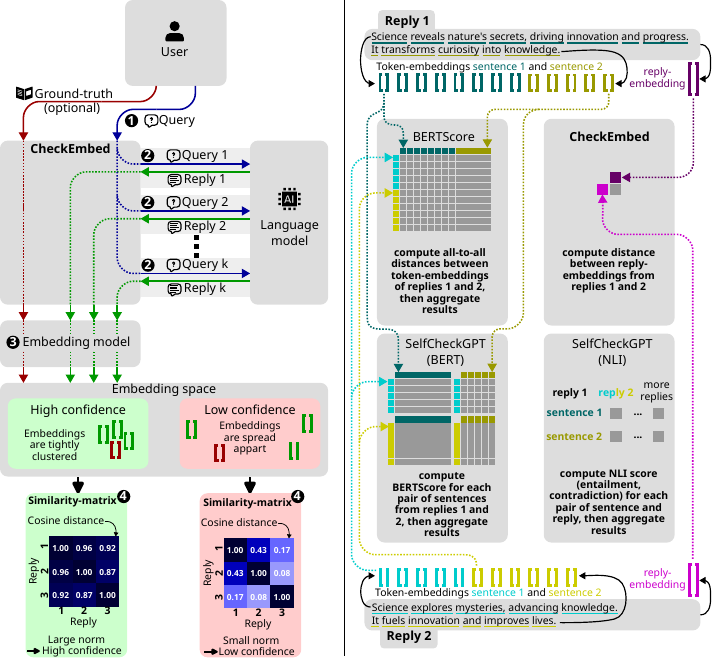}
  \caption{Overview of the \nameS pipeline (left) and comparison between BERTScore, SelfCheckGPT, and \nameS (right).}
  \label{fig:overview}
\end{figure}
\else
\begin{figure}[ht]
    \centering
        \includegraphics[width=1.0\linewidth]{plots/checkembed_overview.pdf}
        \vspaceSQ{-1.5em}
    \caption{Overview of the \nameS pipeline (left) and comparison between BERTScore, SelfCheckGPT, and \nameS (right).}
    \label{fig:overview}
\end{figure}
\fi

%\vspaceSQ{-0.75em}
\section{Complexity Analysis}
\label{sec:scalability}
%\vspaceSQ{-0.75em}

We provide a {brief scalability} analysis in Table~\ref{tab:complexity} showing why \nameAS is fundamentally more scalable and faster than other baselines under the main task of verifying an LLM answer or checking for a hallucination. We also consider the complexities of deriving a similarity score between two text passages -- a key building block in many schemes. We consider depth (the latency under unlimited parallel resources) and work (total operation count), two established measures for assessing the performance of parallel algorithms~\cite{fortune1978parallelism, keller2000practical, blelloch2010parallel}. The notation and assumptions are explained in the table caption. Derivation details for comparison baselines are in Appendix~\ref{sec:app:complexity}.

%- Differences:
% Keep in mind: All this numbers do not consider sampling time (same for every one of the approaches).

In {\nameA}, to verify an answer, one first generates $k$ answers and computes $k$ embeddings. This is $O(D_I + D_M)$ depth, as executing the language and embedding models can be parallelized. The work is $O(k(W_M + W_I))$ ($D$ and $W$ are the depth and work of a single model inference run; cf.~the caption of Table~\ref{tab:complexity} for full notation details). Then, we compare these embeddings pairwise ($k^2 / 2$ comparisons). This is $O(\log d)$ depth, as all comparisons can be parallelized, but the cosine similarity depth is $O(\log d)$ due to the reduction when summing $d$ terms. The work is equal to $O(k(W_M + W_I) + k^2 d)$ as we execute $k$ inference runs and then conduct $k^2 / 2$ cosine similarity score comparisons.

\begin{table}[t]
\centering
\footnotesize
%\small
\scriptsize
%\ssmall
%
\setlength{\tabcolsep}{1.5pt}
\ifsq
\renewcommand{\arraystretch}{0.8}
\fi
\caption{\textbf{\underline{Comparison of the complexities of different verification schemes}}.
We analyze the complexity of a full verification run (\emph{open-ended task answer}),
and additionally specify the included cost of comparing two multi-sentence passages (\emph{computing similarity}).
``Depth'' is the runtime complexity assuming unlimited parallel resources. 
``Work'' is the total number of all operations.
$k$: the number of answers requested from the LLM (in stability-based schemes).
$d$: embedding dimensionality.
$W_x, D_x$: work and depth complexity of a single model inference run, respectively,
for inference ($x=I$) and embedding ($x=M$). 
$s$: the (average) number of sentences in a single text passage.
$t$: the (average) number of tokens in a single sentence.
For any embedding based methods, we assume the same embedding dimensionality of all used embeddings and that computing a score of two embeddings takes $O(\log d)$ latency (e.g., Numpy supports highly efficient Pearson correlation and cosine similarity). When comparing the baselines, we consider counts of the two most computationally intensive operations within the pipeline: the number of embeddings to be constructed and the number of similarity operations to be conducted.
%
%\maciej{would it be more appropriate than $O(1)$ that we had before, Lorenzo? Or any convincing references why it's O(1)?} 
%\maciej{Add Mind~\citep{su2024unsupervised}, HELM to the table?}
%
}
\label{tab:complexity}
\begin{tabular}{@{}lcccc@{}}
\toprule
& \multicolumn{2}{c}{\textbf{Computing similarity of two passages}} & \multicolumn{2}{c}{\textbf{Verifying an open-ended task answer}} \\
\cmidrule{2-3} \cmidrule{4-5} 
\textbf{Scheme} & 
Latency \textbf{(depth)} & Total operation count \textbf{(work)} & Latency \textbf{(depth)} & Total operation count \textbf{(work)} \\
\midrule
\multicolumn{5}{c}{\textbf{Text--quality scorers}} \\
%
% \midrule
%
BARTScore~\cite{yuan2021bartscore} & $O(\log(st))$ & $O(t)$ & $D_I + O(\log(st))$ & $W_I + O(st)$ \\
UniEval~\cite{zhong2022towards} & n/a & n/a & $D_{I}$ & $W_{I}$ \\
%BLEU~\cite{10.3115/1073083.1073135} & \faN \\
%BLEURT~\cite{sellam2020bleurt} \\
%ROUGE~\cite{lin2004rouge} \\
%PRISM~\cite{thompson2020automatic} \\
%
\midrule
\multicolumn{5}{c}{\textbf{Stability--based verification}} \\
%
%\midrule
%
%SCGPT + BERTScore~\cite{manakul2023selfcheckgpt} & $D_M + O(\log d + \log t + \log k)$ & $2 k W_M + O(k t^2 \log d)$ & $D_M + O(\log d + \log t + \log k + \log s)$ & $2 k s W_M + O(k s t^2 \log d)$ \\
SCGPT (BERT)~\cite{manakul2023selfcheckgpt} & $D_M + O(\log (dts))$ & $s(2 W_M + O(t^2 \log d + t^2))$ & $D_M + O(\log (dtks))$ & $2 k s(W_M + O(t^2 \log d + t^2))$ \\
SCGPT (NLI)~\cite{manakul2023selfcheckgpt} & $D_I + O(\log s)$ & $sW_I + O(s)$ & $D_I + O(\log (ks))$ & $k s W_I + O(ks)$\\
HaloCheck~\cite{elaraby2023halo} & $D_I + O(\log s)$ & $s^2 W_I + O(s^2)$ & $D_{I} + O(\log(ks)))$ & $k^2(s^2 W_I + O(s^2) + 1)$\\
\midrule
\multicolumn{5}{c}{\textbf{Embedding--based similarity}} \\
%
% \midrule
%
BERTScore~\cite{zhang2020bertscore}  & $D_M + O(\log (dt))$ & $2W_M + O(t^2 \log d + t^2)$ & n/a & n/a \\
SentenceBERT~\cite{reimers2019sentence} & $D_I + O(\log (dt))$ & $2 W_I + O(t + d)$ & n/a & n/a \\
%Cosine similarity~\cite{??} & & & n/a & n/a \\
%
\midrule
\multicolumn{5}{c}{\textbf{LLM-as-a-Judge}} \\
%
% \midrule
%
G-Eval~\cite{liu2023geval} + GPT-4 & n/a & n/a & $D_{I} + O(\log k)$ & $kW_{I} + O(k)$\\
GPTScore~\cite{fu2024gptscore} & \faU & \faU & $O(D_{I})$ & $O(W_{I})$ \\
% LLM as discriminator~\cite{??} \\
%
%\midrule
%
%\multicolumn{5}{c}{\textbf{Trained verifiers}} \\
%
% \midrule
%
%Reward models~\cite{??, ??, ??} \\
%GenRM~\cite{??} \\
%
% \midrule
%
% \multicolumn{16}{c}{\textbf{Tool-augmented verification}} \\
%
\midrule
\makecell[l]{\textbf{\nameAS [This work]}} &  $D_M + O(\log d)$ & $k W_M + O(d)$ & $D_{I} + D_{M} + O(\log d)$ & $k(W_{M} + W_I) + O(k^2 d)$ \\
\bottomrule
\end{tabular}
\vspaceSQ{-1em}
\end{table}

\section{Evaluation}
\label{sec:eval}
%\vspaceSQ{-1em}

We now show the advantages of \nameAS over the state of the art. For full reproducibility, we provide detailed prompt specifications in Appendix~\ref{sec:app:prompts} and supplementary evaluation setup details in Appendix~\ref{sec:app:setup}. Due to a large amount of data and considered baselines, we present selected representative outcomes, full results are in Appendix~\ref{sec:app:eval}.

\maciej{The approach requires validation on standard text similarity or QA evaluation tasks to substantiate claims of superiority over more granular methods.} \maciej{Public dataset evaluations are minimal}

\maciej{"Despite claiming to verify open-ended tasks, the method effectively only works for hallucination detection tasks with definitive answers. Multiple divergent responses could all be valid for truly open-ended queries (e.g., "Tell me a joke"), which the method fails to accommodate."}

\maciej{"The legal document analysis experiments are limited (only two cases shown), lacking statistical significance tests to support performance claims. While the WikiBio experiments provide some representativeness, deeper analysis is needed to determine whether performance gains stem from the \nameS framework or the GTE model, as BERTScore and SelfCheckGPT could potentially be enhanced with GTE."}

\maciej{"Reliance on synthetic data: Section 4.1 uses synthetically generated data for the evaluation, which may limit the validity of broad claims drawn from this analysis. For instance, the specific generation prompts used could inherently favor the answer-level embedding technique. Given that this subtask is central to the \nameS approach, additional real-world evaluation would be valuable for supporting the conclusions."}

\maciej{"Lack of sampling details: The paper lacks detail on specific LLM sampling parameters (e.g., temperature, top\_k values), which could significantly impact the histogram-based approaches in \nameS. This information seems pretty important, as sampling + aggregation is one of the key contributions, and the variability of LLM responses (due to sampling parameters) could significantly affect the resultant stability/confidence results."}

\maciej{"Some of the interesting observations, such as in Section 4.2 (e.g., "Figure 4 shows that whenever \nameS has high confidence in the LLM replies, there is a high likelihood that these replies are close to the corresponding GT"), appear more like case studies than generalized findings. While it would be interesting if \name’s inter-sample similarity correlates well with correctness, the two legal examples provided aren't convincing. Additional evidence across the entire legal dataset, for instance, would strengthen this claim."}

\maciej{"Using histogram to do thresholding: Have you explored any methods for automatically deriving 'useful' confidence thresholds?"}

\maciej{"What sampling methods did you use/explore when generating LLM responses, and how might these impact your results? This is a pretty crucial detail which could effect your pipeline performance."}

\maciej{"Can you provide additional details on the proprietary and synthetic datasets used in evaluation? E.g. how many examples were these results computed over? Can you release the synthetic datasets produced?"}

\maciej{What other real-world LLM benchmarks can we use?}

\maciej{Typos / Presentation Suggestions
l248: noticely
ll356 – Consistently
Figure 5 and Figure 6 – can you add trendlines to these?
Figure 9/10/11 – can you clearly outline the different parameters at the beginning for better readability? (e.g. [GPT-4 sampling, Stella 1.5B embeddings])}

\maciej{"The evaluation appears to be designed specifically to highlight flaws in BERTScore. The test setting using similar-but-different text is not sufficiently rigorous. While Figure 1 shows data where most tokens and patterns are similar with only a few different tokens, BERTScore would calculate many redundant token pairs, potentially overlooking key differences. However, realistic scenarios involve much more complex data patterns than these contrived examples. A more robust evaluation using data with sufficient pattern complexity should be added to prove the method's validity."}

\maciej{""When calculating cosine similarity and Pearson correlation between k replies of varying lengths: How are the different lengths handled? For Pearson correlation specifically, is comparing tokens in their natural order meaningful when the data shows similar patterns with only a few token substitutions?""}

\maciej{"Some issues with the finer details of the paper. For instance, the two heatmaps in the main figure (Figure 2) have incorrect y-axis (or x-axis) labels."}

\maciej{"It seems that \nameS does not outperform baseline methods in certain experiments, such as those in Sections 4.2 and 4.3."}

\maciej{"the addition of formulas and examples would improve clarity. Specifically, the examples of similar/different replies and ground truth, the legal documents data point, and the detailed calculation process of Pearson correlation in Section 4.3 are needed."}

\maciej{"Intuitively, BertScore and SelfCheckGPT compare texts at a finer granularity (i.e., token- and sentence-level), while \nameS operates at a passage-level, which is coarser. Finer-grained comparisons usually yield better results, yet the experimental results show otherwise. How can this counterintuitive outcome be explained? (e.g., is it due to differences in embedding models, etc.?)"}

\textbf{Comparison Baselines. }
We compare \nameAS to {SelfCheckGPT with BERTScore}~\cite{manakul2023selfcheckgpt} (it represents {stability-based verifiers that operate at token- and sentence-level embeddings}), SelfCheckGPT with NLI~\cite{manakul2023selfcheckgpt} (it represents {stability-based verifiers that harness additional task-specific models}), HalluDetect~\cite{quevedo2024detecting} -- a scheme which leverages probability distributions from a separate fine-tuned LLM to identify inconsistencies in the generated text (it represents verifiers that incorporate additional task-specific training), BERTScore itself~\cite{zhang2020bertscore} (it represents straightforward application of token- and sentence-level embeddings to verification), and to an LLM-as-a-Judge~\cite{gu2024survey} (it represents delegating verification to modern powerful LLMs).

\textbf{Tasks. }
We consider {verification of answers to open-ended questions} (and implicitly hallucinations in such open-ended settings), {detection of fine-grained hallucinations}, and assessing similarity of text passages. While we focus on text, we also provide evidence that \nameAS can be applied to images.

\textbf{Considered Embedding Models. }
When embedding LLM replies with \nameA, we experiment with different embedding models, namely Salesforce/SFR-Embedding-Mistral (SFR)~\citep{SFRAIResearch2024}, intfloat/e5-mistral-7b-instruct (E5)~\citep{wang2022text,wang2024improving}, Alibaba-NLP/gte-Qwen1.5-7B-instruct (GTE)~\citep{li2023towards}, which all have around 7B parameters, as well as smaller models such as NovaSearch/stella\_en\_1.5B\_v5 (STE1.5, 1.5B parameters)~\citep{dunzhang2024stella1.5b, zhang2024jasper} and NovaSearch/stella\_en\_400M\_v5 (STE400, 400M parameters)~\citep{dunzhang2024stella400, zhang2024jasper}. We also use an API-based GPT Text Embedding Large (GPT) model~\citep{OpenAItextEmbeddingLarge}.
For BERTScore and SelfCheckGPT, we use the best possible models available for these baselines (i.e., microsoft/deberta-xlarge-mnli~\citep{he2021deberta} and roberta-large~\citep{liu2019roberta}). For vision, we use CLIP's Vision Transformer Large~\cite{icml/RadfordKHRGASAM21}.
We use the default embedding sizes (listed in the Appendix~\ref{sec:embed_length}).

\textbf{Considered Generative Models. }
For generating LLM answers and for the LLM-as-a-Judge baseline, we experiment with different LLMs, including GPT-4o, GPT-4o-mini, Llama-3.3-70B, and 
Llama-3.1-8B. In the initial phases of the project, we also explored GPT-4 and GPT-3.5; all insights are identical. For vision, we use the open-source Stable Diffusion 3.5 Medium model~\cite{pmlr-v235-esser24a}.

\textbf{Considered Similarity Measures. } When comparing embeddings (in \nameAS, SelfCheckGPT, and BERTScore), we use cosine similarity and the Pearson correlation score. These two follow the same accuracy patterns, so we focus on the cosine similarity. 
\iftr
We then use the Frobenius norm to extract a single value from the cosine similarity matrices as well as Spearman's rank correlation coefficient for summarization.
\fi

\textbf{Considered Datasets. }
We use the WikiBio dataset~\cite{lebret2016neural} as modified by Manakul et al.~\cite{manakul2023selfcheckgpt} for their evaluation of SelfCheckGPT. It consists of 238 documents based on Wikipedia articles, that were used to generate samples in which hallucinations were introduced. Each sentence of those samples was manually labeled as either ``major inaccurate'', ``minor inaccurate'', or ``accurate''.
We also consider RAGTruth~\cite{niu2024ragtruth} which contains 18,000 responses generated by different models in retrieval-augmented generation (RAG)~\citep{lewis2020retrieval, guu2020realm} tasks. The authors selected 450 responses per model for their test set, on which we base our evaluation.
\maciej{See if we manage HaluEval}
%
% Lorenzo:
% HaluEval is a large benchmark with 35,000 hallucinated samples, divided into
% tasks like QA, Dialogue, and Summarization (10k each), plus 5k for general
% tasks. Each sample was generated with GPT-3.5 using two methods, and the more
% hallucinated sample was selected. We generated additional samples for 2.5k
% datapoints in QA, Dialogue, and Summarization to run \nameAS and
% SelfCheckGPT. For HalluDetect training, we selected 1,000 samples from each
% task.
%
% \lorenzo{Unfortunately HaluEval data is an only hallucination dataset, it also present the same flaws of small hallucination detection, the hallucination are inserted using chatGPT and tends to repeat a lot in the samples, which causes our score to go up. This dataset make us look bad. It can maybe be fixed for the thesis and rebuttal, but now is not usable.}
% HaluEval~\citep{li2023halueval}, a large benchmark, consists of 35,000
% hallucinated samples, divided into different tasks (QA, dialogue, summarization
% and general tasks). It contains a mix of automatically generated
% samples as well as well as samples, which were annotated by humans.

\subsection{Analysis of Distinguishing Similar and Different Text Passages}
\label{sec:eval-distinguish}

\ifconf
\begin{wrapfigure}{r}{0.4\textwidth} % 'r' for right, 'l' for left
    \centering
    \vspaceSQ{-1em}
      \includegraphics[width=1.0\linewidth]{plots/plot_eval_violins_gpt_generic_gpt4o.pdf}
    \vspaceSQ{-1.0em}
    \caption{\textbf{Advantages of \nameAS in distinguishing similar and different LLM replies}. We vary the used embedding model (for \nameA) and the used generative model (for LLM-as-a-Judge).}
    \label{fig:eval-violins-gpt-generic-gpt4o}
    \vspaceSQ{-1.5em} 
\end{wrapfigure}
\else
\begin{figure}[ht]
    \centering
    \vspaceSQ{-1em}
    \includegraphics[width=0.8\linewidth]{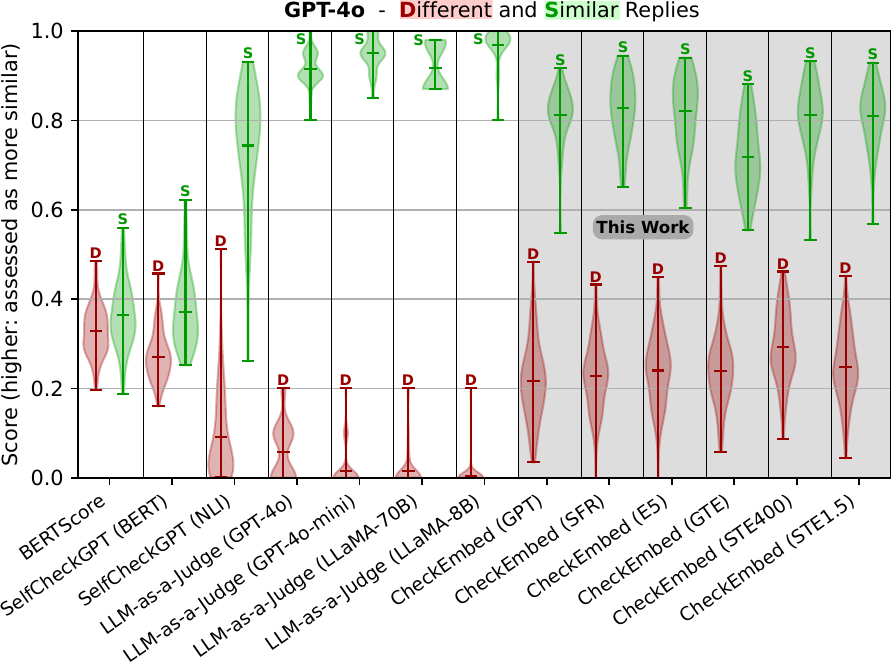}
    \caption{\textbf{Advantages of \nameAS in distinguishing similar and different LLM replies}. We vary the used embedding model (for \nameA) and the used generative model (for LLM-as-a-Judge).}
    \label{fig:eval-violins-gpt-generic-gpt4o}
\end{figure}
\fi

We start the evaluation by assessing an important building block of LLM verification stability-based methods (SelfCheckGPT, BERTScore, and \nameA) whether a given method is able to clearly distinguish two passages of text that (1) look similar, but come with very different meanings (``Different replies'', see the left side of Figure~\ref{fig:posterchild} for an example), as well as (2) look different, but have similar or identical meanings (``Similar replies'', see the right side of Figure~\ref{fig:posterchild} for an example). These methods rely fundamentally on comparing multiple LLM answers and, as such, recognizing differences between these answers is a crucial prerequisite for their effectiveness.
We illustrate the results for a large amount of such pairs in Figure~\ref{fig:eval-violins-gpt-generic-gpt4o}.

We observe that \nameAS appropriately assigns consistently high and low similarity scores to pairs of similar and to pairs of different replies, respectively (there is little to no overlap between these groups of data points), regardless of the embedding or generation model used. Contrarily, there is a large overlap between these groups of data points for both BERTScore and SelfCheckGPT, indicating that these baselines perform worse in distinguishing such replies effectively. SelfCheckGPT (NLI) shows a better (but still noticely inferior to \nameA) distinction between those two groups, but its runtime is significantly worse. 
We additionally plot the LLM-as-a-Judge results -- even though it does not rely on comparing text passages for verification, it serves as the top baseline in this task, considering its training objective, which enables (among others) distinguishing complex nuances in text.
The \textbf{key takeaway} is that \nameAS effectively recognizes the similarities and differences in the \textit{meaning} of the considered text passages, regardless of their length and style as well as used models, approaching the best performance of LLM-as-a-Judge -- all while being {highly scalable}.

\if 0
% Robert: was in the original arXiv version; I think it does not fit anymore
An interesting feature of \nameAS is that, while it \textit{does} distinguish similar and different passages very effectively, it gives \textit{relatively high} scores to the \textit{different} passages; these scores are usually \textit{higher} than the BERTScore or SelfCheckGPT scores for \textit{similar} passages. 
%
% This indicates that one has to compare these baselines somewhat carefully -- a relatively high score (e.g., 0.75) from \nameAS could already indicate different passages. 
%
Despite this, it is still straightforward to distinguish between answers implying similar or different passages, because the \nameAS scores for \textit{similar} passages are \textit{consistently} {very high} (e.g., with means higher than 0.9 for SFR or E5).
\fi

\begin{wraptable}{r}{0.40\linewidth}
    \centering
    \ifsq
    \renewcommand{\arraystretch}{0.6}
    \fi
    \vspaceSQ{-3em}
    \scriptsize
    \caption{Passage level correlation on WikiBio-gpt3 using Pearson (PE) and Spearman (SP).}
    \label{tab:wikibio_overview}
    \begin{tabular}{lcc}
        \toprule
        \textbf{Method} & \textbf{PE} & \textbf{SP} \\
        \midrule        
        BertScore & 67.7 & 67.9  \\
        SelfCheckGPT (BERTScore) & 57.4 & 54.6 \\
        SelfCheckGPT (NLI)       & \textbf{74.1} & 73.8 \\
        \midrule
        LLM-as-a-Judge (4o)	                  &  39.7 & 39.4 \\
        LLM-as-a-Judge (4o mini)	          &  27.1 & 31.4 \\
        LLM-as-a-Judge (Llama70b)	          &  -7.8 & -8.7 \\
        LLM-as-a-Judge (Llama8b)	          &   2.3 &  2.9 \\
        \midrule
        \nameS (GPT)    &  66.8 &  72.6 \\ 
        \nameS (STE400) &  68.5 &  72.9 \\    
        \nameS (STE1.5) &  69.9 &  73.8 \\
        \nameS (E5)     &  71.6 &  74.1 \\
        \nameS (SFR)    &  72.2 &  76.2 \\
        \nameS (GTE)    &  73.6 &  \textbf{76.2} \\
        \bottomrule   
    \end{tabular}
    \vspaceSQ{-3em}
\end{wraptable}
\subsection{Analysis of LLM Answer Verification}
\label{sec:eval-heatmaps}

We now illustrate how \nameAS enables effective verification of LLM answers.

\textbf{The WikiBio Dataset. }
First, we discuss the \nameAS performance on an existing benchmark, WikiBio, also used to assess SelfCheckGPT. Their subset consists of 238 documents based on Wikipedia articles with introduced hallucinations. Each sentence of those samples was manually labeled as either ``major inaccurate'', ``minor inaccurate'', or ``accurate''.
Consistent with the SelfCheckGPT evaluation by~\citet{manakul2023selfcheckgpt}, we employed a passage scoring system that aggregates sentence scores: assigning 0 for major inaccuracies, 0.5 for minor inaccuracies, and 1 for accurate sentences—before calculating the average score. This construction allows the utilization of Pearson and Spearman correlation scores to reflect a more nuanced output to quantify the extent of hallucination within passages over more simplistic black-and-white approaches.

An overview of the results is in Table~\ref{tab:wikibio_overview}. \nameAS demonstrates robust performance compared to existing baselines, particularly in Spearman's correlation, where its results are significantly higher. For Pearson's correlation, \nameAS is marginally outperformed by SelfCheckGPT's NLI variant, but it is more than 30$\times$ faster to compute. LLM-as-a-Judge performs significantly worse, as even powerful LLMs are inherently susceptible to hallucinations themselves.

\if 0 % some old WikiBio data...
\begin{wraptable}{r}{0.35\linewidth}
    \vspaceSQ{-1em}
    \scriptsize
    \caption{Passage level correlation on the WikiBio-gpt3 dataset using Pearson and Spearman.}
    \label{tab:wikibio_overview}
    \begin{tabular}{lcc}
        \toprule
        \textbf{Method} & \textbf{Pearson} & \textbf{Spearman} \\
        \midrule        
        BertScore & 67.7 & 67.9  \\
        \rowcolor[HTML]{EEEEEE} \multicolumn{3}{l}{SelfCheckGPT} \\
        w/ BERTScore & 57.4 & 54.6 \\
        w/ NLI       & \textbf{74.1} & 73.8 \\
        \hline 
        \rowcolor[HTML]{EEEEEE} \multicolumn{3}{l}{\textbf{\name}} \\
        w/ GPT    & 66.8 &  72.6 \\ 
        w/ STE400 &  68.5 &  72.9 \\    
        w/ STE1.5 &  69.9 &  73.8 \\
        w/ E5     &  71.6 &  74.1 \\
        w/ SFR    &  72.2 &  76.2 \\
        w/ GTE    &  73.6 &  \textbf{76.2} \\
        \bottomrule   
    \end{tabular}
    \vspaceSQ{-1em}
\end{wraptable}
\fi

% Robert: move here for better layout
\begin{figure}[t]
    \centering
    \includegraphics[width=0.95\textwidth]{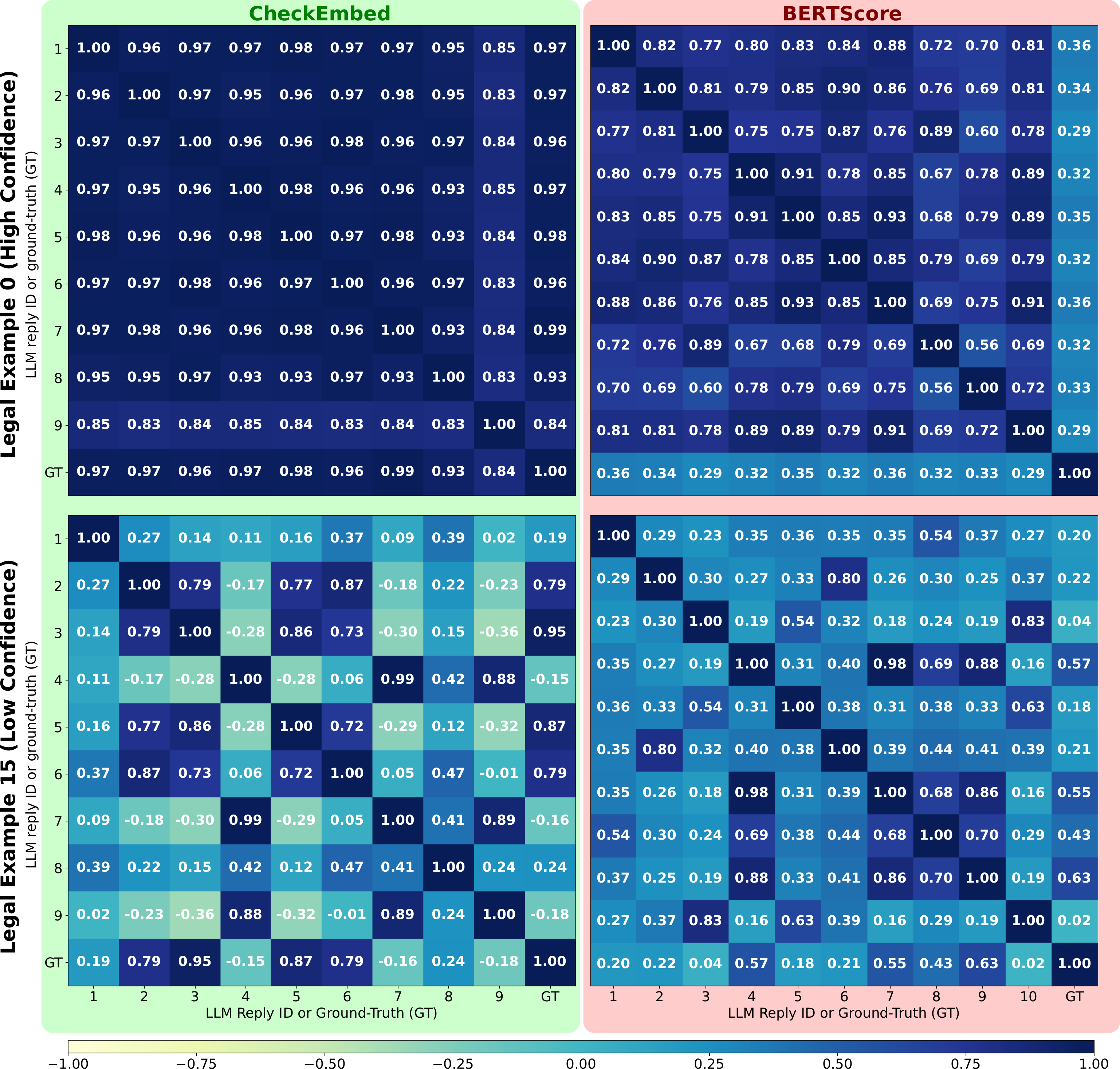}
    \caption{\textbf{Analysis of the verification of LLM answers.} We compare to BERTScore; SelfCheckGPT (with BERT) comes with significantly higher runtimes (detailed in Section~\ref{sec:runtimes}) and less competitive scores as it does not focus on open-ended answer-level analysis. The results form a heatmap of the \nameA's, or BERTScore's, cosine similarity between all LLM replies, and between each reply and the human expert prepared ground-truth (GT). Rows correspond to two representative legal documents, that come with -- respectively -- high and low LLM confidence in its replies. Embedding model used: GPT Text Embedding Large. Generative model used: GPT-4o.}
    \label{fig:heatmap_gpt4o}
    \vspace{-2em}
\end{figure}

\textbf{Real-World Legal Use Case (Verifying Open-Ended Tasks). }
Next, we extract terms and their definitions from legal documents; the used data is real and it comes from an in-house legal analytics project.
In this use case, a prompt to the LLM consists of the contents of a legal document (e.g., an NDA), as well as a request to extract respective terms and their definitions. The prompt sizes used in this task are in the range of 25--600 tokens (we split the documents into chunks as whole documents are often very long and come with total token counts that significantly exceed the recommended maximal sizes for the input of the used embedding models). \nameAS asks the LLM to generate 10 replies ($k=10$).
We use this use case to showcase the interpretability of \nameAS outcomes by plotting full heatmaps; illustrate example results for \nameAS vs.~BERTScore are in Figure~\ref{fig:heatmap_gpt4o} (other baselines and models are, as usual, in Appendix~\ref{sec:app:eval}).
Each figure shows the cosine similarity between LLM replies, and also between each reply and the ground-truth (GT) that has been prepared by a human expert.

The results illustrate that whenever \nameAS has very high confidence in its answer (top row in Figure~\ref{fig:heatmap_gpt4o}), which is visible by consistently having very high similarities between different replies, it corresponds to very high similarity scores between the LLM replies and the ground-truth. This is the case for all the considered models. Other baselines show mixed results for individual replies, and low similarities between their replies and GT. 
%
% This outcome directly corresponds to the results from Section~\ref{sec:eval-distinguish} and Figures~\ref{fig:eval-violins-gpt-generic} and~\ref{fig:eval-violins-gpt-precise} -- high scores form a cluster that has little overlap with low scores. \maciej{TODO: check with Lorenzo}
%
It shows that, whenever \nameAS has high confidence it the LLM replies, there is high likelihood that these replies are close to the corresponding GT.

\maciej{"The simple embedding-based approach may overlook complex logical relationships and reasoning errors, particularly in specialized domains (legal, medical), and fails to consider contextual coherence or detect plausible but contradictory content."}

In the bottom row of the figure, we provide example results where \nameAS indicates low or mixed LLM's confidence. While many scores are still high (e.g., 0.97), many are much lower, even negative. We manually verified that these particularly low individual scores correspond to LLM replies of very low quality (e.g., only a single term with its definition has been extracted). The low scores overall indicate model's low confidence, which is further supported by corresponding low similarity scores to GT. Here, BERTScore also has low confidence -- overall, its scores have a smaller ranger than those of \nameA, but its relative drop in similarity to GT is similarly as low as that of \nameA.

\iftr
Note that the results in the heatmaps directly correspond to the results from Section~\ref{sec:eval-distinguish} and Figure~\ref{fig:eval-violins-gpt-generic-gpt4o} in that very high \nameAS scores (e.g., 0.9) indicate high confidence while scores that are lower
% (e.g., 0.75) -- but still higher than BERTScore --
consistently mean low LLM's confidence.
% This indicates that whenever using such baselines together, one  may want to consider rescaling the results accordingly.
\fi

A useful simple \nameAS measure that indicates the low quality of the LLM answer is a selected summarization measure for a heatmap, for example mean or a matrix norm combined with a standard deviation (std). Whenever the mean is \textit{very high} (e.g., $>$0.9) and the std is \textit{low} (e.g., $<$0.05), the answer is of high quality with very high likelihood. Otherwise, one may want to investigate a given situation in more detail.
For example, in the top row (example~0), the LLM is very certain of what the answer is; the mean is 0.95 with very low std of 0.06; BERTScore seems to imply hallucinations with lower scores and even more importantly, an std of 0.18.

\if 0

In the bottom row (Example~11), we lower the score under 0.8, which start being a not good score and moreover we can easily tell the LLM is starting hallucinating looking at the std dev of 0.19. Value that is very high for us.
BertScore lower the score much, but has a std dev that is lower than the one from prompt0, it seems like the llm was hallucinating more looking at samples 0 than on samples 11. But it is really the case?
Looking manually answer from prompt11 differ much more than answer from prompt 0, that just have in some case a sentence more or less one in respect to each other or a rephrasing of the same concept

gpt4
Similar, this time BErtScore has been much more accurate.
Prompt 0 with std dev of 0.01 we are completely not hallucinating, same for Bert

prompt 8, we start seeing some hallucination, stronger in  aa particular answer, so overall score isn't too much bad, but std dev is already up 0.17 that is a warning factor 

Prompt4
hallucination, the answers are really diferent from each other, some of them are similar, that's true, but the overall score and std dev indicates high level of uncertainity (it looks correct this analysis looking manually)

\fi

\textbf{Real-World Legal Use Case (Detecting Fine-Grained Hallucinations). }
While \nameAS is primarily targeted at verification of open-ended tasks, we also investigate whether \nameAS can be used to detect small fine-grained hallucinations, such as mistakes in individual facts.
The results are in Figure~\ref{fig:hallucinate_gpt4-legal}. The task analyzed is summarizing legal articles. For each article considered, we generate a summary with no errors (labeled as ``ground truth''), and we also ask the LLM to summarize these documents, while forcing deliberate small fact-level mistakes, from 1 to 10 mistakes per summary.
\nameS is able to recognize when samples contains no errors, as illustrated by very large scores for GT. We can observe that the amount of low-confidence scores is somewhat increasing with the growing number of introduced errors. However, this increase only starts to be distinctive beyond 5 errors. The trends for BERTScore and SelfCheckGPT are similar, which shows that these baselines perform well for their intended use case. LLM-as-a-Judge comes with much higher spread, confirming its uncertainty for hallucination detection already observed in the WikiBio dataset analysis. Interestingly for other datasets, such as scientific descriptions illustrated in Figure~\ref{fig:hallucinate_gpt4}, \nameAS can recognize hallucinations after introducing a single error, as visible by no overlap between the GT and the consecutive data points.

\iftr
\begin{figure}[t]
\else
\begin{figure}[h]
\fi
    \centering
    \includegraphics[width=\textwidth]{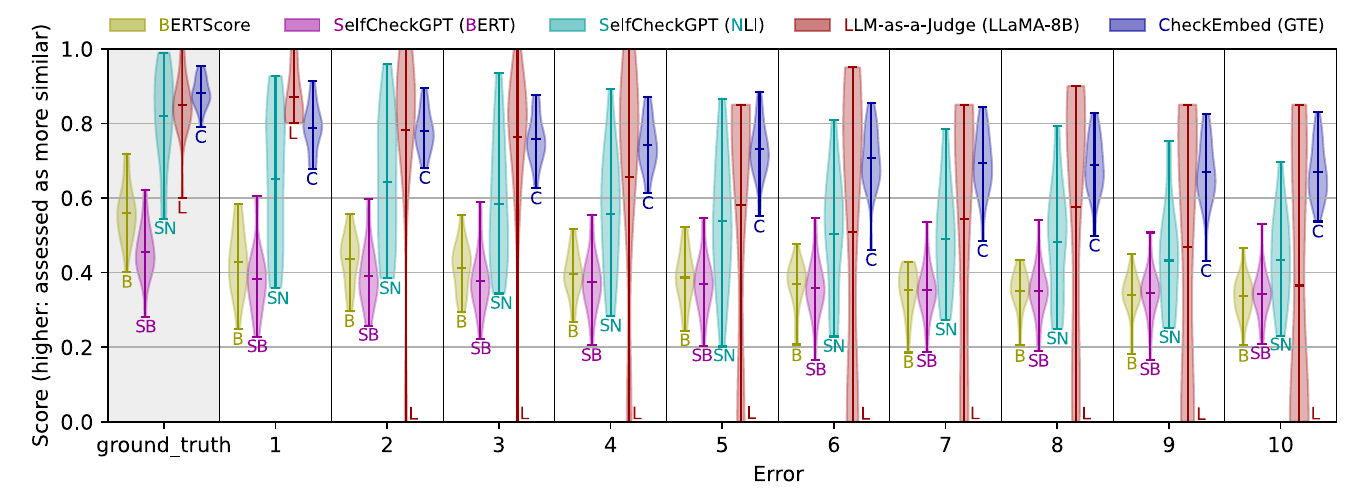}
    \caption{\textbf{Analysis of fine-grained hallucination verification of LLM answers (GPT-4o) when summarizing legal documents}.}
    \label{fig:hallucinate_gpt4-legal}
    \vspaceSQ{-1em}
\end{figure}

\subsection{Beyond LLMs To Other Modalities}
\label{sec:beyond_LLMs}
% - verification of outcomes in vision models and in graph foundation models (GFMs) is still very preliminary
% - \nameAS can also be extended here. While the full detailed multi-faceted design and analysis is outside the scope, we provide clear preliminary data showing this.

% \maciej{Eric, could you throw in the data here, and some text? I'd start as I do above (let's use/extend my story idea and add some refs). Maybe you could also grab Tanja's results and describe them as well. Plot-wise, some small summaries - full data (and these cool pictures) can go to the appendix.}

% \maciej{Eric/Tanja, we may want to compare to existing methods as well, at least maybe adding these methods to the tables 1-2... if you think it makes sense.}

\nameS can be applied straightforwardly across modalities, assuming the availability of a suitable embedding model. Preliminary results demonstrating this based on generative vision models are shown in Figure~\ref{fig:hallucinate_images}. While a full exploration of the design space and all possible modalities is beyond the scope of this work, we provide early evidence of CE's generality. 

\iftr
\begin{wrapfigure}{r}{0.49\linewidth}
    \vspaceSQ{-1em}
    \centering
    \includegraphics[width=0.95\linewidth]{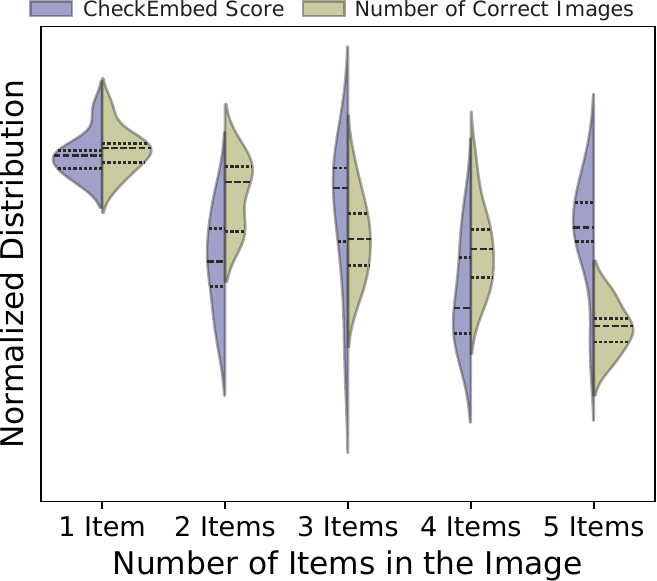}
    \caption{\textbf{Hallucination detection in vision models with \nameA}. We score response quality of a vision model without real-world references or auxiliary models. Both \nameS scores and correct image counts are normalized. As more items are requested, hallucinations rise, reducing correctness and \nameS scores.}
    \label{fig:hallucinate_images}
    \vspaceSQ{-1em}
\end{wrapfigure}
\fi

Verification in vision models remains at an early stage, typically relying on auxiliary models for evaluation~\cite{inceptionscore,lim2025evaluatingimagehallucinationtexttoimage} or on similar real world data for statistical comparison~\cite{Frechet_Inception_Distance,Tiv_Hallucination_MICCAI2024}. In contrast, \nameS operates without assuming access to a reference model or real world data, and still achieves strong performance.

\ifconf
\begin{wrapfigure}{r}{0.49\linewidth}
    \vspaceSQ{-1em}
    \centering
    \includegraphics[width=0.95\linewidth]{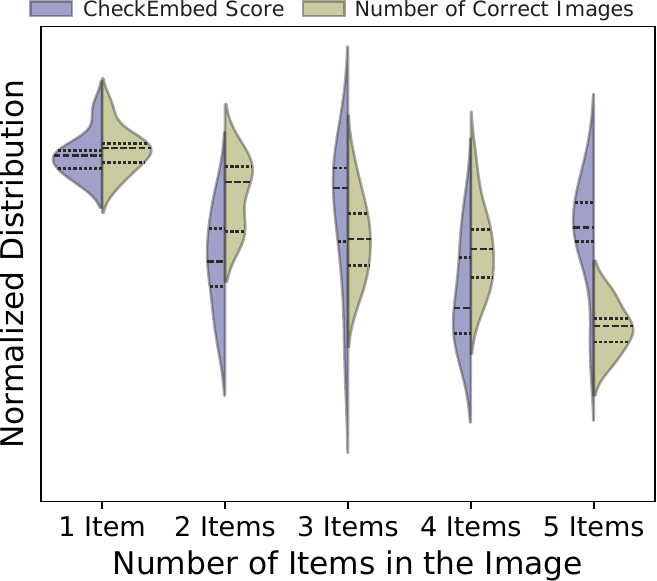}
    \caption{\textbf{Hallucination detection in vision models with \nameA}. We score response quality of a vision model without real-world references or auxiliary models. Both \nameS scores and correct image counts are normalized. As more items are requested, hallucinations rise, reducing correctness and \nameS scores.}
    \label{fig:hallucinate_images}
    \vspaceSQ{-2em}
\end{wrapfigure}
\fi

Hallucinations in images refer to instances where generative models produce visual content that includes non-existent, incorrect, or extraneous elements not grounded in the input prompt or data. These may manifest as anatomically implausible features (e.g., a hand with six fingers), omitted or miscounted objects, or the inclusion of unrelated artifacts such as logos or text.

We induce hallucinations by prompting the model to generate a specified number of objects in the image. This enables a controlled degradation: generation is consistently accurate with a single object, while accuracy declines as the requested count increases. Due to the need for manual scoring to ensure quality, the evaluation scope is limited, resulting in higher variance in the results. Still, Figure~\ref{fig:hallucinate_images} shows a correlation between the number of correctly rendered images and the corresponding \nameS score. The complete setup and further examples are in Appendix~\ref{sec:app:multimodal}.

\subsection{Analysis of Scalability}
\label{sec:runtimes}

\begin{wrapfigure}{r}{0.4\textwidth}
  \ifconf
  \vspaceSQ{-4em}
  \else
  \vspaceSQ{-2em}
  \fi
  \centering
  \includegraphics[width=0.95\linewidth]{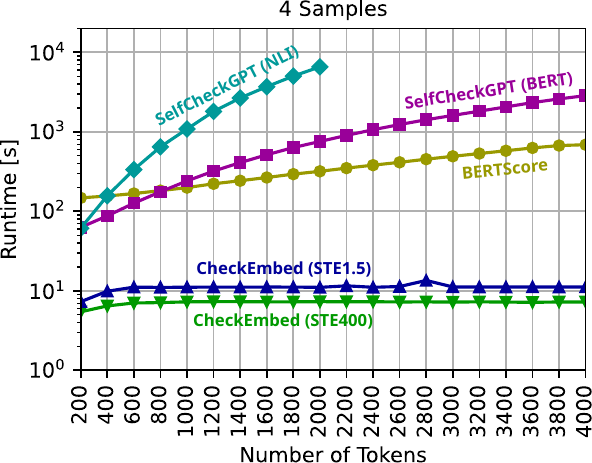}
  \caption{\textbf{Advantages of the \nameAS runtime over other baselines} while varying the answer token size.}
  %We used an NVIDIA RTX3090 GPU for this experiment. Please note the logscale y axis.
  \label{fig:runtimes}
  \vspace{-1.5em}
\end{wrapfigure}

We also investigate the running times of baselines that are highly effective for answer verification and hallucination detection (SelfCheckGPT, BERTScore, \nameA).
Example results are in Figure~\ref{fig:runtimes}. The numbers for each datapoint
correspond to the total runtime required to construct 20 embeddings and to
compute similarity scores between all embedding pairs. We show runtimes for
\nameAS with the Stella models as their smaller model sizes (435M, 1.5B) are
comparable to the best available bidirectional embedding models that can be used
with BERTScore and SelfCheckGPT (e.g., microsoft/deberta-xlarge-mnli has 750M
parameters). \nameA, while using the Stella models, maintains a constant
evaluation time regardless of the sample size or token number for the text
chunks. All comparison baselines exhibit an inflation of their runtime, as we increase the
number of samples or the token length of the inputs, making \nameAS 30$\times$--300$\times$
faster. We present additional results for GPT and other embedding models in Appendix~\ref{sec:additional_runtime_results}. These
results further showcases the high performance of \nameA, rooted in its
simplicity: \textit{all that is required to compute is a single embedding of a textual
answer or its chunk}.

\if 0
\begin{figure}[hbt!]
  \includegraphics[width=\linewidth]{plots/runtime_with_hallu.pdf}
  \caption{\textbf{Comparison of running times of \nameAS and other baselines while varying the number of samples per datapoint.} We used an NVIDIA RTX3090 GPU for this experiment. Please note the logscale y axis.}
  \label{fig:runtimes}
\end{figure}

\begin{wrapfigure}{l}{0.58\linewidth}
  \centering
  \vspace{-1em}
  \includegraphics[width=\linewidth]{plots/runtime_hallu.pdf}
  \vspace{-1em}
  \caption{\textbf{Comparison of running times of \nameAS and other baselines while varying the number of samples per datapoint.} We used an NVIDIA RTX3090 GPU for this experiment. Please note the logscale y axis.}
  \label{fig:runtimes}
\end{wrapfigure}

\lorenzo{Do we want to add in some way this plot for HalluDetect inference times?
\begin{itemize}
    \item We can argue that it is fast, surely faster than SCGPT and BERTScore, but it is as fast as we currently are. Maybe it is a little faster than us, but it must be considered that the faster models for HalluDetect aren't so great actually.
    \item It must also be considered it needs to be trained to work properly, training depending on the dataset size and the modek used can take from few minutes to several hours.
\end{itemize}}
\fi

\subsection{Analysis of Varying the Number of Sampled LLM Answers}

\begin{wrapfigure}{r}{0.4\linewidth}
  \centering
  \vspaceSQ{-2em}
  \includegraphics[width=0.8\linewidth]{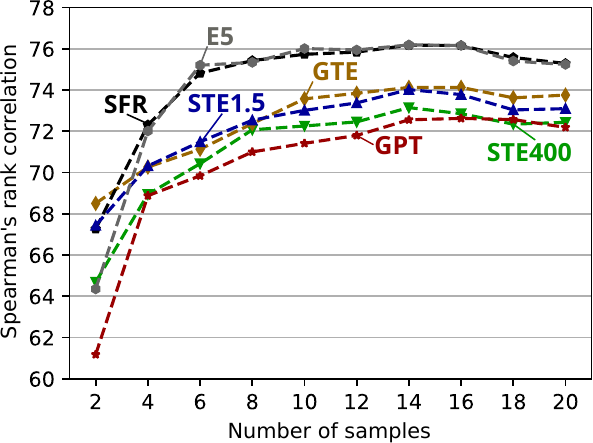}
  \vspace{-0.75em}
  \caption{{Accuracy of \nameAS with different embedding models while varying the number of sampled LLM answers.}}
  \label{fig:accuracy}
  \ifconf
  \vspace{-2em}
  \else
  \vspace{-3em}
  \fi
\end{wrapfigure}

Finally, we also study how the accuracy of \nameAS is influenced by the sample size per datapoint. We conducted this evaluation on the WikiBio dataset and plot the Spearman's rank correlation coefficient while varying the number of samples in Figure~\ref{fig:accuracy}.
While all embedding models show an accuracy increase with more samples, the accuracy starts to stabilize with 8 samples (6 samples for SFR and E5), at which point the gain from using additional samples might be offset by the additional cost.

\maciej{"how different embedding models affect the results?"}

\maciej{For evaluation, to show more clearly our advantages, how about adding a new (additional one, besides others) metric, where we normalize the accuracy per time or cost (e.g., number of answers verified per second? Number of pairs compared per second?). --> " comparable performance by other metrics (e.g., NLI variant of SelfCheckGPT) "}

\section{Related Work}
\label{sec:rw}

We have already discussed other \textbf{verification methods} in Section~\ref{sec:intro} and in Table~\ref{tab:qualitative}.

\textbf{Embedding-based similarity methods} such as BERTScore or Sentence-BERT compute semantic alignment between textual sequences by encoding tokens, sentences, or entire passages into dense vector representations and comparing them using similarity metrics like cosine distance. These methods serve as core building blocks in many verification and hallucination detection schemes (including \name), providing a mechanism to compare generated outputs against references or other outputs. They often operate at the token or sentence level, leading to computational inefficiencies for long, open-ended responses. Here, \nameS leverages whole-answer embeddings, allowing it to assess global semantic agreement across complete outputs using a single vector per answer. % This shift enables a more scalable and robust verification process while preserving the semantic fidelity offered by embedding-based comparison.

% \subsection{\nameS and Other Aspects of LLM Verification Landscape}

\textbf{Explainable AI (XAI)} and more generally \textbf{trustworthy AI} aim to make AI models interpretable and understandable~\citep{Longo_2024, zhao2024explainability, luo2024from}. Recent works include self-explaining models~\citep{huang2023can,madsen2024are} and post-hoc explanation methods~\citep{vale2022explainable,kroeger2024are}. \nameAS supports interpretability through embedding-based heatmaps and summary statistics that visualize semantic agreement across model outputs.

\textbf{Hallucination detection in LLMs} has received significant attention~\citep{rawte2023survey, zhang2023sirens, huang2023survey}. Proposed solutions include fact-checking~\citep{zhang2024knowhalu, chern2023factool}, self-consistency~\citep{manakul2023selfcheckgpt}, and RAG~\citep{zhu2024large}. \nameAS addresses hallucination detection in both open- and closed-ended settings without external tools or references, and with higher semantic fidelity than token- or fact-level comparators. Its design also allows integration with retrieval or user feedback systems if desired.

\textbf{LLM-based agents}~\citep{xi2023rise, wei2022chain, yao2023tree, besta2023graph} are gaining traction for complex autonomous tasks. These agents benefit from accurate internal self-evaluation and answer validation. \nameAS provides a scalable verification mechanism for agent-generated outputs, enabling real-time quality control and decision gating without human feedback or retraining.

% \textbf{Hallucinations in vision models} are typically evaluated using auxiliary models~\cite{inceptionscore,lim2025evaluatingimagehallucinationtexttoimage} including classifiers or by statistically comparing to real-world data~\cite{Frechet_Inception_Distance,Tiv_Hallucination_MICCAI2024} to assess plausibility. In contrast, \nameAS detects hallucinations without dependencies, offering a reference-free evaluation signal.

\textbf{Philosophical and epistemological questions} in AI concern how machine learning systems produce and validate knowledge~\citep{Fleisher_2022, shanahan2023talking, mahowald2024dissociating}. \nameAS contributes to this line of work by constituting a mechanism for assessing the internal consistency and plausibility of AI-generated knowledge, aligning with efforts to improve the interpretability, reliability, and epistemic grounding of AI systems.

\section{Conclusion}
%\vspace{-0.5em}

LLMs are revolutionizing various domains, yet effective verification for open-ended tasks remains a significant challenge. Established methods, which focus on token- and sentence-level analysis, fall short in scalability and effectiveness. Addressing this gap is crucial as applications of LLMs expand, necessitating robust mechanisms to ensure the accuracy and reliability of their outputs.

To this end, we introduce \name, a scalable approach to LLM verification. \nameS leverages the effectiveness of answer-level embeddings to compare LLM answers with one another and the potential ground-truth. By transforming complex textual answers into individual embeddings using modern decoder-only based models like GPT Text Embedding Large, \nameS makes the verification process simple, accurate, and scalable. This straightforward methodology integrates seamlessly with modern data analytics infrastructure, highlighting its practical applicability and ease of deployment.

Our comprehensive verification pipeline includes metrics and tools for assessing the veracity of LLM answers, such as heatmaps of similarities between embeddings of answers, the ground-truth, and statistical summaries. These tools provide detailed insights into the quality of LLM outputs and facilitate practical decision-making in real-world deployments.

\nameS has been tested on numerous tasks, for example term extraction in long legal documents. The results demonstrated significant improvements in accuracy and runtime performance compared to existing methods such as BERTScore, SelfCheckGPT, and LLM-as-a-Judge. We also provide evidence that \nameS can effectively detect hallucinations in vision models. These findings underscore the potential of \nameS to enhance  verification not only for LLMs but also for other classes of models.

% Need citation fo LLM-as-a-Judge? nope, I removed citations to others above ;) faster - saw that

% \lorenzo{LIMITATIONS:
% \begin{itemize}
%    \item We work on the assumption that when LLM hallucinate, they will hallucinate in different ways when prompting them N times with the same exact request. This obviously is possible only when LLM use sampling to choose the next token and when temperature is more than 0. If LLM do not use sampling and/or have a temperature of 0 they will always output the same answer, thus all the sample will have the exact same content resulting in a perfect score from our method.
%    This can also happen when sampling: the hallucinated sequence of tokens has a very high probability during inference and thus gets chosen always, our method would result inefficient to detect this types of hallucinations, but from the observed result we can at least say that this is not happening much or we would have much worst scores.
%    \item We work on the passage level hallucination detection and thus we are not able to detect where the hallucination is present, it is true that also SCGPT do not individuate the position perfectly, but at least it shrink down to the sentence level the position of the hellucination.
%    So CE its really useful, scalable and fast in context where you do not want to find the hallucination and fix it, but you just want to know if the answer is hallucinated or not.
% \end{itemize}
% }

\begin{ack}
We thank Hussein Harake, Colin McMurtrie, Mark Klein, Angelo Mangili, and the whole CSCS team granting access to the Ault, Daint and Alps machines, and for their excellent technical support. We thank Timo Schneider for immense help with infrastructure at SPCL. This project received funding from the European Research Council (Project PSAP, No.~101002047), and the European High-Performance Computing Joint Undertaking (JU) under grant agreement No.~955513 (MAELSTROM). This project was supported by the ETH Future Computing Laboratory (EFCL), financed by a donation from Huawei Technologies. This project received funding from the European Union’s HE research and innovation programme under the grant agreement No. 101070141 (Project GLACIATION). We also acknowledge Polish high-performance computing infrastructure PLGrid (HPC Center: ACK Cyfronet AGH) for providing computer facilities and support within computational grant no.~PLG/2024/017103.
\end{ack}

\bibliographystyle{ACM-Reference-Format}
\bibliography{references.complete}

%%% -*-BibTeX-*-
%%% Do NOT edit. File created by BibTeX with style
%%% ACM-Reference-Format-Journals [18-Jan-2012].

\begin{thebibliography}{71}

%%% ====================================================================
%%% NOTE TO THE USER: you can override these defaults by providing
%%% customized versions of any of these macros before the \bibliography
%%% command.  Each of them MUST provide its own final punctuation,
%%% except for \shownote{}, \showDOI{}, and \showURL{}.  The latter two
%%% do not use final punctuation, in order to avoid confusing it with
%%% the Web address.
%%%
%%% To suppress output of a particular field, define its macro to expand
%%% to an empty string, or better, \unskip, like this:
%%%
%%% \newcommand{\showDOI}[1]{\unskip}   % LaTeX syntax
%%%
%%% \def \showDOI #1{\unskip}           % plain TeX syntax
%%%
%%% ====================================================================

\ifx \showCODEN    \undefined \def \showCODEN     #1{\unskip}     \fi
\ifx \showDOI      \undefined \def \showDOI       #1{#1}\fi
\ifx \showISBNx    \undefined \def \showISBNx     #1{\unskip}     \fi
\ifx \showISBNxiii \undefined \def \showISBNxiii  #1{\unskip}     \fi
\ifx \showISSN     \undefined \def \showISSN      #1{\unskip}     \fi
\ifx \showLCCN     \undefined \def \showLCCN      #1{\unskip}     \fi
\ifx \shownote     \undefined \def \shownote      #1{#1}          \fi
\ifx \showarticletitle \undefined \def \showarticletitle #1{#1}   \fi
\ifx \showURL      \undefined \def \showURL       {\relax}        \fi
% The following commands are used for tagged output and should be
% invisible to TeX
\providecommand\bibfield[2]{#2}
\providecommand\bibinfo[2]{#2}
\providecommand\natexlab[1]{#1}
\providecommand\showeprint[2][]{arXiv:#2}

\bibitem[Besta et~al\mbox{.}(2024)]%
        {besta2023graph}
\bibfield{author}{\bibinfo{person}{Maciej Besta}, \bibinfo{person}{Nils Blach},
  \bibinfo{person}{Ales Kubicek}, \bibinfo{person}{Robert Gerstenberger},
  \bibinfo{person}{Michal Podstawski}, \bibinfo{person}{Lukas Gianinazzi},
  \bibinfo{person}{Joanna Gajda}, \bibinfo{person}{Tomasz Lehmann},
  \bibinfo{person}{Hubert Niewiadomski}, \bibinfo{person}{Piotr Nyczyk}, {and}
  \bibinfo{person}{Torsten Hoefler}.} \bibinfo{year}{2024}\natexlab{}.
\newblock \showarticletitle{Graph of Thoughts: Solving Elaborate Problems with
  Large Language Models}.
\newblock \bibinfo{journal}{\emph{Proceedings of the AAAI Conference on
  Artificial Intelligence}} \bibinfo{volume}{38}, \bibinfo{number}{16}
  (\bibinfo{date}{March} \bibinfo{year}{2024}), \bibinfo{pages}{17682--17690}.
\newblock
\urldef\tempurl%
\url{https://doi.org/10.1609/aaai.v38i16.29720}
\showDOI{\tempurl}


\bibitem[Blelloch and Maggs(2010)]%
        {blelloch2010parallel}
\bibfield{author}{\bibinfo{person}{Guy~E. Blelloch} {and}
  \bibinfo{person}{Bruce~M. Maggs}.} \bibinfo{year}{2010}\natexlab{}.
\newblock \bibinfo{booktitle}{\emph{Parallel Algorithms} (\bibinfo{edition}{2}
  ed.)}.
\newblock \bibinfo{publisher}{Chapman \& Hall/CRC}, \bibinfo{address}{London,
  United Kingdom}, Chapter~25, \bibinfo{pages}{1--44}.
\newblock
\showISBNx{9781584888208}


\bibitem[Chang et~al\mbox{.}(2024)]%
        {chang2023survey}
\bibfield{author}{\bibinfo{person}{Yupeng Chang}, \bibinfo{person}{Xu Wang},
  \bibinfo{person}{Jindong Wang}, \bibinfo{person}{Yuan Wu},
  \bibinfo{person}{Linyi Yang}, \bibinfo{person}{Kaijie Zhu},
  \bibinfo{person}{Hao Chen}, \bibinfo{person}{Xiaoyuan Yi},
  \bibinfo{person}{Cunxiang Wang}, \bibinfo{person}{Yidong Wang},
  \bibinfo{person}{Wei Ye}, \bibinfo{person}{Yue Zhang}, \bibinfo{person}{Yi
  Chang}, \bibinfo{person}{Philip~S. Yu}, \bibinfo{person}{Qiang Yang}, {and}
  \bibinfo{person}{Xing Xie}.} \bibinfo{year}{2024}\natexlab{}.
\newblock \showarticletitle{A Survey on Evaluation of Large Language Models}.
\newblock \bibinfo{journal}{\emph{ACM Trans. Intell. Syst. Technol.}}
  \bibinfo{volume}{15}, \bibinfo{number}{3}, Article \bibinfo{articleno}{39}
  (\bibinfo{date}{March} \bibinfo{year}{2024}), \bibinfo{numpages}{45}~pages.
\newblock
\showISSN{2157-6904}
\urldef\tempurl%
\url{https://doi.org/10.1145/3641289}
\showDOI{\tempurl}


\bibitem[Chern et~al\mbox{.}(2023)]%
        {chern2023factool}
\bibfield{author}{\bibinfo{person}{I-Chun Chern}, \bibinfo{person}{Steffi
  Chern}, \bibinfo{person}{Shiqi Chen}, \bibinfo{person}{Weizhe Yuan},
  \bibinfo{person}{Kehua Feng}, \bibinfo{person}{Chunting Zhou},
  \bibinfo{person}{Junxian He}, \bibinfo{person}{Graham Neubig}, {and}
  \bibinfo{person}{Pengfei Liu}.} \bibinfo{year}{2023}\natexlab{}.
\newblock \bibinfo{title}{FacTool: Factuality Detection in Generative AI -- A
  Tool Augmented Framework for Multi-Task and Multi-Domain Scenarios}.
\newblock
\newblock
\urldef\tempurl%
\url{https://doi.org/10.48550/arXiv.2307.13528}
\showDOI{\tempurl}
\showeprint[arXiv]{2307.13528}~[cs.CL]


\bibitem[Elaraby et~al\mbox{.}(2023)]%
        {elaraby2023halo}
\bibfield{author}{\bibinfo{person}{Mohamed Elaraby}, \bibinfo{person}{Mengyin
  Lu}, \bibinfo{person}{Jacob Dunn}, \bibinfo{person}{Xueying Zhang},
  \bibinfo{person}{Yu Wang}, \bibinfo{person}{Shizhu Liu},
  \bibinfo{person}{Pingchuan Tian}, \bibinfo{person}{Yuping Wang}, {and}
  \bibinfo{person}{Yuxuan Wang}.} \bibinfo{year}{2023}\natexlab{}.
\newblock \bibinfo{title}{Halo: Estimation and Reduction of Hallucinations in
  Open-Source Weak Large Language Models}.
\newblock
\newblock
\urldef\tempurl%
\url{https://doi.org/10.48550/arXiv.2308.11764}
\showDOI{\tempurl}
\showeprint[arXiv]{2308.11764}~[cs.CL]


\bibitem[Esser et~al\mbox{.}(2024)]%
        {pmlr-v235-esser24a}
\bibfield{author}{\bibinfo{person}{Patrick Esser}, \bibinfo{person}{Sumith
  Kulal}, \bibinfo{person}{Andreas Blattmann}, \bibinfo{person}{Rahim
  Entezari}, \bibinfo{person}{Jonas M\"{u}ller}, \bibinfo{person}{Harry Saini},
  \bibinfo{person}{Yam Levi}, \bibinfo{person}{Dominik Lorenz},
  \bibinfo{person}{Axel Sauer}, \bibinfo{person}{Frederic Boesel},
  \bibinfo{person}{Dustin Podell}, \bibinfo{person}{Tim Dockhorn},
  \bibinfo{person}{Zion English}, {and} \bibinfo{person}{Robin Rombach}.}
  \bibinfo{year}{2024}\natexlab{}.
\newblock \showarticletitle{Scaling Rectified Flow Transformers for
  High-Resolution Image Synthesis}. In \bibinfo{booktitle}{\emph{Proceedings of
  the 41st International Conference on Machine Learning (ICML '24)}} (Vienna
  Austria) \emph{(\bibinfo{series}{Proceedings of Machine Learning Research},
  Vol.~\bibinfo{volume}{235})}, \bibfield{editor}{\bibinfo{person}{Ruslan
  Salakhutdinov}, \bibinfo{person}{Zico Kolter}, \bibinfo{person}{Katherine
  Heller}, \bibinfo{person}{Adrian Weller}, \bibinfo{person}{Nuria Oliver},
  \bibinfo{person}{Jonathan Scarlett}, {and} \bibinfo{person}{Felix
  Berkenkamp}} (Eds.). \bibinfo{publisher}{PMLR}, \bibinfo{address}{New York,
  NY, USA}, \bibinfo{pages}{12606--12633}.
\newblock
\urldef\tempurl%
\url{https://proceedings.mlr.press/v235/esser24a.html}
\showURL{%
\tempurl}


\bibitem[Fleisher(2022)]%
        {Fleisher_2022}
\bibfield{author}{\bibinfo{person}{Will Fleisher}.}
  \bibinfo{year}{2022}\natexlab{}.
\newblock \showarticletitle{Understanding, Idealization, and Explainable AI}.
\newblock \bibinfo{journal}{\emph{Episteme}} \bibinfo{volume}{19},
  \bibinfo{number}{4} (\bibinfo{date}{Dec.} \bibinfo{year}{2022}),
  \bibinfo{pages}{534--560}.
\newblock
\urldef\tempurl%
\url{https://doi.org/10.1017/epi.2022.39}
\showDOI{\tempurl}


\bibitem[Fortune and Wyllie(1978)]%
        {fortune1978parallelism}
\bibfield{author}{\bibinfo{person}{Steven Fortune} {and} \bibinfo{person}{James
  Wyllie}.} \bibinfo{year}{1978}\natexlab{}.
\newblock \showarticletitle{Parallelism in Random Access Machines}. In
  \bibinfo{booktitle}{\emph{Proceedings of the Tenth Annual ACM Symposium on
  Theory of Computing}} (San Diego, CA, USA) \emph{(\bibinfo{series}{STOC
  '78})}. \bibinfo{publisher}{Association for Computing Machinery},
  \bibinfo{address}{New York, NY, USA}, \bibinfo{pages}{114--118}.
\newblock
\showISBNx{9781450374378}
\urldef\tempurl%
\url{https://doi.org/10.1145/800133.804339}
\showDOI{\tempurl}


\bibitem[Fu et~al\mbox{.}(2024)]%
        {fu2024gptscore}
\bibfield{author}{\bibinfo{person}{Jinlan Fu}, \bibinfo{person}{See-Kiong Ng},
  \bibinfo{person}{Zhengbao Jiang}, {and} \bibinfo{person}{Pengfei Liu}.}
  \bibinfo{year}{2024}\natexlab{}.
\newblock \showarticletitle{GPTScore: Evaluate as You Desire}. In
  \bibinfo{booktitle}{\emph{Proceedings of the 2024 Conference of the North
  American Chapter of the Association for Computational Linguistics: Human
  Language Technologies (Volume 1: Long Papers)}} (Mexico City, Mexico)
  \emph{(\bibinfo{series}{NAACL '24})},
  \bibfield{editor}{\bibinfo{person}{Kevin Duh}, \bibinfo{person}{Helena
  Gomez}, {and} \bibinfo{person}{Steven Bethard}} (Eds.).
  \bibinfo{publisher}{Association for Computational Linguistics},
  \bibinfo{address}{Kerrville, TX, USA}, \bibinfo{pages}{6556--6576}.
\newblock
\urldef\tempurl%
\url{https://doi.org/10.18653/v1/2024.naacl-long.365}
\showDOI{\tempurl}


\bibitem[Gu et~al\mbox{.}(2025)]%
        {gu2024survey}
\bibfield{author}{\bibinfo{person}{Jiawei Gu}, \bibinfo{person}{Xuhui Jiang},
  \bibinfo{person}{Zhichao Shi}, \bibinfo{person}{Hexiang Tan},
  \bibinfo{person}{Xuehao Zhai}, \bibinfo{person}{Chengjin Xu},
  \bibinfo{person}{Wei Li}, \bibinfo{person}{Yinghan Shen},
  \bibinfo{person}{Shengjie Ma}, \bibinfo{person}{Honghao Liu},
  \bibinfo{person}{Saizhuo Wang}, \bibinfo{person}{Kun Zhang},
  \bibinfo{person}{Yuanzhuo Wang}, \bibinfo{person}{Wen Gao},
  \bibinfo{person}{Lionel Ni}, {and} \bibinfo{person}{Jian Guo}.}
  \bibinfo{year}{2025}\natexlab{}.
\newblock \bibinfo{title}{A Survey on LLM-as-a-Judge}.
\newblock
\newblock
\urldef\tempurl%
\url{https://doi.org/10.48550/arXiv.2411.15594}
\showDOI{\tempurl}
\showeprint[arXiv]{2411.15594}~[cs.CL]


\bibitem[Guu et~al\mbox{.}(2020)]%
        {guu2020realm}
\bibfield{author}{\bibinfo{person}{Kelvin Guu}, \bibinfo{person}{Kenton Lee},
  \bibinfo{person}{Zora Tung}, \bibinfo{person}{Panupong Pasupat}, {and}
  \bibinfo{person}{Ming-Wei Chang}.} \bibinfo{year}{2020}\natexlab{}.
\newblock \showarticletitle{Retrieval-Augmented Language Model Pre-Training}.
  In \bibinfo{booktitle}{\emph{Proceedings of the 37th International Conference
  on Machine Learning (ICML '20)}} (Virtual Event)
  \emph{(\bibinfo{series}{Proceedings of Machine Learning Research},
  Vol.~\bibinfo{volume}{119})}, \bibfield{editor}{\bibinfo{person}{Hal~Daumé
  III} {and} \bibinfo{person}{Aarti Singh}} (Eds.). \bibinfo{publisher}{PMLR},
  \bibinfo{address}{New York, NY, USA}, \bibinfo{pages}{3929--3938}.
\newblock
\urldef\tempurl%
\url{https://proceedings.mlr.press/v119/guu20a.html}
\showURL{%
\tempurl}


\bibitem[He et~al\mbox{.}(2021)]%
        {he2021deberta}
\bibfield{author}{\bibinfo{person}{Pengcheng He}, \bibinfo{person}{Xiaodong
  Liu}, \bibinfo{person}{Jianfeng Gao}, {and} \bibinfo{person}{Weizhu Chen}.}
  \bibinfo{year}{2021}\natexlab{}.
\newblock \showarticletitle{DeBERTa: Decoding-Enhanced BERT With Disentangled
  Attention}. In \bibinfo{booktitle}{\emph{Proccedings of the Ninth
  International Conference on Learning Representations}} (Virtual Event)
  \emph{(\bibinfo{series}{ICLR '21})}. \bibinfo{publisher}{OpenReview},
  \bibinfo{address}{Amherst, MA, USA}, \bibinfo{numpages}{21}~pages.
\newblock
\urldef\tempurl%
\url{https://openreview.net/forum?id=XPZIaotutsD}
\showURL{%
\tempurl}


\bibitem[Heusel et~al\mbox{.}(2017)]%
        {Frechet_Inception_Distance}
\bibfield{author}{\bibinfo{person}{Martin Heusel}, \bibinfo{person}{Hubert
  Ramsauer}, \bibinfo{person}{Thomas Unterthiner}, \bibinfo{person}{Bernhard
  Nessler}, {and} \bibinfo{person}{Sepp Hochreiter}.}
  \bibinfo{year}{2017}\natexlab{}.
\newblock \showarticletitle{GANs Trained by a Two Time-Scale Update Rule
  Converge to a Local Nash Equilibrium}. In
  \bibinfo{booktitle}{\emph{Proceedings of the 31st International Conference on
  Neural Information Processing Systems}} (Long Beach, CA, USA)
  \emph{(\bibinfo{series}{Advances in Neural Information Processing Systems},
  Vol.~\bibinfo{volume}{30})}, \bibfield{editor}{\bibinfo{person}{I.~Guyon},
  \bibinfo{person}{U.~Von Luxburg}, \bibinfo{person}{S.~Bengio},
  \bibinfo{person}{H.~Wallach}, \bibinfo{person}{R.~Fergus},
  \bibinfo{person}{S.~Vishwanathan}, {and} \bibinfo{person}{R.~Garnett}}
  (Eds.). \bibinfo{publisher}{Curran Associates}, \bibinfo{address}{Red Hook,
  NY, USA}, \bibinfo{pages}{6629--6640}.
\newblock
\showISBNx{9781510860964}
\urldef\tempurl%
\url{https://proceedings.neurips.cc/paper_files/paper/2017/hash/8a1d694707eb0fefe65871369074926d-Abstract.html}
\showURL{%
\tempurl}


\bibitem[Huang et~al\mbox{.}(2025)]%
        {huang2023survey}
\bibfield{author}{\bibinfo{person}{Lei Huang}, \bibinfo{person}{Weijiang Yu},
  \bibinfo{person}{Weitao Ma}, \bibinfo{person}{Weihong Zhong},
  \bibinfo{person}{Zhangyin Feng}, \bibinfo{person}{Haotian Wang},
  \bibinfo{person}{Qianglong Chen}, \bibinfo{person}{Weihua Peng},
  \bibinfo{person}{Xiaocheng Feng}, \bibinfo{person}{Bing Qin}, {and}
  \bibinfo{person}{Ting Liu}.} \bibinfo{year}{2025}\natexlab{}.
\newblock \showarticletitle{A Survey on Hallucination in Large Language Models:
  Principles, Taxonomy, Challenges, and Open Questions}.
\newblock \bibinfo{journal}{\emph{ACM Trans. Inf. Syst.}} \bibinfo{volume}{43},
  \bibinfo{number}{2}, Article \bibinfo{articleno}{42} (\bibinfo{date}{Jan.}
  \bibinfo{year}{2025}), \bibinfo{numpages}{55}~pages.
\newblock
\showISSN{1046-8188}
\urldef\tempurl%
\url{https://doi.org/10.1145/3703155}
\showDOI{\tempurl}


\bibitem[Huang et~al\mbox{.}(2023)]%
        {huang2023can}
\bibfield{author}{\bibinfo{person}{Shiyuan Huang}, \bibinfo{person}{Siddarth
  Mamidanna}, \bibinfo{person}{Shreedhar Jangam}, \bibinfo{person}{Yilun Zhou},
  {and} \bibinfo{person}{Leilani~H. Gilpin}.} \bibinfo{year}{2023}\natexlab{}.
\newblock \bibinfo{title}{Can Large Language Models Explain Themselves? A Study
  of LLM-Generated Self-Explanations}.
\newblock
\newblock
\urldef\tempurl%
\url{https://doi.org/10.48550/arXiv.2310.11207}
\showDOI{\tempurl}
\showeprint[arXiv]{2310.11207}~[cs.CL]


\bibitem[Keller et~al\mbox{.}(2000)]%
        {keller2000practical}
\bibfield{author}{\bibinfo{person}{J\"{o}rg Keller}, \bibinfo{person}{Christoph
  Kessler}, {and} \bibinfo{person}{Jesper~Larsson Tr\"{a}ff}.}
  \bibinfo{year}{2000}\natexlab{}.
\newblock \bibinfo{booktitle}{\emph{Practical PRAM Programming}}.
\newblock \bibinfo{publisher}{John Wiley \& Sons}, \bibinfo{address}{Hoboken,
  NJ, USA}.
\newblock
\showISBNx{0471353515}


\bibitem[Kroeger et~al\mbox{.}(2024)]%
        {kroeger2024are}
\bibfield{author}{\bibinfo{person}{Nicholas Kroeger}, \bibinfo{person}{Dan
  Ley}, \bibinfo{person}{Satyapriya Krishna}, \bibinfo{person}{Chirag Agarwal},
  {and} \bibinfo{person}{Himabindu Lakkaraju}.}
  \bibinfo{year}{2024}\natexlab{}.
\newblock \bibinfo{title}{In-Context Explainers: Harnessing LLMs for Explaining
  Black Box Models}.
\newblock
\newblock
\urldef\tempurl%
\url{https://doi.org/10.48550/arXiv.2310.05797}
\showDOI{\tempurl}
\showeprint[arXiv]{2310.05797}~[cs.CL]


\bibitem[Laban et~al\mbox{.}(2022)]%
        {laban2022summac}
\bibfield{author}{\bibinfo{person}{Philippe Laban}, \bibinfo{person}{Tobias
  Schnabel}, \bibinfo{person}{Paul~N. Bennett}, {and} \bibinfo{person}{Marti~A.
  Hearst}.} \bibinfo{year}{2022}\natexlab{}.
\newblock \showarticletitle{SummaC: Re-Visiting NLI-Based Models for
  Inconsistency Detection in Summarization}.
\newblock \bibinfo{journal}{\emph{Transactions of the Association for
  Computational Linguistics}}  \bibinfo{volume}{10} (\bibinfo{year}{2022}),
  \bibinfo{pages}{163--177}.
\newblock
\urldef\tempurl%
\url{https://doi.org/10.1162/tacl_a_00453}
\showDOI{\tempurl}


\bibitem[Lebret et~al\mbox{.}(2016)]%
        {lebret2016neural}
\bibfield{author}{\bibinfo{person}{R{\'e}mi Lebret}, \bibinfo{person}{David
  Grangier}, {and} \bibinfo{person}{Michael Auli}.}
  \bibinfo{year}{2016}\natexlab{}.
\newblock \showarticletitle{Neural Text Generation from Structured Data with
  Application to the Biography Domain}. In
  \bibinfo{booktitle}{\emph{Proceedings of the 2016 Conference on Empirical
  Methods in Natural Language Processing}} (Austin, TX, USA)
  \emph{(\bibinfo{series}{EMNLP '16})}, \bibfield{editor}{\bibinfo{person}{Jian
  Su}, \bibinfo{person}{Kevin Duh}, {and} \bibinfo{person}{Xavier Carreras}}
  (Eds.). \bibinfo{publisher}{Association for Computational Linguistics},
  \bibinfo{address}{Kerrville, TX, USA}, \bibinfo{pages}{1203--1213}.
\newblock
\urldef\tempurl%
\url{https://doi.org/10.18653/v1/D16-1128}
\showDOI{\tempurl}


\bibitem[Lee et~al\mbox{.}(2025)]%
        {lee2024nv}
\bibfield{author}{\bibinfo{person}{Chankyu Lee}, \bibinfo{person}{Rajarshi
  Roy}, \bibinfo{person}{Mengyao Xu}, \bibinfo{person}{Jonathan Raiman},
  \bibinfo{person}{Mohammad Shoeybi}, \bibinfo{person}{Bryan Catanzaro}, {and}
  \bibinfo{person}{Wei Ping}.} \bibinfo{year}{2025}\natexlab{}.
\newblock \showarticletitle{NV-Embed: Improved Techniques for Training LLMs as
  Generalist Embedding Models}. In \bibinfo{booktitle}{\emph{Proceedings of the
  Thirteenth International Conference on Learning Representations}} (Singapore)
  \emph{(\bibinfo{series}{ICLR '25})}. \bibinfo{publisher}{OpenReview},
  \bibinfo{address}{Amherst, MA, USA}, \bibinfo{numpages}{24}~pages.
\newblock
\urldef\tempurl%
\url{https://openreview.net/forum?id=lgsyLSsDRe}
\showURL{%
\tempurl}


\bibitem[Lewis et~al\mbox{.}(2020)]%
        {lewis2020retrieval}
\bibfield{author}{\bibinfo{person}{Patrick Lewis}, \bibinfo{person}{Ethan
  Perez}, \bibinfo{person}{Aleksandra Piktus}, \bibinfo{person}{Fabio Petroni},
  \bibinfo{person}{Vladimir Karpukhin}, \bibinfo{person}{Naman Goyal},
  \bibinfo{person}{Heinrich K\"{u}ttler}, \bibinfo{person}{Mike Lewis},
  \bibinfo{person}{Wen-tau Yih}, \bibinfo{person}{Tim Rockt\"{a}schel},
  \bibinfo{person}{Sebastian Riedel}, {and} \bibinfo{person}{Douwe Kiela}.}
  \bibinfo{year}{2020}\natexlab{}.
\newblock \showarticletitle{Retrieval-Augmented Generation for
  Knowledge-Intensive NLP Tasks}. In \bibinfo{booktitle}{\emph{Proceedings of
  the Thirty-Fourth Annual Conference on Neural Information Processing Systems
  (NeurIPS '20)}} (Virtual Event) \emph{(\bibinfo{series}{Advances in Neural
  Information Processing Systems}, Vol.~\bibinfo{volume}{33})},
  \bibfield{editor}{\bibinfo{person}{H.~Larochelle},
  \bibinfo{person}{M.~Ranzato}, \bibinfo{person}{R.~Hadsell},
  \bibinfo{person}{M.F. Balcan}, {and} \bibinfo{person}{H.~Lin}} (Eds.).
  \bibinfo{publisher}{Curran Associates}, \bibinfo{address}{Red Hook, NY, USA},
  \bibinfo{pages}{9459--9474}.
\newblock
\urldef\tempurl%
\url{https://proceedings.neurips.cc/paper_files/paper/2020/hash/6b493230205f780e1bc26945df7481e5-Abstract.html}
\showURL{%
\tempurl}


\bibitem[Li et~al\mbox{.}(2025)]%
        {li2024generation}
\bibfield{author}{\bibinfo{person}{Dawei Li}, \bibinfo{person}{Bohan Jiang},
  \bibinfo{person}{Liangjie Huang}, \bibinfo{person}{Alimohammad Beigi},
  \bibinfo{person}{Chengshuai Zhao}, \bibinfo{person}{Zhen Tan},
  \bibinfo{person}{Amrita Bhattacharjee}, \bibinfo{person}{Yuxuan Jiang},
  \bibinfo{person}{Canyu Chen}, \bibinfo{person}{Tianhao Wu},
  \bibinfo{person}{Kai Shu}, \bibinfo{person}{Lu Cheng}, {and}
  \bibinfo{person}{Huan Liu}.} \bibinfo{year}{2025}\natexlab{}.
\newblock \bibinfo{title}{From Generation to Judgment: Opportunities and
  Challenges of LLM-as-a-Judge}.
\newblock
\newblock
\urldef\tempurl%
\url{https://doi.org/10.48550/arXiv.2411.16594}
\showDOI{\tempurl}
\showeprint[arXiv]{2411.16594}~[cs.AI]


\bibitem[Li et~al\mbox{.}(2023)]%
        {li2023towards}
\bibfield{author}{\bibinfo{person}{Zehan Li}, \bibinfo{person}{Xin Zhang},
  \bibinfo{person}{Yanzhao Zhang}, \bibinfo{person}{Dingkun Long},
  \bibinfo{person}{Pengjun Xie}, {and} \bibinfo{person}{Meishan Zhang}.}
  \bibinfo{year}{2023}\natexlab{}.
\newblock \bibinfo{title}{Towards General Text Embeddings with Multi-Stage
  Contrastive Learning}.
\newblock
\newblock
\urldef\tempurl%
\url{https://doi.org/10.48550/arXiv.2308.03281}
\showDOI{\tempurl}
\showeprint[arXiv]{2308.03281}~[cs.CL]


\bibitem[Lim et~al\mbox{.}(2025)]%
        {lim2025evaluatingimagehallucinationtexttoimage}
\bibfield{author}{\bibinfo{person}{Youngsun Lim}, \bibinfo{person}{Hojun Choi},
  {and} \bibinfo{person}{Hyunjung Shim}.} \bibinfo{year}{2025}\natexlab{}.
\newblock \bibinfo{title}{Evaluating Image Hallucination in Text-to-Image
  Generation with Question-Answering}.
\newblock
\newblock
\urldef\tempurl%
\url{https://doi.org/10.48550/arXiv.2409.12784}
\showDOI{\tempurl}
\showeprint[arXiv]{2409.12784}~[cs.CV]


\bibitem[Lin(2004)]%
        {lin2004rouge}
\bibfield{author}{\bibinfo{person}{Chin-Yew Lin}.}
  \bibinfo{year}{2004}\natexlab{}.
\newblock \showarticletitle{ROUGE: A Package for Automatic Evaluation of
  Summaries}. In \bibinfo{booktitle}{\emph{Proceedings of the Text
  Summarization Branches Out Workshop}} (Barcelona, Spain).
  \bibinfo{publisher}{Association for Computational Linguistics},
  \bibinfo{address}{Kerrville, TX, USA}, \bibinfo{pages}{74--81}.
\newblock
\urldef\tempurl%
\url{https://aclanthology.org/W04-1013/}
\showURL{%
\tempurl}


\bibitem[Liu et~al\mbox{.}(2023)]%
        {liu2023geval}
\bibfield{author}{\bibinfo{person}{Yang Liu}, \bibinfo{person}{Dan Iter},
  \bibinfo{person}{Yichong Xu}, \bibinfo{person}{Shuohang Wang},
  \bibinfo{person}{Ruochen Xu}, {and} \bibinfo{person}{Chenguang Zhu}.}
  \bibinfo{year}{2023}\natexlab{}.
\newblock \showarticletitle{G-Eval: NLG Evaluation Using GPT-4 with Better
  Human Alignment}. In \bibinfo{booktitle}{\emph{Proceedings of the 2023
  Conference on Empirical Methods in Natural Language Processing}} (Singapore)
  \emph{(\bibinfo{series}{EMNLP '23})},
  \bibfield{editor}{\bibinfo{person}{Houda Bouamor}, \bibinfo{person}{Juan
  Pino}, {and} \bibinfo{person}{Kalika Bali}} (Eds.).
  \bibinfo{publisher}{Association for Computational Linguistics},
  \bibinfo{address}{Kerrville, TX, USA}, \bibinfo{pages}{2511--2522}.
\newblock
\urldef\tempurl%
\url{https://doi.org/10.18653/v1/2023.emnlp-main.153}
\showDOI{\tempurl}


\bibitem[Liu et~al\mbox{.}(2019)]%
        {liu2019roberta}
\bibfield{author}{\bibinfo{person}{Yinhan Liu}, \bibinfo{person}{Myle Ott},
  \bibinfo{person}{Naman Goyal}, \bibinfo{person}{Jingfei Du},
  \bibinfo{person}{Mandar Joshi}, \bibinfo{person}{Danqi Chen},
  \bibinfo{person}{Omer Levy}, \bibinfo{person}{Mike Lewis},
  \bibinfo{person}{Luke Zettlemoyer}, {and} \bibinfo{person}{Veselin
  Stoyanov}.} \bibinfo{year}{2019}\natexlab{}.
\newblock \bibinfo{title}{RoBERTa: A Robustly Optimized BERT Pretraining
  Approach}.
\newblock
\newblock
\urldef\tempurl%
\url{https://doi.org/10.48550/arXiv.1907.11692}
\showDOI{\tempurl}
\showeprint[arXiv]{1907.11692}~[cs.CL]


\bibitem[Longo et~al\mbox{.}(2024)]%
        {Longo_2024}
\bibfield{author}{\bibinfo{person}{Luca Longo}, \bibinfo{person}{Mario Brcic},
  \bibinfo{person}{Federico Cabitza}, \bibinfo{person}{Jaesik Choi},
  \bibinfo{person}{Roberto Confalonieri}, \bibinfo{person}{Javier~Del Ser},
  \bibinfo{person}{Riccardo Guidotti}, \bibinfo{person}{Yoichi Hayashi},
  \bibinfo{person}{Francisco Herrera}, \bibinfo{person}{Andreas Holzinger},
  \bibinfo{person}{Richard Jiang}, \bibinfo{person}{Hassan Khosravi},
  \bibinfo{person}{Freddy Lecue}, \bibinfo{person}{Gianclaudio Malgieri},
  \bibinfo{person}{Andrés Páez}, \bibinfo{person}{Wojciech Samek},
  \bibinfo{person}{Johannes Schneider}, \bibinfo{person}{Timo Speith}, {and}
  \bibinfo{person}{Simone Stumpf}.} \bibinfo{year}{2024}\natexlab{}.
\newblock \showarticletitle{Explainable Artificial Intelligence (XAI) 2.0: A
  Manifesto of Open Challenges and Interdisciplinary Research Directions}.
\newblock \bibinfo{journal}{\emph{Information Fusion}}  \bibinfo{volume}{106},
  Article \bibinfo{articleno}{102301} (\bibinfo{date}{June}
  \bibinfo{year}{2024}), \bibinfo{numpages}{22}~pages.
\newblock
\showISSN{1566-2535}
\urldef\tempurl%
\url{https://doi.org/10.1016/j.inffus.2024.102301}
\showDOI{\tempurl}


\bibitem[Luo and Specia(2024)]%
        {luo2024from}
\bibfield{author}{\bibinfo{person}{Haoyan Luo} {and} \bibinfo{person}{Lucia
  Specia}.} \bibinfo{year}{2024}\natexlab{}.
\newblock \bibinfo{title}{From Understanding to Utilization: A Survey on
  Explainability for Large Language Models}.
\newblock
\newblock
\urldef\tempurl%
\url{https://doi.org/10.48550/arXiv.2401.12874}
\showDOI{\tempurl}
\showeprint[arXiv]{2401.12874}~[cs.CL]


\bibitem[Madsen et~al\mbox{.}(2024)]%
        {madsen2024are}
\bibfield{author}{\bibinfo{person}{Andreas Madsen}, \bibinfo{person}{Sarath
  Chandar}, {and} \bibinfo{person}{Siva Reddy}.}
  \bibinfo{year}{2024}\natexlab{}.
\newblock \showarticletitle{Are Self-Explanations from Large Language Models
  Faithful?}. In \bibinfo{booktitle}{\emph{Findings of the Association for
  Computational Linguistics: ACL 2024}} (Bangkok, Thailand),
  \bibfield{editor}{\bibinfo{person}{Lun-Wei Ku}, \bibinfo{person}{Andre
  Martins}, {and} \bibinfo{person}{Vivek Srikumar}} (Eds.).
  \bibinfo{publisher}{Association for Computational Linguistics},
  \bibinfo{address}{Kerrville, TX, USA}, \bibinfo{pages}{295--337}.
\newblock
\urldef\tempurl%
\url{https://doi.org/10.18653/v1/2024.findings-acl.19}
\showDOI{\tempurl}


\bibitem[Mahowald et~al\mbox{.}(2024)]%
        {mahowald2024dissociating}
\bibfield{author}{\bibinfo{person}{Kyle Mahowald}, \bibinfo{person}{Anna~A.
  Ivanova}, \bibinfo{person}{Idan~A. Blank}, \bibinfo{person}{Nancy Kanwisher},
  \bibinfo{person}{Joshua~B. Tenenbaum}, {and} \bibinfo{person}{Evelina
  Fedorenko}.} \bibinfo{year}{2024}\natexlab{}.
\newblock \showarticletitle{Dissociating Language and Thought in Large Language
  Models}.
\newblock \bibinfo{journal}{\emph{Trends in Cognitive Sciences}}
  \bibinfo{volume}{28}, \bibinfo{number}{6} (\bibinfo{date}{March}
  \bibinfo{year}{2024}), \bibinfo{pages}{517--540}.
\newblock
\urldef\tempurl%
\url{https://doi.org/10.1016/j.tics.2024.01.011}
\showDOI{\tempurl}


\bibitem[Manakul et~al\mbox{.}(2023)]%
        {manakul2023selfcheckgpt}
\bibfield{author}{\bibinfo{person}{Potsawee Manakul}, \bibinfo{person}{Adian
  Liusie}, {and} \bibinfo{person}{Mark Gales}.}
  \bibinfo{year}{2023}\natexlab{}.
\newblock \showarticletitle{SelfCheckGPT: Zero-Resource Black-Box Hallucination
  Detection for Generative Large Language Models}. In
  \bibinfo{booktitle}{\emph{Proceedings of the 2023 Conference on Empirical
  Methods in Natural Language Processing}} (Singapore)
  \emph{(\bibinfo{series}{EMNLP '23})},
  \bibfield{editor}{\bibinfo{person}{Houda Bouamor}, \bibinfo{person}{Juan
  Pino}, {and} \bibinfo{person}{Kalika Bali}} (Eds.).
  \bibinfo{publisher}{Association for Computational Linguistics},
  \bibinfo{address}{Kerrville, TX, USA}, \bibinfo{pages}{9004--9017}.
\newblock
\urldef\tempurl%
\url{https://doi.org/10.18653/v1/2023.emnlp-main.557}
\showDOI{\tempurl}


\bibitem[Meng et~al\mbox{.}(2024)]%
        {SFRAIResearch2024}
\bibfield{author}{\bibinfo{person}{Rui Meng}, \bibinfo{person}{Ye Liu},
  \bibinfo{person}{Shafiq~Rayhan Joty}, \bibinfo{person}{Caiming Xiong},
  \bibinfo{person}{Yingbo Zhou}, {and} \bibinfo{person}{Semih Yavuz}.}
  \bibinfo{year}{2024}\natexlab{}.
\newblock \bibinfo{title}{SFR-Embedding-Mistral: Enhance Text Retrieval with
  Transfer Learning}.
\newblock \bibinfo{howpublished}{Salesforce AI Research Blog}.
\newblock
\urldef\tempurl%
\url{https://www.salesforce.com/blog/sfr-embedding/}
\showURL{%
\tempurl}
\newblock
\shownote{Accessed: 2025-05-05}.


\bibitem[Minaee et~al\mbox{.}(2025)]%
        {minaee2024large}
\bibfield{author}{\bibinfo{person}{Shervin Minaee}, \bibinfo{person}{Tomas
  Mikolov}, \bibinfo{person}{Narjes Nikzad}, \bibinfo{person}{Meysam
  Chenaghlu}, \bibinfo{person}{Richard Socher}, \bibinfo{person}{Xavier
  Amatriain}, {and} \bibinfo{person}{Jianfeng Gao}.}
  \bibinfo{year}{2025}\natexlab{}.
\newblock \bibinfo{title}{Large Language Models: A Survey}.
\newblock
\newblock
\urldef\tempurl%
\url{https://doi.org/10.48550/arXiv.2402.06196}
\showDOI{\tempurl}
\showeprint[arXiv]{2402.06196}~[cs.CL]


\bibitem[Niu et~al\mbox{.}(2024)]%
        {niu2024ragtruth}
\bibfield{author}{\bibinfo{person}{Cheng Niu}, \bibinfo{person}{Yuanhao Wu},
  \bibinfo{person}{Juno Zhu}, \bibinfo{person}{Siliang Xu},
  \bibinfo{person}{KaShun Shum}, \bibinfo{person}{Randy Zhong},
  \bibinfo{person}{Juntong Song}, {and} \bibinfo{person}{Tong Zhang}.}
  \bibinfo{year}{2024}\natexlab{}.
\newblock \showarticletitle{RAGTruth: A Hallucination Corpus for Developing
  Trustworthy Retrieval-Augmented Language Models}. In
  \bibinfo{booktitle}{\emph{Proceedings of the 62nd Annual Meeting of the
  Association for Computational Linguistics (Volume 1: Long Papers)}} (Bangkok,
  Thailand) \emph{(\bibinfo{series}{ACL '24})},
  \bibfield{editor}{\bibinfo{person}{Lun-Wei Ku}, \bibinfo{person}{Andre
  Martins}, {and} \bibinfo{person}{Vivek Srikumar}} (Eds.).
  \bibinfo{publisher}{Association for Computational Linguistics},
  \bibinfo{address}{Kerrville, TX, USA}, \bibinfo{pages}{10862--10878}.
\newblock
\urldef\tempurl%
\url{https://doi.org/10.18653/v1/2024.acl-long.585}
\showDOI{\tempurl}


\bibitem[NovaSearch(2024a)]%
        {dunzhang2024stella1.5b}
\bibfield{author}{\bibinfo{person}{NovaSearch}.}
  \bibinfo{year}{2024}\natexlab{a}.
\newblock \bibinfo{title}{NovaSearch/stella\_en\_1.5B\_v5}.
\newblock \bibinfo{howpublished}{Hugging Face}.
\newblock
\urldef\tempurl%
\url{https://huggingface.co/NovaSearch/stella_en_1.5B_v5}
\showURL{%
\tempurl}
\newblock
\shownote{Accessed: 2025-05-05}.


\bibitem[NovaSearch(2024b)]%
        {dunzhang2024stella400}
\bibfield{author}{\bibinfo{person}{NovaSearch}.}
  \bibinfo{year}{2024}\natexlab{b}.
\newblock \bibinfo{title}{NovaSearch/stella\_en\_400M\_v5}.
\newblock \bibinfo{howpublished}{Hugging Face}.
\newblock
\urldef\tempurl%
\url{https://huggingface.co/NovaSearch/stella_en_400M_v5}
\showURL{%
\tempurl}
\newblock
\shownote{Accessed: 2025-05-05}.


\bibitem[Papineni et~al\mbox{.}(2002)]%
        {10.3115/1073083.1073135}
\bibfield{author}{\bibinfo{person}{Kishore Papineni}, \bibinfo{person}{Salim
  Roukos}, \bibinfo{person}{Todd Ward}, {and} \bibinfo{person}{Wei-Jing Zhu}.}
  \bibinfo{year}{2002}\natexlab{}.
\newblock \showarticletitle{Bleu: A Method for Automatic Evaluation of Machine
  Translation}. In \bibinfo{booktitle}{\emph{Proceedings of the 40th Annual
  Meeting on Association for Computational Linguistics}} (Philadelphia, PA,
  USA) \emph{(\bibinfo{series}{ACL '02})},
  \bibfield{editor}{\bibinfo{person}{Pierre Isabelle}, \bibinfo{person}{Eugene
  Charniak}, {and} \bibinfo{person}{Dekang Lin}} (Eds.).
  \bibinfo{publisher}{Association for Computational Linguistics},
  \bibinfo{address}{Kerrville, TX, USA}, \bibinfo{pages}{311--318}.
\newblock
\urldef\tempurl%
\url{https://doi.org/10.3115/1073083.1073135}
\showDOI{\tempurl}


\bibitem[Petroni et~al\mbox{.}(2019)]%
        {petroni2019language}
\bibfield{author}{\bibinfo{person}{Fabio Petroni}, \bibinfo{person}{Tim
  Rockt{\"a}schel}, \bibinfo{person}{Sebastian Riedel},
  \bibinfo{person}{Patrick Lewis}, \bibinfo{person}{Anton Bakhtin},
  \bibinfo{person}{Yuxiang Wu}, {and} \bibinfo{person}{Alexander Miller}.}
  \bibinfo{year}{2019}\natexlab{}.
\newblock \showarticletitle{Language Models as Knowledge Bases?}. In
  \bibinfo{booktitle}{\emph{Proceedings of the 2019 Conference on Empirical
  Methods in Natural Language Processing and the 9th International Joint
  Conference on Natural Language Processing}} (Hong Kong, China)
  \emph{(\bibinfo{series}{EMNLP-IJCNLP '19})},
  \bibfield{editor}{\bibinfo{person}{Kentaro Inui}, \bibinfo{person}{Jing
  Jiang}, \bibinfo{person}{Vincent Ng}, {and} \bibinfo{person}{Xiaojun Wan}}
  (Eds.). \bibinfo{publisher}{Association for Computational Linguistics},
  \bibinfo{address}{Kerrville, TX, USA}, \bibinfo{pages}{2463--2473}.
\newblock
\urldef\tempurl%
\url{https://doi.org/10.18653/v1/D19-1250}
\showDOI{\tempurl}


\bibitem[Quevedo et~al\mbox{.}(2024)]%
        {quevedo2024detecting}
\bibfield{author}{\bibinfo{person}{Ernesto Quevedo},
  \bibinfo{person}{Jorge~Yero Salazar}, \bibinfo{person}{Rachel Koerner},
  \bibinfo{person}{Pablo Rivas}, {and} \bibinfo{person}{Tomas Cerny}.}
  \bibinfo{year}{2024}\natexlab{}.
\newblock \showarticletitle{Detecting Hallucinations in Large Language Model
  Generation: A Token Probability Approach}. In
  \bibinfo{booktitle}{\emph{Proceedings of the 26th International Conference on
  Artificial Intelligence and Applications (ICAI '24)}} (Las Vegas, NV, USA)
  \emph{(\bibinfo{series}{Communications in Computer and Information Science
  (CCIS)}, Vol.~\bibinfo{volume}{2252})},
  \bibfield{editor}{\bibinfo{person}{Hamid~R. Arabnia},
  \bibinfo{person}{Leonidas Deligiannidis}, \bibinfo{person}{Soheyla Amirian},
  \bibinfo{person}{Farzan Shenavarmasouleh}, \bibinfo{person}{Farid
  Ghareh~Mohammadi}, {and} \bibinfo{person}{David de~la Fuente}} (Eds.).
  \bibinfo{publisher}{Springer Nature Switzerland}, \bibinfo{address}{Cham,
  Switzerland}, \bibinfo{pages}{154--173}.
\newblock
\showISBNx{978-3-031-86623-4}
\urldef\tempurl%
\url{https://doi.org/10.1007/978-3-031-86623-4_13}
\showDOI{\tempurl}


\bibitem[Radford et~al\mbox{.}(2021)]%
        {icml/RadfordKHRGASAM21}
\bibfield{author}{\bibinfo{person}{Alec Radford}, \bibinfo{person}{Jong~Wook
  Kim}, \bibinfo{person}{Chris Hallacy}, \bibinfo{person}{Aditya Ramesh},
  \bibinfo{person}{Gabriel Goh}, \bibinfo{person}{Sandhini Agarwal},
  \bibinfo{person}{Girish Sastry}, \bibinfo{person}{Amanda Askell},
  \bibinfo{person}{Pamela Mishkin}, \bibinfo{person}{Jack Clark},
  \bibinfo{person}{Gretchen Krueger}, {and} \bibinfo{person}{Ilya Sutskever}.}
  \bibinfo{year}{2021}\natexlab{}.
\newblock \showarticletitle{Learning Transferable Visual Models From Natural
  Language Supervision}. In \bibinfo{booktitle}{\emph{Proceedings of the 38th
  International Conference on Machine Learning (ICML '21)}} (Virtual Event)
  \emph{(\bibinfo{series}{Proceedings of Machine Learning Research},
  Vol.~\bibinfo{volume}{139})}, \bibfield{editor}{\bibinfo{person}{Marina
  Meila} {and} \bibinfo{person}{Tong Zhang}} (Eds.). \bibinfo{publisher}{PMLR},
  \bibinfo{address}{New York, NY, USA}, \bibinfo{pages}{8748--8763}.
\newblock
\urldef\tempurl%
\url{http://proceedings.mlr.press/v139/radford21a.html}
\showURL{%
\tempurl}


\bibitem[Radharapu et~al\mbox{.}(2023)]%
        {radharapu2023aart}
\bibfield{author}{\bibinfo{person}{Bhaktipriya Radharapu},
  \bibinfo{person}{Kevin Robinson}, \bibinfo{person}{Lora Aroyo}, {and}
  \bibinfo{person}{Preethi Lahoti}.} \bibinfo{year}{2023}\natexlab{}.
\newblock \showarticletitle{AART: AI-Assisted Red-Teaming with Diverse Data
  Generation for New LLM-Powered Applications}. In
  \bibinfo{booktitle}{\emph{Proceedings of the 2023 Conference on Empirical
  Methods in Natural Language Processing: Industry Track}} (Singapore)
  \emph{(\bibinfo{series}{EMNLP '23})},
  \bibfield{editor}{\bibinfo{person}{Mingxuan Wang} {and} \bibinfo{person}{Imed
  Zitouni}} (Eds.). \bibinfo{publisher}{Association for Computational
  Linguistics}, \bibinfo{address}{Kerrville, TX, USA},
  \bibinfo{pages}{380--395}.
\newblock
\urldef\tempurl%
\url{https://doi.org/10.18653/v1/2023.emnlp-industry.37}
\showDOI{\tempurl}


\bibitem[Rawte et~al\mbox{.}(2023)]%
        {rawte2023survey}
\bibfield{author}{\bibinfo{person}{Vipula Rawte}, \bibinfo{person}{Amit Sheth},
  {and} \bibinfo{person}{Amitava Das}.} \bibinfo{year}{2023}\natexlab{}.
\newblock \bibinfo{title}{A Survey of Hallucination in Large Foundation
  Models}.
\newblock
\newblock
\urldef\tempurl%
\url{https://doi.org/10.48550/arXiv.2309.05922}
\showDOI{\tempurl}
\showeprint[arXiv]{2309.05922}~[cs.AI]


\bibitem[Rei et~al\mbox{.}(2020)]%
        {rei2020comet}
\bibfield{author}{\bibinfo{person}{Ricardo Rei}, \bibinfo{person}{Craig
  Stewart}, \bibinfo{person}{Ana~C. Farinha}, {and} \bibinfo{person}{Alon
  Lavie}.} \bibinfo{year}{2020}\natexlab{}.
\newblock \showarticletitle{COMET: A Neural Framework for MT Evaluation}. In
  \bibinfo{booktitle}{\emph{Proceedings of the 2020 Conference on Empirical
  Methods in Natural Language Processing}} (Virtual Event)
  \emph{(\bibinfo{series}{EMNLP '20})},
  \bibfield{editor}{\bibinfo{person}{Bonnie Webber}, \bibinfo{person}{Trevor
  Cohn}, \bibinfo{person}{Yulan He}, {and} \bibinfo{person}{Yang Liu}} (Eds.).
  \bibinfo{publisher}{Association for Computational Linguistics},
  \bibinfo{address}{Kerrville, TX, USA}, \bibinfo{pages}{2685--2702}.
\newblock
\urldef\tempurl%
\url{https://doi.org/10.18653/v1/2020.emnlp-main.213}
\showDOI{\tempurl}


\bibitem[Reimers and Gurevych(2019)]%
        {reimers2019sentence}
\bibfield{author}{\bibinfo{person}{Nils Reimers} {and} \bibinfo{person}{Iryna
  Gurevych}.} \bibinfo{year}{2019}\natexlab{}.
\newblock \showarticletitle{Sentence-BERT: Sentence Embeddings using Siamese
  BERT-Networks}. In \bibinfo{booktitle}{\emph{Proceedings of the 2019
  Conference on Empirical Methods in Natural Language Processing and the 9th
  International Joint Conference on Natural Language Processing}} (Hong Kong,
  China) \emph{(\bibinfo{series}{EMNLP-IJCNLP '19})},
  \bibfield{editor}{\bibinfo{person}{Kentaro Inui}, \bibinfo{person}{Jing
  Jiang}, \bibinfo{person}{Vincent Ng}, {and} \bibinfo{person}{Xiaojun Wan}}
  (Eds.). \bibinfo{publisher}{Association for Computational Linguistics},
  \bibinfo{address}{Kerrville, TX, USA}, \bibinfo{pages}{3982--3992}.
\newblock
\urldef\tempurl%
\url{https://doi.org/10.18653/v1/D19-1410}
\showDOI{\tempurl}


\bibitem[Sai et~al\mbox{.}(2022)]%
        {sai2022survey}
\bibfield{author}{\bibinfo{person}{Ananya~B. Sai}, \bibinfo{person}{Akash~Kumar
  Mohankumar}, {and} \bibinfo{person}{Mitesh~M. Khapra}.}
  \bibinfo{year}{2022}\natexlab{}.
\newblock \showarticletitle{A Survey of Evaluation Metrics Used for NLG
  Systems}.
\newblock \bibinfo{journal}{\emph{ACM Comput. Surv.}} \bibinfo{volume}{55},
  \bibinfo{number}{2}, Article \bibinfo{articleno}{26} (\bibinfo{date}{Jan.}
  \bibinfo{year}{2022}), \bibinfo{numpages}{39}~pages.
\newblock
\showISSN{0360-0300}
\urldef\tempurl%
\url{https://doi.org/10.1145/3485766}
\showDOI{\tempurl}


\bibitem[Salimans et~al\mbox{.}(2016)]%
        {inceptionscore}
\bibfield{author}{\bibinfo{person}{Tim Salimans}, \bibinfo{person}{Ian
  Goodfellow}, \bibinfo{person}{Wojciech Zaremba}, \bibinfo{person}{Vicki
  Cheung}, \bibinfo{person}{Alec Radford}, {and} \bibinfo{person}{Xi Chen}.}
  \bibinfo{year}{2016}\natexlab{}.
\newblock \showarticletitle{Improved Techniques for Training GANs}. In
  \bibinfo{booktitle}{\emph{Proceedings of the 30th International Conference on
  Neural Information Processing Systems (NIPS '16)}} (Barcelona, Spain)
  \emph{(\bibinfo{series}{Advances in Neural Information Processing Systems},
  Vol.~\bibinfo{volume}{29})}, \bibfield{editor}{\bibinfo{person}{D.~Lee},
  \bibinfo{person}{M.~Sugiyama}, \bibinfo{person}{U.~Luxburg},
  \bibinfo{person}{I.~Guyon}, {and} \bibinfo{person}{R.~Garnett}} (Eds.).
  \bibinfo{publisher}{Curran Associates}, \bibinfo{address}{Red Hook, NY, USA},
  \bibinfo{pages}{2234--2242}.
\newblock
\showISBNx{9781510838819}
\urldef\tempurl%
\url{https://proceedings.neurips.cc/paper_files/paper/2016/hash/8a3363abe792db2d8761d6403605aeb7-Abstract.html}
\showURL{%
\tempurl}


\bibitem[Sellam et~al\mbox{.}(2020)]%
        {sellam2020bleurt}
\bibfield{author}{\bibinfo{person}{Thibault Sellam}, \bibinfo{person}{Dipanjan
  Das}, {and} \bibinfo{person}{Ankur Parikh}.} \bibinfo{year}{2020}\natexlab{}.
\newblock \showarticletitle{BLEURT: Learning Robust Metrics for Text
  Generation}. In \bibinfo{booktitle}{\emph{Proceedings of the 58th Annual
  Meeting of the Association for Computational Linguistics}} (Virtual Event)
  \emph{(\bibinfo{series}{ACL '20})}, \bibfield{editor}{\bibinfo{person}{Dan
  Jurafsky}, \bibinfo{person}{Joyce Chai}, \bibinfo{person}{Natalie Schluter},
  {and} \bibinfo{person}{Joel Tetreault}} (Eds.).
  \bibinfo{publisher}{Association for Computational Linguistics},
  \bibinfo{address}{Kerrville, TX, USA}, \bibinfo{pages}{7881--7892}.
\newblock
\urldef\tempurl%
\url{https://doi.org/10.18653/v1/2020.acl-main.704}
\showDOI{\tempurl}


\bibitem[Shanahan(2024)]%
        {shanahan2023talking}
\bibfield{author}{\bibinfo{person}{Murray Shanahan}.}
  \bibinfo{year}{2024}\natexlab{}.
\newblock \showarticletitle{Talking about Large Language Models}.
\newblock \bibinfo{journal}{\emph{Commun. ACM}} \bibinfo{volume}{67},
  \bibinfo{number}{2} (\bibinfo{date}{Jan.} \bibinfo{year}{2024}),
  \bibinfo{pages}{68--79}.
\newblock
\showISSN{0001-0782}
\urldef\tempurl%
\url{https://doi.org/10.1145/3624724}
\showDOI{\tempurl}


\bibitem[Sulem et~al\mbox{.}(2018)]%
        {sulem2018bleu}
\bibfield{author}{\bibinfo{person}{Elior Sulem}, \bibinfo{person}{Omri Abend},
  {and} \bibinfo{person}{Ari Rappoport}.} \bibinfo{year}{2018}\natexlab{}.
\newblock \showarticletitle{BLEU Is Not Suitable for the Evaluation of Text
  Simplification}. In \bibinfo{booktitle}{\emph{Proceedings of the 2018
  Conference on Empirical Methods in Natural Language Processing}} (Brussels,
  Belgium) \emph{(\bibinfo{series}{EMNLP '18})},
  \bibfield{editor}{\bibinfo{person}{Ellen Riloff}, \bibinfo{person}{David
  Chiang}, \bibinfo{person}{Julia Hockenmaier}, {and}
  \bibinfo{person}{Jun{'}ichi Tsujii}} (Eds.). \bibinfo{publisher}{Association
  for Computational Linguistics}, \bibinfo{address}{Kerrville, TX, USA},
  \bibinfo{pages}{738--744}.
\newblock
\urldef\tempurl%
\url{https://doi.org/10.18653/v1/D18-1081}
\showDOI{\tempurl}


\bibitem[Thompson and Post(2020)]%
        {thompson2020automatic}
\bibfield{author}{\bibinfo{person}{Brian Thompson} {and} \bibinfo{person}{Matt
  Post}.} \bibinfo{year}{2020}\natexlab{}.
\newblock \showarticletitle{Automatic Machine Translation Evaluation in Many
  Languages via Zero-Shot Paraphrasing}. In
  \bibinfo{booktitle}{\emph{Proceedings of the 2020 Conference on Empirical
  Methods in Natural Language Processing}} (Virtual Event)
  \emph{(\bibinfo{series}{EMNLP '20})},
  \bibfield{editor}{\bibinfo{person}{Bonnie Webber}, \bibinfo{person}{Trevor
  Cohn}, \bibinfo{person}{Yulan He}, {and} \bibinfo{person}{Yang Liu}} (Eds.).
  \bibinfo{publisher}{Association for Computational Linguistics},
  \bibinfo{address}{Kerrville, TX, USA}, \bibinfo{pages}{90--121}.
\newblock
\urldef\tempurl%
\url{https://doi.org/10.18653/v1/2020.emnlp-main.8}
\showDOI{\tempurl}


\bibitem[Tivnan et~al\mbox{.}(2024)]%
        {Tiv_Hallucination_MICCAI2024}
\bibfield{author}{\bibinfo{person}{Matthew Tivnan}, \bibinfo{person}{Siyeop
  Yoon}, \bibinfo{person}{Zhennong Chen}, \bibinfo{person}{Xiang Li},
  \bibinfo{person}{Dufan Wu}, {and} \bibinfo{person}{Quanzheng Li}.}
  \bibinfo{year}{2024}\natexlab{}.
\newblock \showarticletitle{Hallucination Index: An Image Quality Metric for
  Generative Reconstruction Models}. In \bibinfo{booktitle}{\emph{Proceedings
  of 27th International Conference on Medical Image Computing and Computer
  Assisted Intervention (MICCAI '24)}} (Marrakesh, Morocco)
  \emph{(\bibinfo{series}{Lecture Notes in Computer Science (LNCS)},
  Vol.~\bibinfo{volume}{15010})},
  \bibfield{editor}{\bibinfo{person}{Marius~George Linguraru},
  \bibinfo{person}{Qi~Dou}, \bibinfo{person}{Aasa Feragen},
  \bibinfo{person}{Stamatia Giannarou}, \bibinfo{person}{Ben Glocker},
  \bibinfo{person}{Karim Lekadir}, {and} \bibinfo{person}{Julia~A. Schnabel}}
  (Eds.). \bibinfo{publisher}{Springer Nature}, \bibinfo{address}{Cham,
  Switzerland}, \bibinfo{pages}{449--458}.
\newblock
\showISBNx{978-3-031-72117-5}
\urldef\tempurl%
\url{https://doi.org/10.1007/978-3-031-72117-5_42}
\showDOI{\tempurl}


\bibitem[Vale et~al\mbox{.}(2022)]%
        {vale2022explainable}
\bibfield{author}{\bibinfo{person}{Daniel Vale}, \bibinfo{person}{Ali
  El-Sharif}, {and} \bibinfo{person}{Muhammed Ali}.}
  \bibinfo{year}{2022}\natexlab{}.
\newblock \showarticletitle{Explainable Artificial Intelligence (XAI) Post-Hoc
  Explainability Methods: Risks and Limitations in Non-Discrimination Law}.
\newblock \bibinfo{journal}{\emph{AI and Ethics}} \bibinfo{volume}{2},
  \bibinfo{number}{4} (\bibinfo{date}{March} \bibinfo{year}{2022}),
  \bibinfo{pages}{815--826}.
\newblock
\showISSN{2730-5961}
\urldef\tempurl%
\url{https://doi.org/10.1007/s43681-022-00142-y}
\showDOI{\tempurl}


\bibitem[Wang et~al\mbox{.}(2024a)]%
        {wang2024usercentric}
\bibfield{author}{\bibinfo{person}{Jiayin Wang}, \bibinfo{person}{Fengran Mo},
  \bibinfo{person}{Weizhi Ma}, \bibinfo{person}{Peijie Sun},
  \bibinfo{person}{Min Zhang}, {and} \bibinfo{person}{Jian-Yun Nie}.}
  \bibinfo{year}{2024}\natexlab{a}.
\newblock \showarticletitle{A User-Centric Multi-Intent Benchmark for
  Evaluating Large Language Models}. In \bibinfo{booktitle}{\emph{Proceedings
  of the 2024 Conference on Empirical Methods in Natural Language Processing}}
  (Miami, FL, USA) \emph{(\bibinfo{series}{EMNLP '24})},
  \bibfield{editor}{\bibinfo{person}{Yaser Al-Onaizan}, \bibinfo{person}{Mohit
  Bansal}, {and} \bibinfo{person}{Yun-Nung Chen}} (Eds.).
  \bibinfo{publisher}{Association for Computational Linguistics},
  \bibinfo{address}{Kerrville, TX, USA}, \bibinfo{pages}{3588--3612}.
\newblock
\urldef\tempurl%
\url{https://doi.org/10.18653/v1/2024.emnlp-main.210}
\showDOI{\tempurl}


\bibitem[Wang et~al\mbox{.}(2024b)]%
        {wang2022text}
\bibfield{author}{\bibinfo{person}{Liang Wang}, \bibinfo{person}{Nan Yang},
  \bibinfo{person}{Xiaolong Huang}, \bibinfo{person}{Binxing Jiao},
  \bibinfo{person}{Linjun Yang}, \bibinfo{person}{Daxin Jiang},
  \bibinfo{person}{Rangan Majumder}, {and} \bibinfo{person}{Furu Wei}.}
  \bibinfo{year}{2024}\natexlab{b}.
\newblock \bibinfo{title}{Text Embeddings by Weakly-Supervised Contrastive
  Pre-Training}.
\newblock
\newblock
\urldef\tempurl%
\url{https://doi.org/10.48550/arXiv.2212.03533}
\showDOI{\tempurl}
\showeprint[arXiv]{2212.03533}~[cs.CL]


\bibitem[Wang et~al\mbox{.}(2024c)]%
        {wang2024improving}
\bibfield{author}{\bibinfo{person}{Liang Wang}, \bibinfo{person}{Nan Yang},
  \bibinfo{person}{Xiaolong Huang}, \bibinfo{person}{Linjun Yang},
  \bibinfo{person}{Rangan Majumder}, {and} \bibinfo{person}{Furu Wei}.}
  \bibinfo{year}{2024}\natexlab{c}.
\newblock \showarticletitle{Improving Text Embeddings with Large Language
  Models}. In \bibinfo{booktitle}{\emph{Proceedings of the 62nd Annual Meeting
  of the Association for Computational Linguistics (Volume 1: Long Papers)}}
  (Bangkok, Thailand) \emph{(\bibinfo{series}{ACL '24})},
  \bibfield{editor}{\bibinfo{person}{Lun-Wei Ku}, \bibinfo{person}{Andre
  Martins}, {and} \bibinfo{person}{Vivek Srikumar}} (Eds.).
  \bibinfo{publisher}{Association for Computational Linguistics},
  \bibinfo{address}{Kerrville, TX, USA}, \bibinfo{pages}{11897--11916}.
\newblock
\urldef\tempurl%
\url{https://doi.org/10.18653/v1/2024.acl-long.642}
\showDOI{\tempurl}


\bibitem[Wei et~al\mbox{.}(2022)]%
        {wei2022chain}
\bibfield{author}{\bibinfo{person}{Jason Wei}, \bibinfo{person}{Xuezhi Wang},
  \bibinfo{person}{Dale Schuurmans}, \bibinfo{person}{Maarten Bosma},
  \bibinfo{person}{Brian Ichter}, \bibinfo{person}{Fei Xia},
  \bibinfo{person}{Ed Chi}, \bibinfo{person}{Quoc~V. Le}, {and}
  \bibinfo{person}{Denny Zhou}.} \bibinfo{year}{2022}\natexlab{}.
\newblock \showarticletitle{Chain-of-Thought Prompting Elicits Reasoning in
  Large Language Models}. In \bibinfo{booktitle}{\emph{Proceedings of the
  Thirty-Sixth Annual Conference on Neural Information Processing Systems
  (NeurIPS '22)}} (New Orleans, LA, USA) \emph{(\bibinfo{series}{Advances in
  Neural Information Processing Systems}, Vol.~\bibinfo{volume}{35})},
  \bibfield{editor}{\bibinfo{person}{S.~Koyejo}, \bibinfo{person}{S.~Mohamed},
  \bibinfo{person}{A.~Agarwal}, \bibinfo{person}{D.~Belgrave},
  \bibinfo{person}{K.~Cho}, {and} \bibinfo{person}{A.~Oh}} (Eds.).
  \bibinfo{publisher}{Curran Associates}, \bibinfo{address}{Red Hook, NY, USA},
  \bibinfo{pages}{24824--24837}.
\newblock
\urldef\tempurl%
\url{https://proceedings.neurips.cc/paper_files/paper/2022/hash/9d5609613524ecf4f15af0f7b31abca4-Abstract-Conference.html}
\showURL{%
\tempurl}


\bibitem[Xi et~al\mbox{.}(2025)]%
        {xi2023rise}
\bibfield{author}{\bibinfo{person}{Zhiheng Xi}, \bibinfo{person}{Wenxiang
  Chen}, \bibinfo{person}{Xin Guo}, \bibinfo{person}{Wei He},
  \bibinfo{person}{Yiwen Ding}, \bibinfo{person}{Boyang Hong},
  \bibinfo{person}{Ming Zhang}, \bibinfo{person}{Junzhe Wang},
  \bibinfo{person}{Senjie Jin}, \bibinfo{person}{Enyu Zhou},
  \bibinfo{person}{Rui Zheng}, \bibinfo{person}{Xiaoran Fan},
  \bibinfo{person}{Xiao Wang}, \bibinfo{person}{Limao Xiong},
  \bibinfo{person}{Yuhao Zhou}, \bibinfo{person}{Weiran Wang},
  \bibinfo{person}{Changhao Jiang}, \bibinfo{person}{Yicheng Zou},
  \bibinfo{person}{Xiangyang Liu}, \bibinfo{person}{Zhangyue Yin},
  \bibinfo{person}{Shihan Dou}, \bibinfo{person}{Rongxiang Weng},
  \bibinfo{person}{Wenjuan Qin}, \bibinfo{person}{Yongyan Zheng},
  \bibinfo{person}{Xipeng Qiu}, \bibinfo{person}{Xuanjing Huang},
  \bibinfo{person}{Qi Zhang}, {and} \bibinfo{person}{Tao Gui}.}
  \bibinfo{year}{2025}\natexlab{}.
\newblock \showarticletitle{The Rise and Potential of Large Language Model
  Based Agents: A Survey}.
\newblock \bibinfo{journal}{\emph{Science China Information Sciences}}
  \bibinfo{volume}{68}, \bibinfo{number}{2}, Article
  \bibinfo{articleno}{121101} (\bibinfo{date}{Jan.} \bibinfo{year}{2025}).
\newblock
\showISSN{1869-1919}
\urldef\tempurl%
\url{https://doi.org/10.1007/s11432-024-4222-0}
\showDOI{\tempurl}


\bibitem[Xiao et~al\mbox{.}(2024)]%
        {xiao2024cpack}
\bibfield{author}{\bibinfo{person}{Shitao Xiao}, \bibinfo{person}{Zheng Liu},
  \bibinfo{person}{Peitian Zhang}, \bibinfo{person}{Niklas Muennighoff},
  \bibinfo{person}{Defu Lian}, {and} \bibinfo{person}{Jian-Yun Nie}.}
  \bibinfo{year}{2024}\natexlab{}.
\newblock \showarticletitle{C-Pack: Packed Resources For General Chinese
  Embeddings}. In \bibinfo{booktitle}{\emph{Proceedings of the 47th
  International ACM SIGIR Conference on Research and Development in Information
  Retrieval}} (Washington, DC, USA) \emph{(\bibinfo{series}{SIGIR '24})}.
  \bibinfo{publisher}{Association for Computing Machinery},
  \bibinfo{address}{New York, NY, USA}, \bibinfo{pages}{641--649}.
\newblock
\showISBNx{9798400704314}
\urldef\tempurl%
\url{https://doi.org/10.1145/3626772.3657878}
\showDOI{\tempurl}


\bibitem[Yao et~al\mbox{.}(2023)]%
        {yao2023tree}
\bibfield{author}{\bibinfo{person}{Shunyu Yao}, \bibinfo{person}{Dian Yu},
  \bibinfo{person}{Jeffrey Zhao}, \bibinfo{person}{Izhak Shafran},
  \bibinfo{person}{Tom Griffiths}, \bibinfo{person}{Yuan Cao}, {and}
  \bibinfo{person}{Karthik Narasimhan}.} \bibinfo{year}{2023}\natexlab{}.
\newblock \showarticletitle{Tree of Thoughts: Deliberate Problem Solving with
  Large Language Models}. In \bibinfo{booktitle}{\emph{Proceedings of the
  Thirty-Seventh Annual Conference on Neural Information Processing Systems
  (NeurIPS '23)}} (New Orleans, LA, USA) \emph{(\bibinfo{series}{Advances in
  Neural Information Processing Systems}, Vol.~\bibinfo{volume}{36})},
  \bibfield{editor}{\bibinfo{person}{A.~Oh}, \bibinfo{person}{T.~Naumann},
  \bibinfo{person}{A.~Globerson}, \bibinfo{person}{K.~Saenko},
  \bibinfo{person}{M.~Hardt}, {and} \bibinfo{person}{S.~Levine}} (Eds.).
  \bibinfo{publisher}{Curran Associates}, \bibinfo{address}{Red Hook, NY, USA},
  \bibinfo{pages}{11809--11822}.
\newblock
\urldef\tempurl%
\url{https://proceedings.neurips.cc/paper_files/paper/2023/hash/271db9922b8d1f4dd7aaef84ed5ac703-Abstract-Conference.html}
\showURL{%
\tempurl}


\bibitem[Yuan et~al\mbox{.}(2021)]%
        {yuan2021bartscore}
\bibfield{author}{\bibinfo{person}{Weizhe Yuan}, \bibinfo{person}{Graham
  Neubig}, {and} \bibinfo{person}{Pengfei Liu}.}
  \bibinfo{year}{2021}\natexlab{}.
\newblock \showarticletitle{BARTScore: Evaluating Generated Text as Text
  Generation}. In \bibinfo{booktitle}{\emph{Proceedings of the Thirty-Fifth
  Annual Conference on Neural Information Processing Systems (NeurIPS '21)}}
  (Virtual Event) \emph{(\bibinfo{series}{Advances in Neural Information
  Processing Systems}, Vol.~\bibinfo{volume}{34})},
  \bibfield{editor}{\bibinfo{person}{M.~Ranzato},
  \bibinfo{person}{A.~Beygelzimer}, \bibinfo{person}{Y.~Dauphin},
  \bibinfo{person}{P.S. Liang}, {and} \bibinfo{person}{J.~Wortman Vaughan}}
  (Eds.). \bibinfo{publisher}{Curran Associates}, \bibinfo{address}{Red Hook,
  NY, USA}, \bibinfo{pages}{27263--27277}.
\newblock
\urldef\tempurl%
\url{https://proceedings.neurips.cc/paper_files/paper/2021/hash/e4d2b6e6fdeca3e60e0f1a62fee3d9dd-Abstract.html}
\showURL{%
\tempurl}


\bibitem[Zhang et~al\mbox{.}(2025a)]%
        {zhang2024jasper}
\bibfield{author}{\bibinfo{person}{Dun Zhang}, \bibinfo{person}{Jiacheng Li},
  \bibinfo{person}{Ziyang Zeng}, {and} \bibinfo{person}{Fulong Wang}.}
  \bibinfo{year}{2025}\natexlab{a}.
\newblock \bibinfo{title}{Jasper and Stella: Distillation of SOTA Embedding
  Models}.
\newblock
\newblock
\urldef\tempurl%
\url{https://doi.org/10.48550/arXiv.2412.19048}
\showDOI{\tempurl}
\showeprint[arXiv]{2412.19048}~[cs.IR]


\bibitem[Zhang et~al\mbox{.}(2025b)]%
        {zhang2024knowhalu}
\bibfield{author}{\bibinfo{person}{Jiawei Zhang}, \bibinfo{person}{Chejian Xu},
  \bibinfo{person}{Yu Gai}, \bibinfo{person}{Freddy Lecue},
  \bibinfo{person}{Shuang Yang}, \bibinfo{person}{Dawn Song}, {and}
  \bibinfo{person}{Bo Li}.} \bibinfo{year}{2025}\natexlab{b}.
\newblock \showarticletitle{KnowHalu: Hallucination Detection via Multi-Form
  Knowledge Based Factual Checking}. In \bibinfo{booktitle}{\emph{Proceedings
  of the Workshop on Foundation Models in the Wild}} (Singapore)
  \emph{(\bibinfo{series}{FM-Wild '25})}. \bibinfo{publisher}{OpenReview},
  \bibinfo{address}{Amherst, MA, USA}, \bibinfo{numpages}{30}~pages.
\newblock
\urldef\tempurl%
\url{https://openreview.net/forum?id=RFwyhpcYZK}
\showURL{%
\tempurl}


\bibitem[Zhang et~al\mbox{.}(2020)]%
        {zhang2020bertscore}
\bibfield{author}{\bibinfo{person}{Tianyi Zhang}, \bibinfo{person}{Varsha
  Kishore}, \bibinfo{person}{Felix Wu}, \bibinfo{person}{Kilian~Q. Weinberger},
  {and} \bibinfo{person}{Yoav Artzi}.} \bibinfo{year}{2020}\natexlab{}.
\newblock \showarticletitle{BERTScore: Evaluating Text Generation with BERT}.
  In \bibinfo{booktitle}{\emph{Proceedings of the Eighth International
  Conference on Learning Representations}} (Virtual Event)
  \emph{(\bibinfo{series}{ICLR '20})}. \bibinfo{publisher}{OpenReview},
  \bibinfo{address}{Amherst, MA, USA}, \bibinfo{numpages}{43}~pages.
\newblock
\urldef\tempurl%
\url{https://openreview.net/forum?id=SkeHuCVFDr}
\showURL{%
\tempurl}


\bibitem[Zhang et~al\mbox{.}(2023)]%
        {zhang2023sirens}
\bibfield{author}{\bibinfo{person}{Yue Zhang}, \bibinfo{person}{Yafu Li},
  \bibinfo{person}{Leyang Cui}, \bibinfo{person}{Deng Cai},
  \bibinfo{person}{Lemao Liu}, \bibinfo{person}{Tingchen Fu},
  \bibinfo{person}{Xinting Huang}, \bibinfo{person}{Enbo Zhao},
  \bibinfo{person}{Yu Zhang}, \bibinfo{person}{Yulong Chen},
  \bibinfo{person}{Longyue Wang}, \bibinfo{person}{Anh~Tuan Luu},
  \bibinfo{person}{Wei Bi}, \bibinfo{person}{Freda Shi}, {and}
  \bibinfo{person}{Shuming Shi}.} \bibinfo{year}{2023}\natexlab{}.
\newblock \bibinfo{title}{Siren's Song in the AI Ocean: A Survey on
  Hallucination in Large Language Models}.
\newblock
\newblock
\urldef\tempurl%
\url{https://doi.org/10.48550/arXiv.2309.01219}
\showDOI{\tempurl}
\showeprint[arXiv]{2309.01219}~[cs.CL]


\bibitem[Zhao et~al\mbox{.}(2024)]%
        {zhao2024explainability}
\bibfield{author}{\bibinfo{person}{Haiyan Zhao}, \bibinfo{person}{Hanjie Chen},
  \bibinfo{person}{Fan Yang}, \bibinfo{person}{Ninghao Liu},
  \bibinfo{person}{Huiqi Deng}, \bibinfo{person}{Hengyi Cai},
  \bibinfo{person}{Shuaiqiang Wang}, \bibinfo{person}{Dawei Yin}, {and}
  \bibinfo{person}{Mengnan Du}.} \bibinfo{year}{2024}\natexlab{}.
\newblock \showarticletitle{Explainability for Large Language Models: A
  Survey}.
\newblock \bibinfo{journal}{\emph{ACM Trans. Intell. Syst. Technol.}}
  \bibinfo{volume}{15}, \bibinfo{number}{2}, Article \bibinfo{articleno}{20}
  (\bibinfo{date}{Feb.} \bibinfo{year}{2024}), \bibinfo{numpages}{38}~pages.
\newblock
\showISSN{2157-6904}
\urldef\tempurl%
\url{https://doi.org/10.1145/3639372}
\showDOI{\tempurl}


\bibitem[Zhao et~al\mbox{.}(2025)]%
        {zhao2023survey}
\bibfield{author}{\bibinfo{person}{Wayne~Xin Zhao}, \bibinfo{person}{Kun Zhou},
  \bibinfo{person}{Junyi Li}, \bibinfo{person}{Tianyi Tang},
  \bibinfo{person}{Xiaolei Wang}, \bibinfo{person}{Yupeng Hou},
  \bibinfo{person}{Yingqian Min}, \bibinfo{person}{Beichen Zhang},
  \bibinfo{person}{Junjie Zhang}, \bibinfo{person}{Zican Dong},
  \bibinfo{person}{Yifan Du}, \bibinfo{person}{Chen Yang},
  \bibinfo{person}{Yushuo Chen}, \bibinfo{person}{Zhipeng Chen},
  \bibinfo{person}{Jinhao Jiang}, \bibinfo{person}{Ruiyang Ren},
  \bibinfo{person}{Yifan Li}, \bibinfo{person}{Xinyu Tang},
  \bibinfo{person}{Zikang Liu}, \bibinfo{person}{Peiyu Liu},
  \bibinfo{person}{Jian-Yun Nie}, {and} \bibinfo{person}{Ji-Rong Wen}.}
  \bibinfo{year}{2025}\natexlab{}.
\newblock \bibinfo{title}{A Survey of Large Language Models}.
\newblock
\newblock
\urldef\tempurl%
\url{https://doi.org/10.48550/arXiv.2303.18223}
\showDOI{\tempurl}
\showeprint[arXiv]{2303.18223}~[cs.CL]


\bibitem[Zheng et~al\mbox{.}(2023)]%
        {zheng2023judging}
\bibfield{author}{\bibinfo{person}{Lianmin Zheng}, \bibinfo{person}{Wei-Lin
  Chiang}, \bibinfo{person}{Ying Sheng}, \bibinfo{person}{Siyuan Zhuang},
  \bibinfo{person}{Zhanghao Wu}, \bibinfo{person}{Yonghao Zhuang},
  \bibinfo{person}{Zi Lin}, \bibinfo{person}{Zhuohan Li},
  \bibinfo{person}{Dacheng Li}, \bibinfo{person}{Eric Xing},
  \bibinfo{person}{Hao Zhang}, \bibinfo{person}{Joseph~E. Gonzalez}, {and}
  \bibinfo{person}{Ion Stoica}.} \bibinfo{year}{2023}\natexlab{}.
\newblock \showarticletitle{Judging LLM-as-a-Judge with MT-Bench and Chatbot
  Arena}. In \bibinfo{booktitle}{\emph{Proceedings of the Thirty-Seventh Annual
  Conference on Neural Information Processing Systems (NeurIPS '23)}} (New
  Orleans, LA, USA) \emph{(\bibinfo{series}{Advances in Neural Information
  Processing Systems}, Vol.~\bibinfo{volume}{36})},
  \bibfield{editor}{\bibinfo{person}{A.~Oh}, \bibinfo{person}{T.~Naumann},
  \bibinfo{person}{A.~Globerson}, \bibinfo{person}{K.~Saenko},
  \bibinfo{person}{M.~Hardt}, {and} \bibinfo{person}{S.~Levine}} (Eds.).
  \bibinfo{publisher}{Curran Associates}, \bibinfo{address}{Red Hook, NY, USA},
  \bibinfo{pages}{46595--46623}.
\newblock
\urldef\tempurl%
\url{https://proceedings.neurips.cc/paper_files/paper/2023/hash/91f18a1287b398d378ef22505bf41832-Abstract-Datasets_and_Benchmarks.html}
\showURL{%
\tempurl}


\bibitem[Zhong et~al\mbox{.}(2022)]%
        {zhong2022towards}
\bibfield{author}{\bibinfo{person}{Ming Zhong}, \bibinfo{person}{Yang Liu},
  \bibinfo{person}{Da Yin}, \bibinfo{person}{Yuning Mao},
  \bibinfo{person}{Yizhu Jiao}, \bibinfo{person}{Pengfei Liu},
  \bibinfo{person}{Chenguang Zhu}, \bibinfo{person}{Heng Ji}, {and}
  \bibinfo{person}{Jiawei Han}.} \bibinfo{year}{2022}\natexlab{}.
\newblock \showarticletitle{Towards a Unified Multi-Dimensional Evaluator for
  Text Generation}. In \bibinfo{booktitle}{\emph{Proceedings of the 2022
  Conference on Empirical Methods in Natural Language Processing}} (Abu Dhabi,
  United Arab Emirates) \emph{(\bibinfo{series}{EMNLP '22})},
  \bibfield{editor}{\bibinfo{person}{Yoav Goldberg}, \bibinfo{person}{Zornitsa
  Kozareva}, {and} \bibinfo{person}{Yue Zhang}} (Eds.).
  \bibinfo{publisher}{Association for Computational Linguistics},
  \bibinfo{address}{Kerrville, TX, USA}, \bibinfo{pages}{2023--2038}.
\newblock
\urldef\tempurl%
\url{https://doi.org/10.18653/v1/2022.emnlp-main.131}
\showDOI{\tempurl}


\bibitem[Zhu et~al\mbox{.}(2024)]%
        {zhu2024large}
\bibfield{author}{\bibinfo{person}{Yutao Zhu}, \bibinfo{person}{Huaying Yuan},
  \bibinfo{person}{Shuting Wang}, \bibinfo{person}{Jiongnan Liu},
  \bibinfo{person}{Wenhan Liu}, \bibinfo{person}{Chenlong Deng},
  \bibinfo{person}{Haonan Chen}, \bibinfo{person}{Zheng Liu},
  \bibinfo{person}{Zhicheng Dou}, {and} \bibinfo{person}{Ji-Rong Wen}.}
  \bibinfo{year}{2024}\natexlab{}.
\newblock \bibinfo{title}{Large Language Models for Information Retrieval: A
  Survey}.
\newblock
\newblock
\urldef\tempurl%
\url{https://doi.org/10.48550/arXiv.2308.07107}
\showDOI{\tempurl}
\showeprint[arXiv]{2308.07107}~[cs.CL]


\bibitem[Zhuang et~al\mbox{.}(2024)]%
        {OpenAItextEmbeddingLarge}
\bibfield{author}{\bibinfo{person}{Juntang Zhuang}, \bibinfo{person}{Paul
  Baltescu}, \bibinfo{person}{Joy Jiao}, \bibinfo{person}{Arvind Neelakantan},
  \bibinfo{person}{Andrew Braunstein}, \bibinfo{person}{Jeff Harris},
  \bibinfo{person}{Logan Kilpatrick}, \bibinfo{person}{Leher Pathak},
  \bibinfo{person}{Enoch Cheung}, \bibinfo{person}{Ted Sanders},
  \bibinfo{person}{Yutian Liu}, \bibinfo{person}{Anushree Agrawal},
  \bibinfo{person}{Andrew Peng}, \bibinfo{person}{Ian Kivlichan},
  \bibinfo{person}{Mehmet Yatbaz}, \bibinfo{person}{Madelaine Boyd},
  \bibinfo{person}{Anna-Luisa Brakman}, \bibinfo{person}{Florencia~Leoni
  Aleman}, \bibinfo{person}{Henry Head}, \bibinfo{person}{Molly Lin},
  \bibinfo{person}{Meghan Shah}, \bibinfo{person}{Chelsea Carlson},
  \bibinfo{person}{Sam Toizer}, \bibinfo{person}{Ryan Greene},
  \bibinfo{person}{Alison Harmon}, \bibinfo{person}{Denny Jin},
  \bibinfo{person}{Karolis Kosas}, \bibinfo{person}{Marie Inuzuka},
  \bibinfo{person}{Peter Bakkum}, \bibinfo{person}{Barret Zoph},
  \bibinfo{person}{Luke Metz}, \bibinfo{person}{Jiayi Weng},
  \bibinfo{person}{Randall Lin}, \bibinfo{person}{Yash Patil},
  \bibinfo{person}{Mianna Chen}, \bibinfo{person}{Andrew Kondrich},
  \bibinfo{person}{Brydon Eastman}, \bibinfo{person}{Liam Fedus},
  \bibinfo{person}{John Schulman}, \bibinfo{person}{Vlad Fomenko},
  \bibinfo{person}{Andrej Karpathy}, \bibinfo{person}{Aidan Clark}, {and}
  \bibinfo{person}{Owen Campbell-Moore}.} \bibinfo{year}{2024}\natexlab{}.
\newblock \bibinfo{title}{OpenAI Text-Embedding-Large: New Embedding Models and
  API Updates}.
\newblock \bibinfo{howpublished}{OpenAI Research}.
\newblock
\urldef\tempurl%
\url{https://openai.com/index/new-embedding-models-and-api-updates/}
\showURL{%
\tempurl}
\newblock
\shownote{Accessed: 2025-05-02}.


\end{thebibliography}

\ifconf
\newpage
\section*{NeurIPS Paper Checklist}

\begin{enumerate}

\item {\bf Claims}
    \item[] Question: Do the main claims made in the abstract and introduction accurately reflect the paper's contributions and scope?
    \item[] Answer: \answerYes{} % Replace by \answerYes{}, \answerNo{}, or \answerNA{}.
    \item[] Justification: We describe the design and pipeline of \nameS in Section~\ref{sec:design} and provide a scalability analysis in Section~\ref{sec:scalability}. The datasets and metrics are described in Section~\ref{sec:eval}, which also contains the empirical results.

\item {\bf Limitations}
    \item[] Question: Does the paper discuss the limitations of the work performed by the authors?
    \item[] Answer: \answerYes{} % Replace by \answerYes{}, \answerNo{}, or \answerNA{}.
    \item[] Justification: We discuss the limitations of our work in Section~\ref{sec:rw} as well as in Sections~\ref{sec:eval-distinguish} and \ref{sec:eval-heatmaps}, .

\item {\bf Theory assumptions and proofs}
    \item[] Question: For each theoretical result, does the paper provide the full set of assumptions and a complete (and correct) proof?
    \item[] Answer: \answerNA{} % Replace by \answerYes{}, \answerNo{}, or \answerNA{}.
    \item[] Justification: Our work suggests using embeddings as way to verify LLM replies and we show the merits of this approach based on empirical evaluation. However we provide a complexity analysis in Section~\ref{sec:scalability}.

    \item {\bf Experimental result reproducibility}
    \item[] Question: Does the paper fully disclose all the information needed to reproduce the main experimental results of the paper to the extent that it affects the main claims and/or conclusions of the paper (regardless of whether the code and data are provided or not)?
    \item[] Answer: \answerYes{} % Replace by \answerYes{}, \answerNo{}, or \answerNA{}.
    \item[] Justification: We describe the architecture in Section~\ref{sec:design}. We will release the code as well as the datasets publicly and submit it as well as part of the supplementary material for the reviewers. We will include a description of the necessary steps to recreate the results from the evaluation section.

\item {\bf Open access to data and code}
    \item[] Question: Does the paper provide open access to the data and code, with sufficient instructions to faithfully reproduce the main experimental results, as described in supplemental material?
    \item[] Answer: \answerYes{} % Replace by \answerYes{}, \answerNo{}, or \answerNA{}.
    \item[] Justification: We will publish the code and the datasets. Included in the supplemental material will be instructions to reproduce the evaluation results.

\item {\bf Experimental setting/details}
    \item[] Question: Does the paper specify all the training and test details (e.g., data splits, hyperparameters, how they were chosen, type of optimizer, etc.) necessary to understand the results?
    \item[] Answer: \answerYes{} % Replace by \answerYes{}, \answerNo{}, or \answerNA{}.
    \item[] Justification: We provide the parameters for the empirical evaluation in Section~\ref{sec:eval} as well as the evaluation figures.

\item {\bf Experiment statistical significance}
    \item[] Question: Does the paper report error bars suitably and correctly defined or other appropriate information about the statistical significance of the experiments?
    \item[] Answer: \answerYes{} % Replace by \answerYes{}, \answerNo{}, or \answerNA{}.
    \item[] Justification: We show the distribution of the empirical data with violin plots, which covers all the measured data without cherry picking.

\item {\bf Experiments compute resources}
    \item[] Question: For each experiment, does the paper provide sufficient information on the computer resources (type of compute workers, memory, time of execution) needed to reproduce the experiments?
    \item[] Answer: \answerYes{} % Replace by \answerYes{}, \answerNo{}, or \answerNA{}.
    \item[] Justification: We provide information on the compute resources in Appendix~\ref{sec:compute_resources}.

\item {\bf Code of ethics}
    \item[] Question: Does the research conducted in the paper conform, in every respect, with the NeurIPS Code of Ethics \url{https://neurips.cc/public/EthicsGuidelines}?
    \item[] Answer: \answerYes{} % Replace by \answerYes{}, \answerNo{}, or \answerNA{}.
    \item[] Justification: The datasets do not include personal information.

\item {\bf Broader impacts}
    \item[] Question: Does the paper discuss both potential positive societal impacts and negative societal impacts of the work performed?
    \item[] Answer: \answerNA{} % Replace by \answerYes{}, \answerNo{}, or \answerNA{}.
    \item[] Justification: Our idea tries to improve the trust in the LLM replies and therefore all positive and negative aspects of LLMs apply to our idea as well.

\item {\bf Safeguards}
    \item[] Question: Does the paper describe safeguards that have been put in place for responsible release of data or models that have a high risk for misuse (e.g., pretrained language models, image generators, or scraped datasets)?
    \item[] Answer: \answerNA{} % Replace by \answerYes{}, \answerNo{}, or \answerNA{}.
    \item[] Justification: The framework tries to clarify, whether a given LLM reply can be trusted, no additional safeguards to the ones, that the LLM should have already in place, are necessary.

\item {\bf Licenses for existing assets}
    \item[] Question: Are the creators or original owners of assets (e.g., code, data, models), used in the paper, properly credited and are the license and terms of use explicitly mentioned and properly respected?
    \item[] Answer: \answerYes{} % Replace by \answerYes{}, \answerNo{}, or \answerNA{}.
    \item[] Justification: We provide citations for all used embedding models and tools.

\item {\bf New assets}
    \item[] Question: Are new assets introduced in the paper well documented and is the documentation provided alongside the assets?
    \item[] Answer: \answerYes{} % Replace by \answerYes{}, \answerNo{}, or \answerNA{}.
    \item[] Justification: We document our code and the necessary steps to execute our evaluation pipeline.

\item {\bf Crowdsourcing and research with human subjects}
    \item[] Question: For crowdsourcing experiments and research with human subjects, does the paper include the full text of instructions given to participants and screenshots, if applicable, as well as details about compensation (if any)?
    \item[] Answer: \answerNA{} % Replace by \answerYes{}, \answerNo{}, or \answerNA{}.
    \item[] Justification: No human subjects were used in the experiments.

\item {\bf Institutional review board (IRB) approvals or equivalent for research with human subjects}
    \item[] Question: Does the paper describe potential risks incurred by study participants, whether such risks were disclosed to the subjects, and whether Institutional Review Board (IRB) approvals (or an equivalent approval/review based on the requirements of your country or institution) were obtained?
    \item[] Answer: \answerNA{} % Replace by \answerYes{}, \answerNo{}, or \answerNA{}.
    \item[] Justification: No human subjects were used in the experiments.

\item {\bf Declaration of LLM usage}
    \item[] Question: Does the paper describe the usage of LLMs if it is an important, original, or non-standard component of the core methods in this research? Note that if the LLM is used only for writing, editing, or formatting purposes and does not impact the core methodology, scientific rigorousness, or originality of the research, declaration is not required.
    %this research?
    \item[] Answer: \answerYes{} % Replace by \answerYes{}, \answerNo{}, or \answerNA{}.
    \item[] Justification: The primary target of \nameS is to verify answers of language models, for which we use various embedding models. We discuss the architecture of the framework in Section~\ref{sec:design}.

\end{enumerate}

%%%%%%%%%%%%%%%%%%%%%%%%%%%%%%%%%%%%%%%%%%%%%%%%%%%%%%%%%%%%
\fi

\newpage
\appendix

\section{Details on Complexity Analysis}
\label{sec:app:complexity}

We estimate the complexity of verifying an open-ended task.
Furthermore, we distinguish additionally the fundamental part that appears in many approaches - computing similarity of two text passages, consisting of $s$ sentences with each one having on average $t$ tokens.

\subsection{Computing Similarity of Two Passages}

\textbf{BARTScore}
This method uses a pretrained text model to evaluate generated text.
The text comparison requires obtaining and summing the probability of each token.
Thus, to compare two multi-sentence passages, $O(st)$ work is needed, which can be parallelized with an additional reduction step to depth $O(\log(st))$.

\textbf{UniEval}
This method uses a single inference operation for the entire evaluation, and thus has no separate step for text passage comparison.

\textbf{BERTScore}
BERTScore is defined as a method for two sentences, not text passages.
This method has four major steps: create the two embeddings for the inputs, compute pairwise cosine similarity between vectors, find the maximum similarity, and perform final importance weighting.
The first step requires two inference calls, one for the reference and one for the candidate, resulting in depth $D_M$ and work $2*W_M$.
In the second step, we compute $O(t^2)$ similarity computations. Since the embedding dimensionality is $d$, and we assume $O(\log(d))$ cost of computing the score of two embeddings.
Each pair of embeddings can be evaluated independently.
Thus, the depth is $O(\log(d))$, and work is $O(t^2 \log(d))$.
Finding the maximum can be done in $O(t^2)$ steps and $O(\log(t))$ time.
Importance weighting uses pre-computed \emph{idf} values, which adds only $O(t)$ arithmetical operations.

Thus, the total depth is $D_M + O(\log(td))$, with work $2W_M + O(t^2 \log(d) + t^2)$.

\textbf{SentenceBERT}
Similarly to BERTScore, this method is defined at the level of individual sentences.
The method applies BERT inference for each sentence and pooling to compute embeddings, and then computes cosine similarity between two embeddings.
The total depth is thus $D_I + O(\log(td))$, and work is $2W_I + O(t + d)$.

\textbf{SelfCheckGPT}

In the \textbf{BERTScore} variant, we compute the maximum BERTScore between a given sentence of the reply and any sentence from each sample, and then compute the average result for all samples.
Thus, we perform the BERTScore operation $s$ times, and we can parallelize all operations except for finding the maximum, 
which adds $O(\log(s))$ to the depth.
Thus, based on the complexity of a single BERTScore application, we obtain the work $s2W_M + O(t^2 \log(d) + t^2)$ and depth $D_M + O(\log(tds))$.

In the \textbf{NLI} variant, SelfCheck uses a fine-tuned NLI model to compute the average probability of contradiction for each sample.
Each sentence of the document is evaluated separately, resulting in $s$ inference operations, and $O(s)$ arithmetical operations of computing the total summary.
Thus, the depth is $D_I + O(\log(s))$ and work is $sW_I + O(s)$.

%This means $O((st)^2) = O(s^2 t^2)$ embedding comparisons have to be performed for any two passages (for each pair of compared sentences, one compares every pair of individual tokens/words), resulting in a total of $O(k^2 s^2 t^2)$ embedding comparisons as this is done for $O(k^2)$ pairs of LLM answers, and a total of $O(k^2)$ embedding constructions.
%
%\lorenzo{Embeddings gets created on each sample (LLm answer) sent to BertScore, we send 2 samples each request, we have $k^2$ comparison to make, so $k^2$ calls to Bert, in total $2k^2$ embedding to be made}

\textbf{HaloCheck}
The method requires using $k$ different LLM responses, and then computing the pairwise sentence entanglement between them.
To compute the similarity score, HaloCheck employs the $SummaC$ method~\cite{laban2022summac}, in the zero-shot and sentence-level variant ($SummaC_{zs}$).
This results in one bipartite graph (a matrix) for all responses, which is later simplified to a similarity score for the entire document.
Each document is partitioned into sub-blocks, and an inference with an NLI model is applied pairwise to all sub-blocks of the input and summary document.
The zero-shot approach simplifies the matrix to a vector by computing the maximum of each column, and then a mean is computed to produce a final scalar value.
We assume that all documents are split into sentences, resulting in $s^2$ inference calls.
Since the matrix summary can be computed in parallel with additional reduction steps, the depth is $D_I + O(\log(s))$,
and the work is $s^2 * W_I + O(s^2)$.

%
%\lorenzo{Embeddings gets created on each sample (LLm answer) sent to BertScore, we send 2 samples each request, we have $k^2$ comparison to make, so $k^2$ calls to Bert, in total $2k^2$ embedding to be made}

\textbf{G-Eval}
This method creates a Chain-of-Thought for evaluation with LLM, and does not have a separate step for computing similarity of text passages.

\subsection{Verifying an Open-Ended Task Answer}

%{SelfCheckGPT} assesses a given LLM reply by comparing it to all sample replies collected. To simplify the following derivations, assume that in an individual comparison of two LLM replies, these replies consist of $s_1$ and $s_2$ sentences, respectively. Now, for each such comparison,
%
% it separates into sentences the two LLm answers obtaining $s_1$ and $s_2$ sentences, than it sends two arrays of sentences to bertScore containing each $s_1 s_2$ sentences (The sentences gets repeated respectevly $s_2$ and $s_1$ times. Generalizing it we end up sending $2s^2$ sentences or $O(s^2)$}, 
%
%SelfCheckGPT uses BERTScore, where the two input passages $x$ and $y$ to BERTScore consist of $s_1 s_2$ sentences each, i.e., both passage~$x$ and passage $y$ contain all the sentences from its corresponding LLM reply, repeated as many times as the number of sentences in the other LLM reply (this is conducted to enable comparing all sentences from each reply pairwise). This gives (using the above BERTScore formulae) $O(k s^2)$ embedding constructions (there are $k$ LLM replies) and $O(k s^2 s^2 t^2) = O(k s^4 t^2)$ embedding comparisons. 

\textbf{BARTScore}
For the full task evaluation, a single inference call is needed to obtain token probabilities that are later aggregated, as described in the previous section.

\textbf{UniEval}
This method uses a single LLM inference call for the task.

\textbf{BERTScore, SentenceBERT}
Since these methods operate on a sentence level, they do not handle open-ended verification of multi-sentence passages.

\textbf{SelfCheckGPT}

When evaluating an open-ended task, this method computes the hallucination score of $k$ samples from an LLM.

In the \textbf{BERTScore} variant, we consider all $k$ samples when computing the similarity score, resulting in $s*k$ applications of the score computation, and an additional parallel reduction step of $O(\log(k))$.

In the \textbf{NLI} variant, we need to evaluate each document sentence against all $k$ samples, resulting in $s*k$ inference operations and $O(s*k)$ arithmetical operations for the total summary.

\textbf{HaloCheck}
The full comparison between $k$ responses requires computing the $SummaC$ between all pairs, which creates $k^2$ pairs of documents, and we compute the $SummaC$ score for each.
An additional step of computing the mean of the adjacency matrix adds $O(k^2)$ operations, which can be parallelized with a reduction step of $O(\log(k))$ depth.
Thus, the total depth is $D_I + O(\log(sk))$, and the total work is $k^2 (s^2 * W_I + O(s^2) + 1)$.

\textbf{G-Eval}
This method requires $k=20$ samples to estimate the token probability, resulting in $k$ calls to the LLM inference and $O(k)$ arithmetical steps.

\textbf{GPTScore}
This method computes the score using LLM inference. We could not approximate the total number of operations performed in the paper.

\newpage
\section{Specification of Prompts}
\label{sec:app:prompts}

\definecolor{darkgrey}{HTML}{4A4A4A}
\definecolor{lightgrey}{HTML}{CCCCCC}

\newtcolorbox{prompt}[2][]{%
  enhanced,
  breakable,
  colframe=darkgrey,
  colback=white,
  title style={fill=darkgrey},
  coltitle=white,
  fonttitle=\bfseries,
  title=#2,
  fontupper=\ttfamily\small,
  #1
}

We now show the prompt template used for the query generation of the ``similar description'' use case. A list of ``Generic'' and ``Precise'' topics is used to replace \#\#\# HERE \#\#\# with an actual topic. The aim is to generate two passages of text that look different, but are the same content-wise.

\begin{prompt}{Prompt Template for the Query Generation of the ``Similar Description'' Use Case}
% \begin{table}[h]
%   \centering
%   %\renewcommand{\arraystretch}{1.5}
%   \caption{Prompt template used for the query generation of the ``similar description'' use case. A list of ``Generic'' and ``Precise'' topics is used to replace \#\#\# HERE \#\#\# with an actual topic. The aim is to generate two passages of text that look different, but are the same content-wise.}
%   %\label{app:similar-desc-prompt}
%   \begin{tabularx}{\textwidth}{X}
%   \toprule
\#\#\# INSTRUCTION \#\#\#\\
\\
Hello. Please generate two passages of text. They should both describe the same thing (\#\#\# HERE \#\#\#). However, these two passages should differ VASTLY in their length, style. \\
I want you to give an answer using the following format: \\
\textless{}formatting\textgreater{} \\
\#\#\# DESCRIPTION 1 \#\#\# \\
the actual description here... \\
\#\#\# DESCRIPTION 2 \#\#\# \\
the actual description here... \\
\textless{}/formatting\textgreater{} \\
\\
\#\#\# ANSWER \#\#\#
%   \bottomrule
%   \end{tabularx}
% \end{table}
\end{prompt}

The next prompt template is used for the query generation of the ``different description'' use case. A list of different topics is used to replace \#\#\# HERE 1 \#\#\# and \#\#\# HERE 2 \#\#\# with two actual topics. The aim is to generate two passages of text that seem alike, but are completely different content-wise.

\begin{prompt}{Prompt Template for the Query Generation of the ``Different Description'' Use Case}
% \begin{table}[h]
%   \centering
%   %\renewcommand{\arraystretch}{1.5}
%   \caption{Prompt template used for the query generation of the ``different description'' use case. A list of different topics is used to replace \#\#\# HERE 1 \#\#\# and \#\#\# HERE 2 \#\#\# with two actual topics. The aim is to generate two passages of text that seem alike, but are completely different content-wise.}
%   %\label{app:diff-desc-prompt}
%   \begin{tabularx}{\textwidth}{X}
%   \toprule
\#\#\# INSTRUCTION \#\#\#\\
\\
Hello. Please generate two passages of text. They should describe two different things: \\
1. \#\#\# HERE 1 \#\#\# \\
2. \#\#\# HERE 2 \#\#\# \\
\\
However, these two passages should have the same length and style. \\
I want you to give an answer using the following format: \\
\textless{}formatting\textgreater{} \\
\#\#\# DESCRIPTION 1 \#\#\# \\
the actual description here... \\
\#\#\# DESCRIPTION 2 \#\#\# \\
the actual description here... \\
\textless{}/formatting\textgreater{} \\
\\
\#\#\# ANSWER \#\#\#
%   \bottomrule
%   \end{tabularx}
% \end{table}
\end{prompt}

\newpage

The following prompt template is used to ``extract respective terms and their definitions'' from chunks of legal documentation. Given the complexity of the task, we provide the concrete format as well as an in-context example. [\#\#\# REPLACE WITH CONTEXT \#\#\#] gets replaced by a text chunk from the legal definitions dataset.

\begin{prompt}{Prompt Template for Extraction of Legal Terms and Their Definitions}
% \begin{table}[h]
%   \centering
%   %\renewcommand{\arraystretch}{1.5}
%   \caption{Prompt template used to ``extract respective terms and their definitions'' from chunks of legal documentation. Given the complexity of the task, we provide the concrete format as well as an in-context example. [\#\#\# REPLACE WITH CONTEXT \#\#\#] gets replaced by a text chunk from the legal definitions dataset.}
%   %\label{tab:prompt}
%  \begin{tabularx}{\textwidth}{X}
%   \toprule
\#\#\# INSTRUCTION \#\#\#\\
\\
You are a lawyer.\\
\\
\#\#\# QUESTION \#\#\# \\
\\
Based on the provided context extract all the legal definitions. Answer using the following formatting.\\
\textless{}formatting\textgreater{} \\
Term.Definition \\
Term.Definition \\
...\\
\textless{}/formatting\textgreater{} \\
\textless{}example\textgreater{} \\
{[}...{]}\\
\#\#\# CONTEXT \#\#\#  \\
\\
Preliminary Note \\
The Stock Purchase Agreement sets forth the basic terms of the purchase and sale of the preferred stock to the investors (such as the purchase price, closing date, conditions to closing) and identifies the other financing documents. Generally this agreement does not set forth either (1) the characteristics of the stock being sold (which are defined in the Certificate of Incorporation) or (2) the relationship among the parties after the closing, such as registration rights, rights of first refusal and co-sale and voting arrangements (these matters often implicate persons other than just the Company and the investors in this round of financing and are usually embodied in separate agreements to which those others persons are parties, or in some cases in the Certificate of Incorporation). The main items of negotiation in the Stock Purchase Agreement are therefore the price and number of shares being sold, the representations and warranties that the Company must make to the investors and the closing conditions for the transaction. \\
SERIES A PREFERRED STOCK PURCHASE AGREEMENT \\
THIS SERIES A PREFERRED STOCK PURCHASE AGREEMENT (this “Agreement”), is made as of [], 20[], by and among [\_\_\_\_\_\_], a Delaware corporation (the “Company”), and the investors listed on Exhibit A attached to this Agreement (each a “Purchaser” and together the “Purchasers”). \\
The parties hereby agree as follows: \\
\\
\#\#\# ANSWER \#\#\# \\
\\
Agreement. THIS SERIES A PREFERRED STOCK PURCHASE AGREEMENT \\
Company. Delaware corporation \\
Purchaser. Company or the investors listed on Exhibit A \\
Purchasers. Company and the investors listed on Exhibit A together \\
\textless{}/example\textgreater{} \\
\\
\#\#\# CONTEXT \#\#\# \\ 
\\
{[}\#\#\#REPLACE WITH CONTEXT\#\#\#{]} \\
\\
\#\#\# ANSWER \#\#\#
%   \bottomrule
%   \end{tabularx}
% \end{table}
\end{prompt}

\newpage

The subsequent prompt template is used for the ground-truth generation query of the ``hallucination detection'' use case. A legal or scientific topic is used to replace \#\#\# TOPIC \#\#\#.

\begin{prompt}{Prompt Template for Ground-Truth Generation of ``Hallucination Detection'' Use Case}
% \begin{table}[h]
%   \centering
%   %\renewcommand{\arraystretch}{1.5}
%   \caption{Prompt template used for the ground-truth generation query of the ``hallucination detection'' use case. A legal or scientific topic is used to replace \#\#\# TOPIC \#\#\#.}
%   %\label{tab:prompt}
%   \begin{tabularx}{\textwidth}{X}
%   \toprule
\#\#\# INSTRUCTION \#\#\#\\
\\
Hello. Please generate a passage of text that talks about (\#\#\# TOPIC \#\#\#). \\
\\
Please, use the following format for answering: \\
\textless{}formatting\textgreater{} \\
\#\#\# PASSAGE \#\#\# \\
The passage here.... \\
\textless{}/formatting\textgreater{}
%   \bottomrule
%   \end{tabularx}
% \end{table}
\end{prompt}

The next prompt template is used for the hallucination generation query of the ``hallucination detection'' use case. A legal or scientific topic is used to replace \#\#\# TOPIC \#\#\#. \#\#\# NUMBER \#\#\# is replaced according to an user-specified range of numbers. \#\#\# ERRORS \#\#\# is used during the hallucination generation process, but is removed from the sample output before the embeddings are created.

\begin{prompt}{Prompt Template for Hallucination Generation of ``Hallucination Detection'' Use Case}
% \begin{table}[h]
%   \centering
%   %\renewcommand{\arraystretch}{1.5}
%   \caption{Prompt template used for the hallucination generation query of the ``hallucination detection'' use case. A legal or scientific topic is used to replace \#\#\# TOPIC \#\#\#. \#\#\# NUMBER \#\#\# is replaced according to an user-specified range of numbers. \#\#\# ERRORS \#\#\# is used during the hallucination generation process, but is removed from the sample output before the embeddings are created.}
%   %\label{tab:prompt}
%   \begin{tabularx}{\textwidth}{X}
%   \toprule
\#\#\# INSTRUCTION \#\#\#\\
\\
Hello. Please generate \#\#\# NUMBER \#\#\# completely false information (fact hallucinations) on (\#\#\# TOPIC \#\#\#). \\
Then insert the errors inside a passage of text that talks about (\#\#\# TOPIC \#\#\#). \\
You should convince a reader that the false information are actually correct ones. \\
\\
Please, use the following format for answering: \\
\textless{}formatting\textgreater{} \\
\#\#\# ERRORS \#\#\# \\
List of fact hallucinations to be later included in the passage... \\
\#\#\# PASSAGE \#\#\# \\
The passage here.... \\
\textless{}/formatting\textgreater{}
%   \bottomrule
%   \end{tabularx}
% \end{table}
\end{prompt}

\newpage

\subsection{LLM-as-a-Judge Prompt Templates}

We now show the prompt template employed to assess hallucination scores using the LLM-as-a-Judge methodology for both description tasks (``similar`` and ``different''). In this template, \{description1\} and \{description2\} are substituted with their respective similar or different descriptions of a certain topic.

\begin{prompt}{LLM-as-a-Judge Prompt Template for Scoring Hallucinations in Description Tasks}
\#\#\# INSTRUCTION \#\#\# \\
\\
You are a linguistic expert. You will be given two separate descriptions. Your job is to rate how similar the two descriptions are based on the content of the description. You will need to output a score from 0 to 100, where 0 means the description are about completely different things, and 100 means the descriptions are about the same thing. Use the full range of scores, 0, 1, 2, ... 10, 20, ... 90, 100. \\
\\
\#\#\# OUTPUT \#\#\#\\
\\
The output should be a single number, which is the score from 0 to 100.\\
You CANNOT output any other text. You CANNOT output a decimal number. You MUST output an integer number. You MUST NOT output a number that is less than 0 or greater than 100.\\
\\
\#\#\# INPUT \#\#\#\\
\{description1\}\\
\{description2\}
\end{prompt}

The next prompt template is used to detect hallucination of legal summaries, where \{legal\_summary\} and \{original\_document\} are respectively substituted with the LLM generated summary and the original chunk of text used for the summary creation.

\begin{prompt}{LLM-as-a-Judge Prompt Template for Hallucination Detection of Legal Summaries}
\#\#\# INSTRUCTION \#\#\# \\
\\
You are a linguistic and law expert. You will be given a summary of part of a legal document and the original as well. Your job is to rate how accurate and not hallucinated the summary is based on the content of the original passage. You will need to output a score from 0 to 100, where 0 means the summary is completely hallucinated, and 100 means the summary is completely accurate.\\
\\
\#\#\# OUTPUT \#\#\#\\
\\
The output should be a single number, which is the score from 0 to 100.\\
You CANNOT output any other text. You CANNOT output a decimal number. You MUST output an integer number. You MUST NOT output a number that is less than 0 or greater than 100.\\
\\
\#\#\# INPUT \#\#\#\\
\{legal\_summary\}\\
\\
\#\#\# ORIGINAL PASSAGE \#\#\#\\
\{original\_document\}
\end{prompt}

\newpage

The following prompt template is used for the hallucination detection of scientific descriptions, where \{text\_passage\} is substituted with an increasingly hallucinated description of a well-known scientific topic.

\begin{prompt}{LLM-as-a-Judge Prompt Template for Hallucination Detection of Scientific Descriptions}
\#\#\# INSTRUCTION \#\#\# \\
\\
You are a linguistic expert. You will be given a passage of text containing some scientific descriptions about a topic. Your job is to rate how hallucinated the passage is based on the content of the description. You will need to output a score from 0 to 100, where 0 means the passage is completely hallucinated, and 100 means the passage is completely accurate.\\
\\
\#\#\# OUTPUT \#\#\#\\
\\
The output should be a single number, which is the score from 0 to 100.\\
You CANNOT output any other text. You CANNOT output a decimal number. You MUST output an integer number. You MUST NOT output a number that is less than 0 or greater than 100.\\
\\
\#\#\# INPUT \#\#\#\\
\{text\_passage\}
\end{prompt}

The subsequent prompt template is used for hallucination detection with the WikiBio dataset, where \{biography\} is substituted with a hallucinated biography of a famous person generated by an LLM, published as part of the dataset.

\begin{prompt}{Judge-as-a-LLM Prompt Template for Hallucination Detection with the WikiBio Dataset}
\#\#\# INSTRUCTION \#\#\# \\
\\
You are a linguistic and historian expert. You will be given a passage of text containing a brief biography of a famous person. Your job is to rate how hallucinated the biography is. You will need to output a score from 0 to 100, where 0 means the biography is completely hallucinated, and 100 means the biography is completely correct. Use the full range of scores, 0, 1, 2, ... 10, 20, ... 90, 100.\\
\\
\#\#\# OUTPUT \#\#\#\\
\\
The output should be a single number, which is the score from 0 to 100.\\
You CANNOT output any other text. You CANNOT output a decimal number. You MUST output an integer number. You MUST NOT output a number that is less than 0 or greater than 100.\\
\\
\#\#\# INPUT \#\#\#\\
\{biography\}
\end{prompt}

\newpage

%The next prompt template is used in the context of ``WikiBio Hallucination Detection with Reference'' task. In this template, \{generated\_biography\} and \{original\_biography\} are respectively substituted with an eventual hallucinated biography of famous people and the original biography taken from Wikipedia biography. Both the hallucinated passage and the correct biography are taken from the WikiBio dateset.
The following prompt template also concerns the WikiBio dataset, but augments the previous prompt template with the original biography as a reference, which is also taken from the dataset.

\begin{prompt}{Judge-as-a-LLM Prompt Template for Hallucination Detection with the WikiBio Dataset with the Original Biography as Additional Reference}
\#\#\# INSTRUCTION \#\#\# \\
\\
You are a linguistic and historian expert. You will be given a passage of text containing a brief biography of a famous person, you will also be given the complete original biography of the same famous person. Your job is to rate how hallucinated the biography is compared to the original, longer, one. You will need to output a score from 0 to 100, where 0 means the biography is completely hallucinated, and 100 means the biography is completely correct. Use the full range of scores, 0, 1, 2, ... 10, 20, ... 90, 100. The original biography is always longer, this should be not taken into account as hallucination.\\
\\
\#\#\# OUTPUT \#\#\#\\
\\
The output should be a single number, which is the score from 0 to 100.\\
You CANNOT output any other text. You CANNOT output a decimal number. You MUST output an integer number. You MUST NOT output a number that is less than 0 or greater than 100.\\
\\
\#\#\# INPUT \#\#\#\\
**Passage**:\\
\{generated\_biography\}\\
\\
**Original**:\\
\{original\_biography\}
\end{prompt}

The final prompt template is used for hallucination detection with RAGTruth, where \{generated\_answer\} and \{original\_prompt\} are respectively substituted with an LLM generated answer and the prompt used to generate said answer, which are both taken from the dataset.

\begin{prompt}{Judge-as-a-LLM Prompt Template for Hallucination Detection with RAGTruth}
\#\#\# INSTRUCTION \#\#\# \\
\\
You are an expert highly skilled in Summarization, Question Answering and in Converting Data to Text. You will be given an answer to some kind of task, you will also given the original request made by the task. Your task is to evaluate the answer based on the original request, is the answer correct? Or is it factually incorrect and hallucinated?. You will give a score from 0 to 100, where 0 means the answer is completely hallucinated and 100 means the answer is completely correct. Use the full range of scores, 0, 1, 2, ... 10, 20, ... 90, 100.\\
\\
\#\#\# OUTPUT \#\#\#\\
\\
The output should be a single number, which is the score from 0 to 100.\\
You CANNOT output any other text. You CANNOT output a decimal number. You MUST output an integer number. You MUST NOT output a number that is less than 0 or greater than 100.\\
\\
\#\#\# ANSWER \#\#\#\\
\{generated\_answer\}\\
\\
\#\#\# ORIGINAL REQUEST \#\#\#\\
\{original\_prompt\}
\end{prompt}

\iftr
\FloatBarrier
\fi

\newpage
\section{Experimental Setup: Additional Details}
\label{sec:app:setup}

We provide additional details on the evaluation setup for full reproducibility.

\subsection{Embedding Length and Parameter Size}
\label{sec:embed_length}

\begin{table}[hbt!]
  \centering
  \caption{Embedding length and number of parameters for each model used during the evaluation.}
  %\label{tab:prompt}
  \begin{tabular}{lcc}
  \toprule
  \textbf{Model Name} & \textbf{Length} & \textbf{\#Parameters} \\
  \midrule
GPT Text Embedding Large & 3072 & not public\\
Salesforce/SFR-Embedding-Mistral & 4096 & 7.11B\\
intfloat/e5-mistral-7b-instruct & 4096 & 7.11B\\
Alibaba-NLP/gte-Qwen1.5-7B-instruct & 4096 & 7.72B\\
NovaSearch/stella\_en\_1.5B\_v5 & 4096 & 1.54B \\
NovaSearch/stella\_en\_400M\_v5 & 4096 & 435M \\
microsoft/deberta-xlarge-mnli & 1024 & 750M\\
roberta-large & 1024 & 355M \\
CLIP Vision Transformer Large & 768 & 428M\\
  \bottomrule
  \end{tabular}
\end{table}
% \lorenzo{Stella en 400 and 1.5 have multiple dimensions: 512, 768, 1024, 2048, 4096, 6144 and 8192. (That's were variable comes from) \textbf{We used 4096. Generally speeaking: 1024 loses 0.001 accuracy over 8192 in MTEB benchmark}}
% \lorenzo{GPT Text Embedding Large: Dynamically changing the dimensions enables very flexible usage. For example, when using a vector data store that only supports embeddings up to 1024 dimensions long, developers can now still use our best embedding model text-embedding-3-large and specify a value of 1024 for the dimensions API parameter, which will shorten the embedding down from 3072 dimensions, trading off some accuracy in exchange for the smaller vector size. \textbf{We do not knwo how much accuracy it loses}}

\subsection{Compute Resources}
\label{sec:compute_resources}

Running the pipeline for the dataset of legal definitions generated by three LLMs (GPT-3.5, GPT-4 and GPT-4o) as well as the baselines SelfCheckGPT and BERTScore on a single NVIDIA Tesla V100-PCIE-32GB GPU took roughly 90 minutes. That dataset was used to create the heatmap figures. The pipeline for the datasets with similar and different descriptions, used for the violin plots, was executed on the same hardware in around 80 minutes. The experiments for the runtime comparison took 43 hours respectively for each GPU (NVIDIA A100 and NVIDIA RTX3090). The WikiBio experiments took 120 minutes on a single NVIDIA A100. The RAGTruth benchmark, using the same pipeline and hardware, required 30 hours. We ran HalluDetect for six hours on two NVIDIA A100 GPUs for the RAGTruth benchmark. Finally, most LLM-as-a-Judge executions took around 20 minutes on four GH200 GPUs with the exception of running the RAGTruth dataset, which required approximately two hours.

% \lorenzo{Standard runtimes on the RTX3090 took 1 day and 19 hours}
% \lorenzo{Old runtimes on the A100 took 1 day and 19 hours}
% \lorenzo{This is quite a coincidence, but we must take in consideration that SelfCheckGPT NLI is not present in the old data and it takes most of the time there alone. (There is difference in minutes and seconds though)}

\subsection{Comparison Baselines: Additional Details}

BERTScore was previously introduced, along with its workflow. However, please note
that by default, BERTScore outputs a similarity score between two input
sentences. This approach proved highly ineffective when used within
SelfCheckGPT. As a baseline for comparison, we applied BERTScore using our
pipeline and metrics, which resulted in significantly better performance than
SelfCheckGPT BERTScore, though it still under-performed compared to pure \name.

As we already provide an overview over the general SelfCheckGPT pipeline, we will now
give a more detailed description of each evaluator we considered and the
rationale behind our choices.
\begin{itemize}
\item SelfCheckGPT-NLI employs Natural Language Inference (NLI), a method that
  classifies relationships between texts as entailment, neutral, or
  contradiction. It utilizes a fine-tuned DeBERTa-v3 model to detect
  contradictions in generated text. By providing a passage and the sentence
  to be verified, the model computes a contradiction score based on the logits
  for 'entailment' and 'contradiction'. This score determines whether the
  generated sentence contradicts the provided context. NLI represents the most
  effective evaluation method offered by SelfCheckGPT, excluding the Prompt
  method, and performs significantly better than BERTScore.
\item SelfCheckGPT Prompt – While the Prompt method was the most effective in
  their evaluations, it was impractical for our purposes due to its cost. This
  method requires comparing each sentence in the sample with every sentence in
  the reference by querying a language model for potential hallucinations. This could
  result in tens of thousands of API calls, making it unsuitable for larger
  datasets. Even the SelfCheckGPT authors recommend NLI over Prompt due to its
  more manageable costs.
\item SelfCheckGPT-BERTScore was selected
  as a baseline because its approach most closely aligns with our pipeline. It
  also represents the lowest-performing method in SelfCheckGPT.%, serving as a
  %score we aim to surpass.
\end{itemize}

HalluDetect uses two classifiers—Logistic Regression (LR) and a Simple Neural
Network (SNN)—to detect hallucinations in text generated by LLMs. The classifiers are trained on pairs of texts (condition-text,
generated-text) using four numerical features—Minimum Token Probability,
Average Token Probability, Maximum LLME Probability Deviation, and Minimum
LLME Probability Spread—extracted from token probabilities during a forward
pass in a separate LLM evaluator (LLME). This approach leverages probability
distributions from a distinct LLM to identify inconsistencies in the generated
text, allowing for reliable hallucination detection across various tasks, such
as question answering and summarization. Both models are trained in a
supervised learning framework and evaluated on benchmark datasets to ensure
generalizability and accuracy.

The WikiBio dataset consists of 238 Wikipedia article inputs. We used this
dataset to evaluate our method and other baselines on tasks proposed by
SelfCheckGPT. GPT-3 (text-davinci-003) generated 20 reference samples per
input using the prompt: "This is a Wikipedia passage about {concept}:". Each
sentence in the references was manually labeled as either Major Inaccurate,
Minor Inaccurate, or Accurate. Since \nameS operates at the passage level,
passage scores were calculated by averaging the sentence-level scores,
similar to the SelfCheckGPT approach. We used Pearson Correlation and
Spearman Rank for comparison.

The RAGTruth dataset contains around 18,000 responses generated by models
like GPT-3.5, GPT-4, Llama-2, and Mistral in retrieval-augmented generation
tasks. The authors selected 450 responses per LLM for the test set. % \lorenzo{150 responses per LLM per task, or 450 responses per LLM.}
% We generated additional samples for these responses and trained HalluDetect using 2.5k datapoints from the test set while testing on the remaining 150 responses per model.
We provide further details in Appendix~\ref{sec:app:ragtruth}.

% \lorenzo{Add details about RAGTruth dataset}

% \lorenzo{They are already present in Section D.5, we need to move details here and just have a simple introduction there?}

\newpage
\section{Additional Results}
\label{sec:app:eval}

\subsection{Distinguishing Text Passages}

\begin{figure}[h]
    \centering
        \includegraphics[width=1.0\linewidth]{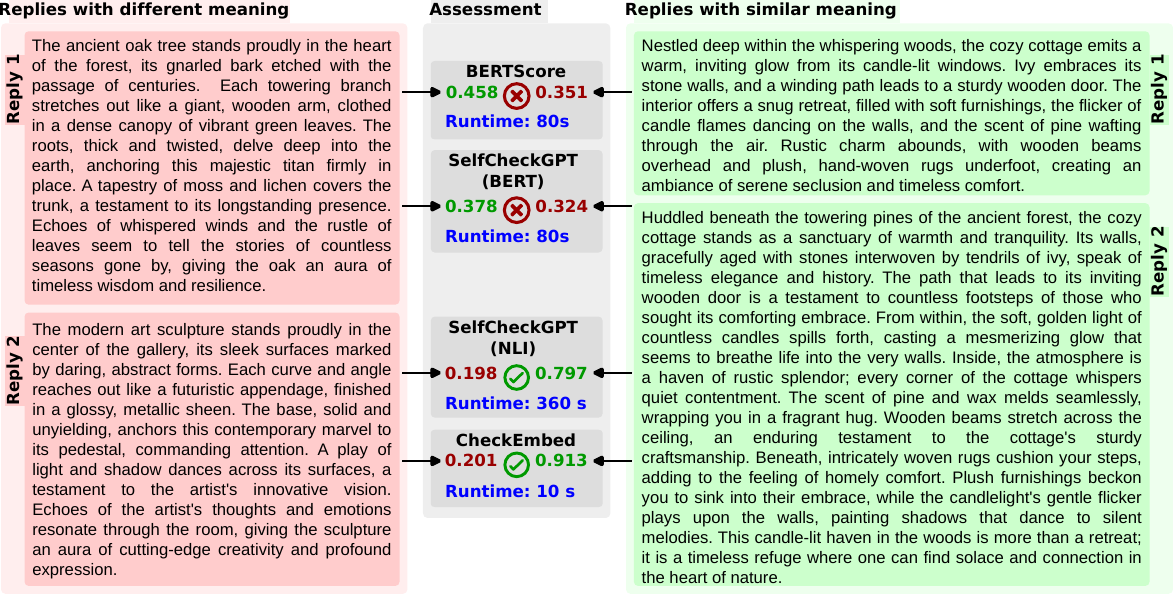}
        \vspaceSQ{-1.5em}
    \caption{\textbf{Comparison of stability-related verification schemes.} We show two sets of two LLM replies each: Replies explaining different concepts using similar wording (left) and ones explaining similar concepts using different wording (right); the queries used to generate these replies can be found in Appendix~\ref{sec:app:prompts}. The task considered is to robustly distinguish the similar pair from the different pair. We used ChatGPT-4o with temperature = 1.0 for generating the replies and Stella 400M for generating embeddings for this example.
    We compare \nameS to two variants of SelfCheckGPT: one that uses BERTScore as a subroutine, and one that harnesses the Natural Language Inference (NLI), which classifies relationships between texts as entailment, neutral, or contradiction, and utilizes a fine-tuned DeBERTa-v3 model~\citep{he2021deberta} to detect textual contradictions by computing a contradiction score based on the logits for 'entailment' and 'contradiction'. We also compare to BERTScore as a standalone baseline. While BERTscore and SelfCheckGPT (BERT) assess the semantically unrelated replies as more related than the related ones (because these two baselines have been designed to mostly target the verification of individual sentences or facts), \textit{\textbf{\nameS} correctly differentiates between semantically related and unrelated replies}, and significantly outperforms SelfCheckGPT (NLI) in speed. 
    }
    %\vspaceSQ{-1.15em}
    \label{fig:posterchild}
\end{figure}

The prompt sizes used for the two groups of similar and different text passages are in the range of 25--250 and 100--200 tokens, respectively. To broaden the analysis, we further consider two subtypes of such passages: ``Generic'' and ``Precise''. The former are brief while the latter are rich in detailed information (e.g., ``Vintage bike'' vs.~``Old, rusted bicycle leaning against a weathered fence''). We illustrate the results for these two subtypes in Figure~\ref{fig:eval-violins-gpt}.% Figures~\ref{fig:eval-violins-gpt-precise} and~\ref{fig:eval-violins-gpt-generic}.

\if 0
%
we asked each LLM to provide two passages of text, for similarity we asked one short and one long description varying in style. We tried giving two different degree of freedom to the LLMs. Generating this passages of text feeding that with super Precise Topics
(Ex... 
"Old, rusted bicycle leaning against a weathered fence",
"Brightly colored hot air balloons rising at dawn") and much more Generic Ones 
(Ex.. "Vintage cars",
"Mountaintop views",
"Old libraries").

Precise topics offer the LLMs less freedom to go off and add much more and diverse details on the long passage of text where it has more space. While giving the LLM purely generic topics offer the possibility to achieve this and add details not specified in the short passage of text. Nevertheless the results from both Precise and Generic show a very consistent behaviour of giving high scores (0.8/0.9) contributing once more to prove that we are able to detect when the topic of two different passages of text are similar.

For the diversity analysis we prompted the llm (Put prompt shceme in appendix) asking for two passages of text of similar lenght and style. But giving two very different topics to describe. \nameS isn't getting fooled by the similarity in style and lenght of the passages.
%
\fi

Interestingly, GPT-4-turbo generates replies that are `the most difficult to distinguish'', i.e., it comes with visible (still very low) overlap between similar and different ones, across all embedding models. Contrarily, GPT-4o comes with no overlap whatsoever, while GPT-3.5 has very minor overlap.

\begin{figure}[t]
  \centering
  \includegraphics[width=1.0\linewidth]{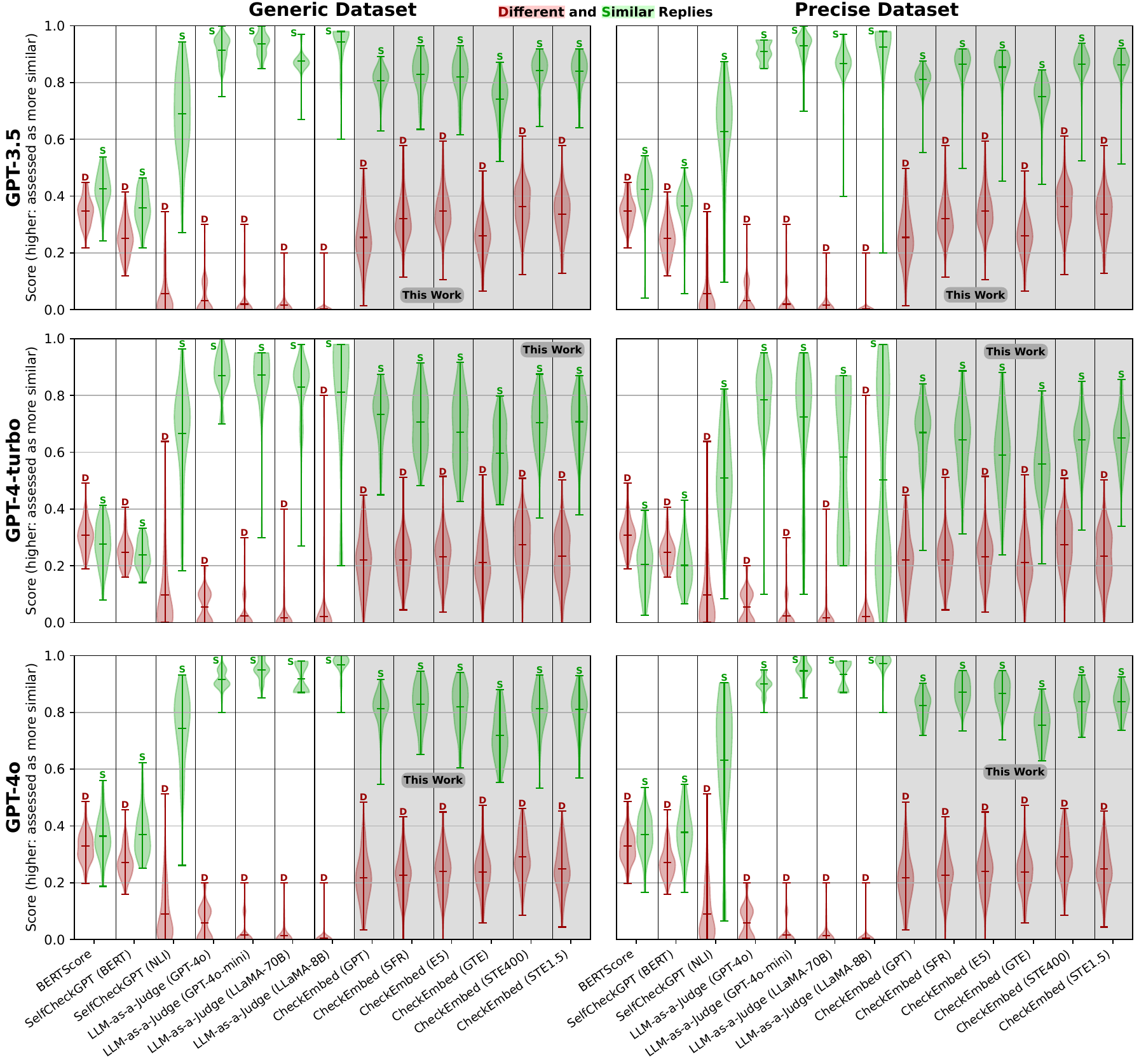}
  \vspaceSQ{-1em}
  \caption{\textbf{Analysis of distinguishing similar and different LLM replies}, details explained in Section~\ref{sec:eval-distinguish}. \nameAS is (highly) effective at appropriately recognizing the similarities and differences in the meaning of the verified text passages. This can be seen from moderate to no overlap between groups of data points corresponding to scores for -- respectively -- similar and different LLM replies, regardless of the model used. Contrarily, there is a large overlap between these groups of data points for both BERTScore and SelfCheckGPT (BERT), indicating that these baselines perform worse in distinguishing such replies effectively, while SelfCheckGPT (NLI) shows a better, but still noticely inferior to \nameA, distinction between those two groups. We additionally plot the LLM-as-a-Judge results – even though it does not rely on comparing text passages for verification, it serves as the top baseline in this task, considering its training objective, which enables (among others) distinguishing complex nuances in text.}
  \vspaceSQ{-1.5em}
  \label{fig:eval-violins-gpt}
\end{figure}

\clearpage

\subsection{Heatmaps}
\label{sec:additional_heatmaps}

We show additional heatmaps based on samples generated by different language models (GPT-3.5 and GPT-4) as well as using different embedding models (GPT Text Embedding Large, Stella 1.5B) in Figures~\ref{fig:heatmap_gpt4} to~\ref{fig:heatmap_gpt35_stella}.

These heatmaps illustrate that whenever \nameS has very high confidence in its answer (top rows of the heatmaps), which is visible by consistently having very high similarities between different replies, it corresponds to very high similarity scores between the LLM replies and the ground-truth. This is the case for all the considered models. Other baselines show mixed results for individual replies, and low similarities between their replies and the ground-truth. It shows that, whenever \nameS has high confidence it the LLM replies, there is high likelihood that these replies are close to the corresponding ground-truth.

\begin{figure}[h]
    \centering
    \includegraphics[width=\textwidth]{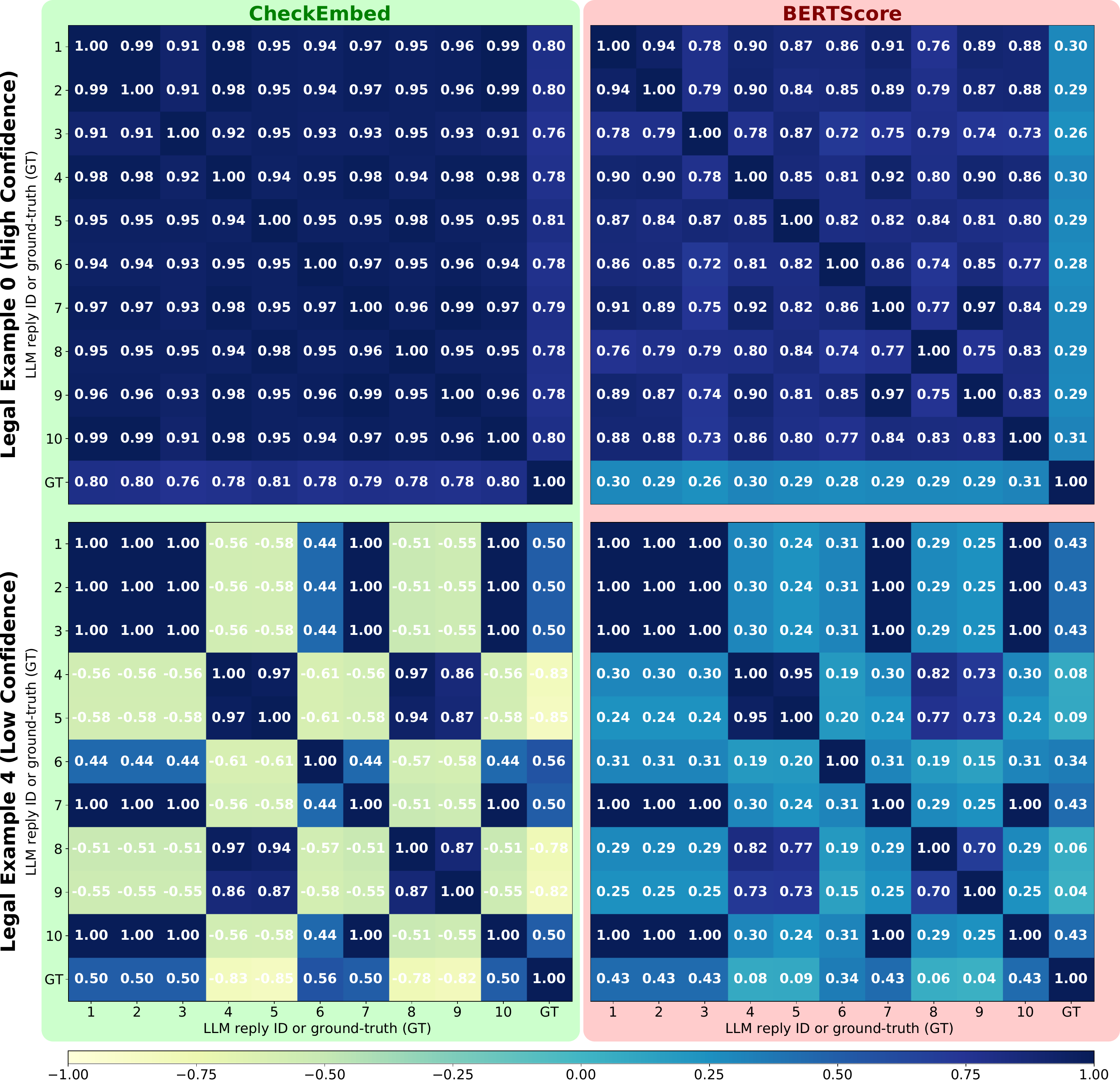}
    \caption{\textbf{Analysis of the verification of LLM answers}, details explained in Section~\ref{sec:eval-heatmaps}. We compare to BERTScore; SelfCheckGPT (BERT) comes with significantly higher runtimes (detailed in Section~\ref{sec:runtimes}) and less competitive scores as it does not focus on open-ended answer-level analysis. The results form a heatmap of the \nameA's, or BERTScore's, cosine similarity between all LLM replies, and between each reply and the human expert prepared ground-truth (GT). Rows correspond to two representative legal documents, that come with -- respectively -- high and low LLM confidence in its replies. Embedding model used: GPT Text Embedding Large. Generative model used: GPT-4.}
    %
    % Data from legal-definitions-task\_small.
    %
    %Frob\_score: 0.98 - 0.83 / 0.83 - 0.57 / 0.75 - 0.59
    %Std\_dev: 0.01 - 0.06 / 0.17 - 0.14 / 0.23 - 0.32
    %\lorenzo{ CHECK }}
    \label{fig:heatmap_gpt4}
\end{figure}

\begin{figure}[hbt!]
    \centering
    \includegraphics[width=\textwidth]{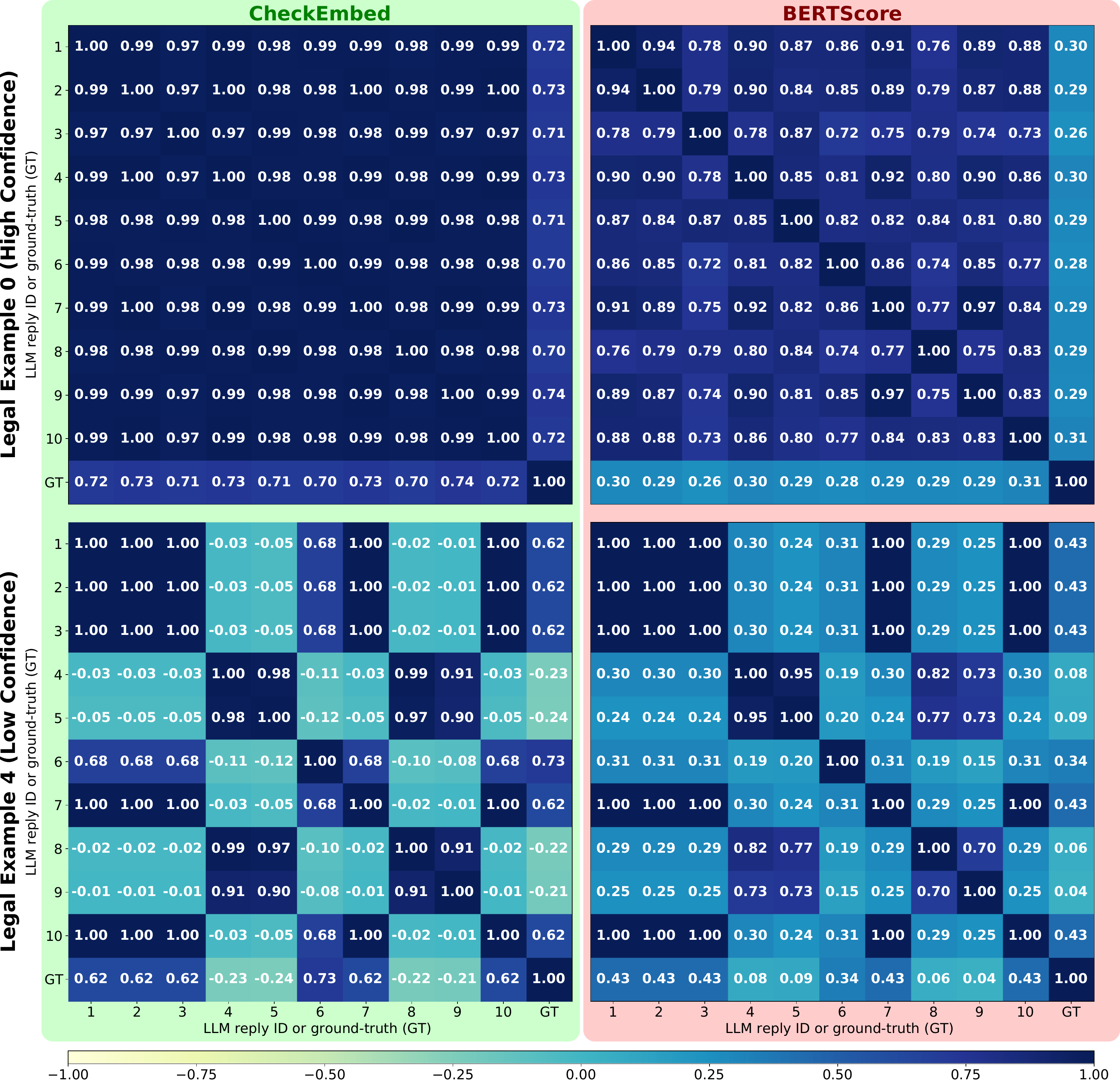}
    \caption{\textbf{Analysis of the verification of LLM answers}, details explained in Section~\ref{sec:eval-heatmaps}. We compare to BERTScore; SelfCheckGPT (BERT) comes with significantly higher runtimes (detailed in Section~\ref{sec:runtimes}) and less competitive scores as it does not focus on open-ended answer-level analysis. The results form a heatmap of the \nameA's, or BERTScore's, cosine similarity between all LLM replies, and between each reply and the human expert prepared ground-truth (GT). Rows correspond to two representative legal documents, that come with -- respectively -- high and low LLM confidence in its replies. Embedding model used: Stella 1.5B. Generative model used: GPT-4.}
    \label{fig:heatmap_gpt4_stella}
\end{figure}

\begin{figure}[hbt!]
    \centering 
    \includegraphics[width=\textwidth]{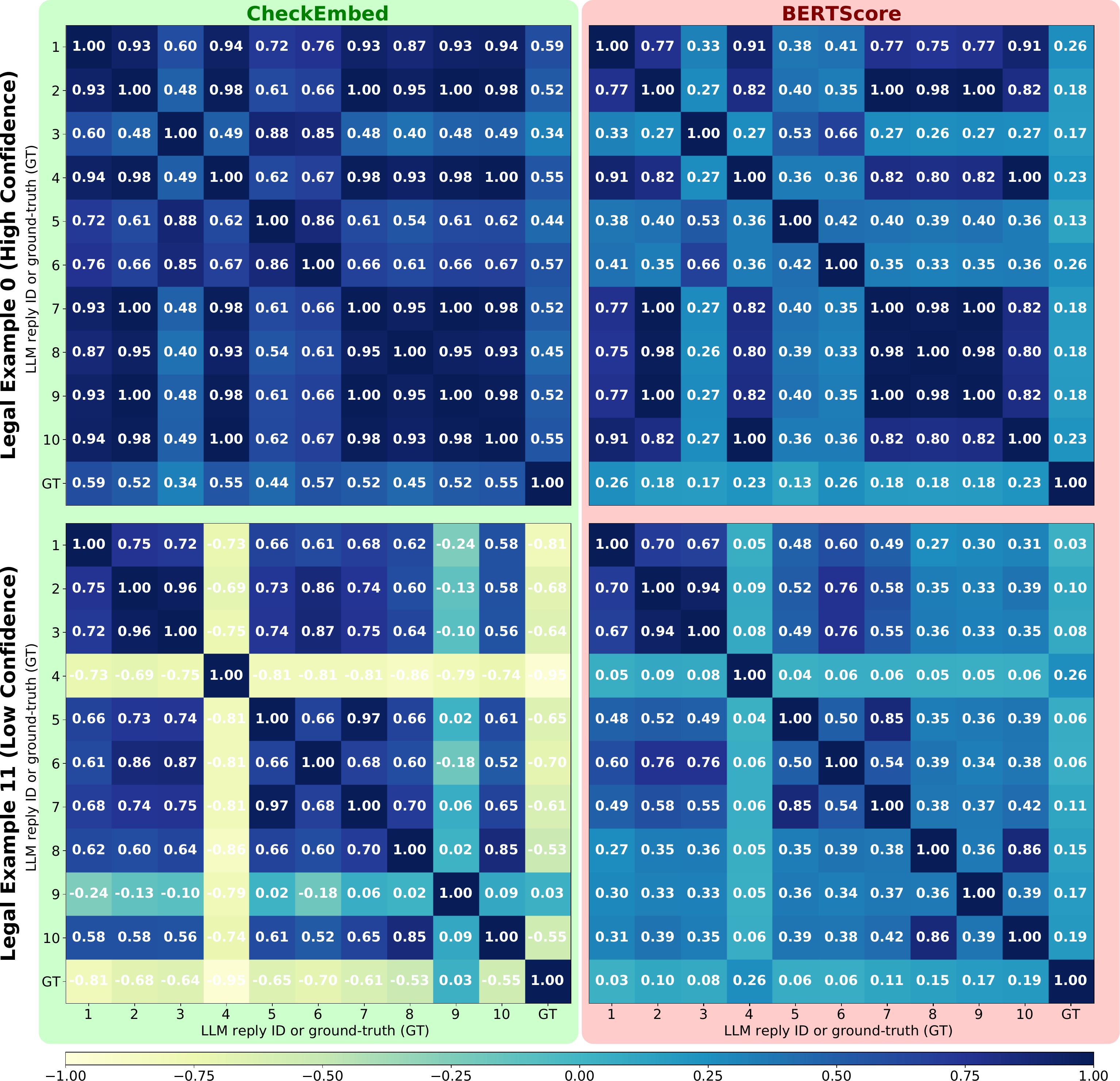}
    \caption{\textbf{Analysis of the verification of LLM answers}, details explained in Section~\ref{sec:eval-heatmaps}. We compare to BERTScore; SelfCheckGPT (BERT) comes with significantly higher runtimes (detailed in Section~\ref{sec:runtimes}) and less competitive scores as it does not focus on open-ended answer-level analysis. The results form a heatmap of the \nameA's, or BERTScore's, cosine similarity between all LLM replies, and between each reply and the human expert prepared ground-truth (GT). Rows correspond to two representative legal documents, that come with -- respectively -- high and low LLM confidence in its replies. Embedding model used: GPT Text Embedding Large. Generative model used: GPT-3.5.}
    %
    % Data from legal-definitions-task\_small.
    %
    %Frob\_score: 0.93 - 0.66 / 0.88 - 0.58 / 0.79 - 0.46
    %Std\_dev: 0.06 - 0.27 / 0.06 - 0.11 / 0.19 - 0.23
    %\lorenzo{ CHECK }}
    \label{fig:heatmap_gpt35}
\end{figure}

\newpage

\begin{figure}[hbt!]
    \centering
    \includegraphics[width=\textwidth]{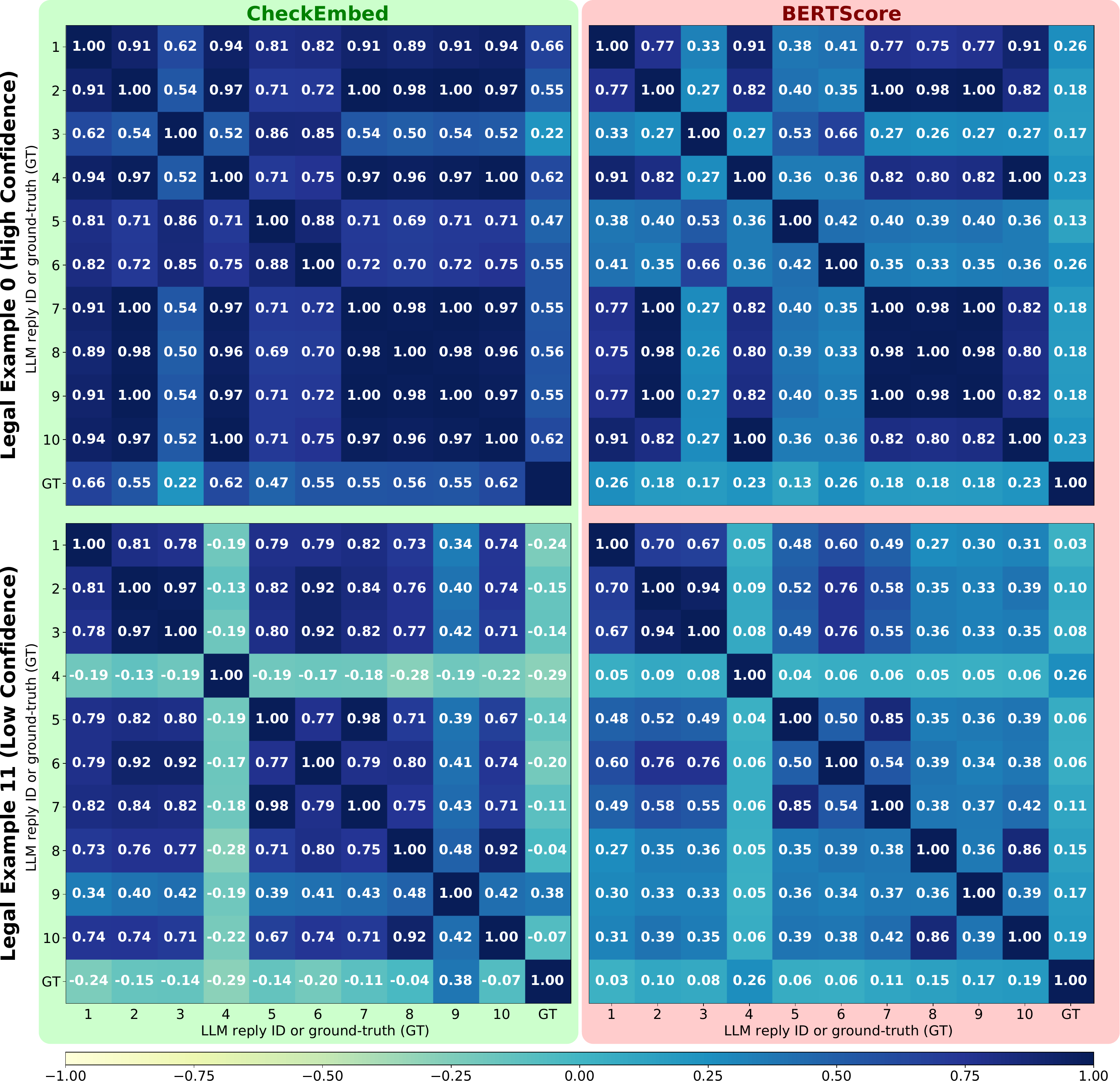}
    \caption{\textbf{Analysis of the verification of LLM answers}, details explained in Section~\ref{sec:eval-heatmaps}. We compare to BERTScore; SelfCheckGPT (BERT) comes with significantly higher runtimes (detailed in Section~\ref{sec:runtimes}) and less competitive scores as it does not focus on open-ended answer-level analysis. The results form a heatmap of the \nameA's, or BERTScore's, cosine similarity between all LLM replies, and between each reply and the human expert prepared ground-truth (GT). Rows correspond to two representative legal documents, that come with -- respectively -- high and low LLM confidence in its replies. Embedding model used: Stella 1.5B. Generative model used: GPT-3.5.}
    \label{fig:heatmap_gpt35_stella}
\end{figure}

\newpage
\clearpage

\subsection{Running Times}
\label{sec:additional_runtime_results}

We illustrate the runtime of \nameS as well as various stability-related verification baselines in Figures~\ref{fig:runtimes_full} to~\ref{fig:runtimes_A100_large}. We omit HalluDetect as it requires training.

% \lorenzo{Justify why HalluDetect and LLM-as-a-Judge are not here? One need training and both need only 1 sample, LLM-as-a-Judge is technically much faster without a doubt}

% \lorenzo{Stability related}

\begin{figure}[hbt!]
  \includegraphics[width=\linewidth]{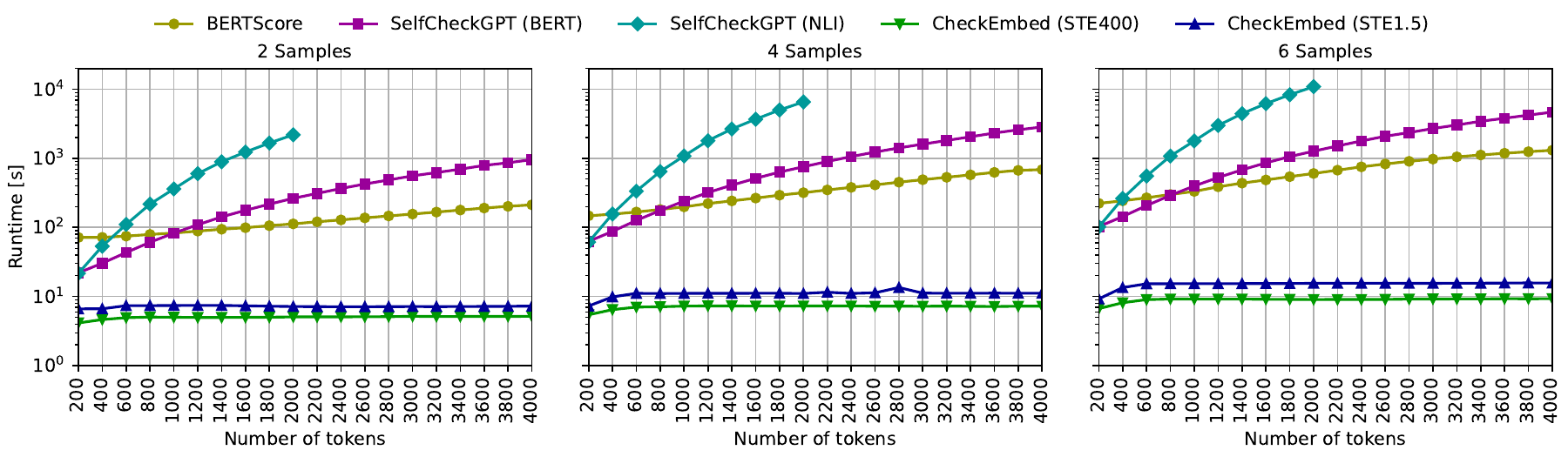}
  \caption{\textbf{Comparison of running times of \nameAS and other baselines while varying the number of samples per datapoint.} We used an NVIDIA RTX3090 GPU for this experiment. Please note the logscale y axis.}
  \label{fig:runtimes_full}
\end{figure}

\begin{figure}[hbt!]
  \includegraphics[width=\linewidth]{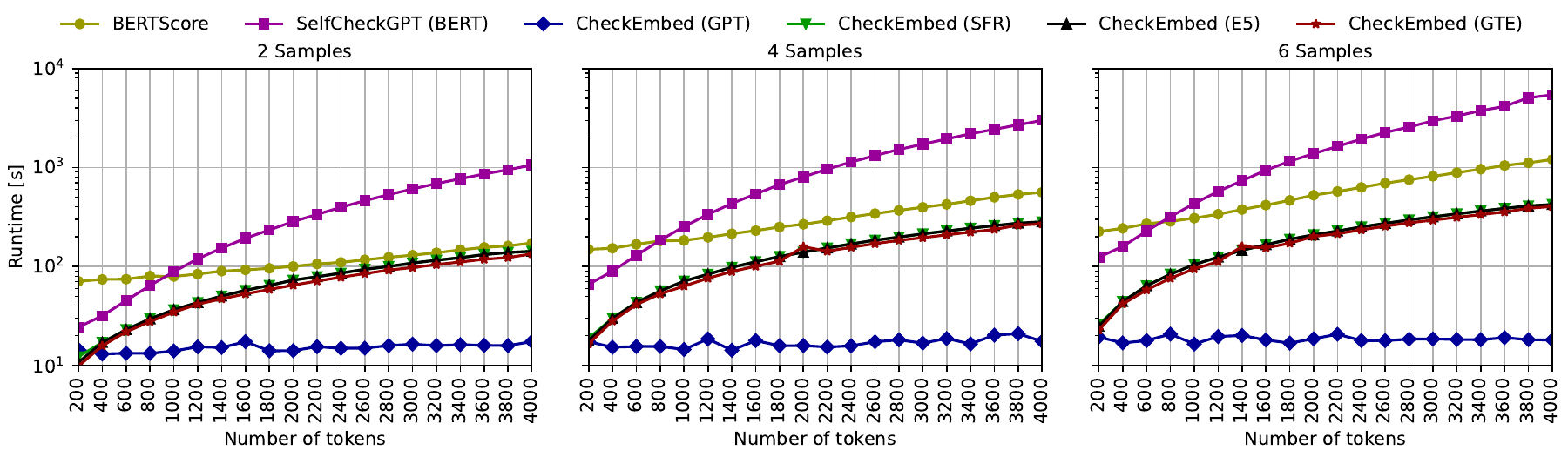}
  \caption{\textbf{Comparison of running times of \nameAS and other baselines while varying the number of samples (2, 4 and 6) per datapoint.} We used an NVIDIA A100 GPU for this experiment. Please note the logscale y axis.}
  \label{fig:runtimes_A100_small}
\end{figure}

\begin{figure}[hbt!]
  \centering
  \includegraphics[width=0.75\linewidth]{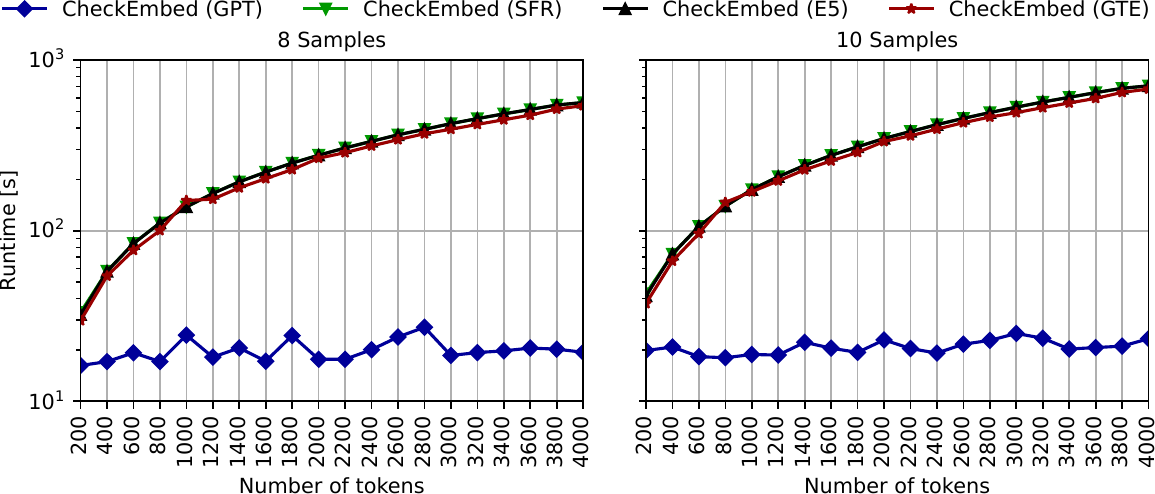}
  \caption{\textbf{Comparison of running times of \nameAS and other baselines while varying the number of samples (8 and 10) per datapoint.} We used an NVIDIA A100 GPU for this experiment. Results for BERTScore and SelfCheckGPT (BERT) are missing, since their execution with larger sample sizes would have taken a long time. Please note the logscale y axis.}
  \label{fig:runtimes_A100_large}
\end{figure}

\if 0
\begin{itemize}
        \item legal-definitions-task\_large has been run on a RTX 3070 8192MB, batch\_size = 1, Time: ~ 5h (local)
        \item legal-definitions-task\_small: Tesla V100-PCIE-32GB, batch\_size = 1 (also for gte\_Qwen1.5-7B-instructor). Time: ~ 1.5h (Ault cluster)
         \item similar-definitions-task: Tesla V100-PCIE-32GB. Time: ~ 0.5h (Ault cluster)
         \item different-definitions-task: Tesla V100-PCIE-32GB.  Time: ~ 0.75h (Autl cluster)
         \item forced-hallucination-task: Grace Hopper 200 - 96 GB. Still running. Time: ~ 10h (?? cluster) (still don't know if it is going to end up in the paper).
\end{itemize}
\fi

\newpage
\subsection{Full WikiBio Results}
\label{sec:wikibio_results}

We provide results for additional LLM-as-a-Judge baselines in Table~\ref{tab:wikibio_overview_full} and the full \nameS results in Table~\ref{tab:wikibio_ce}. As WikiBio consists of only 238 datapoints, we did not use HalluDetect as a baseline, since it requires training, which would not leave enough datapoints for testing.

% \lorenzo{Should we specify somewhere that HalluDetect was only executed for the RAGTruth dataset, seen requiring training we don;t have enough data for the other task to effectively train the model and have enough data left for a proper Test}

\begin{table}[h]
    \centering
    % \renewcommand{\arraystretch}{0.6}
    % \vspaceSQ{-1em}
    % \scriptsize
    \caption{Passage level correlation on the WikiBio-gpt3 dataset using Pearson (PE) and Spearman (SP) with additional LLM-as-a-Judge baselines.}
    \label{tab:wikibio_overview_full}
    \begin{tabular}{lcc}
        \toprule
        \textbf{Method} & \textbf{PE} & \textbf{SP} \\
        \midrule        
        BertScore & 67.7 & 67.9  \\
        SelfCheckGPT (BERT) & 57.4 & 54.6 \\
        SelfCheckGPT (NLI)       & \textbf{74.1} & 73.8 \\
        \midrule
        LLM-as-a-Judge (4o)	                  &  39.7 & 39.4 \\
        LLM-as-a-Judge w/reference (4o)	      &  36.0 & 35.8 \\
        LLM-as-a-Judge (4o mini)              &  27.1 & 31.4 \\
        LLM-as-a-Judge w/reference (4o mini)  &  45.0 & 46.7 \\
        LLM-as-a-Judge (Llama70b)	            &  -7.8 & -8.7 \\
        LLM-as-a-Judge w/reference (Llama70b) & -10.0 & -6.5 \\
        LLM-as-a-Judge (Llama8b)              &   2.3 &  2.9 \\
        LLM-as-a-Judge w/reference (Llama8b)  &  17.7 & 17.3 \\
        \midrule
        \nameS (GPT)    &  66.8 &  72.6 \\ 
        \nameS (STE400) &  68.5 &  72.9 \\
        \nameS (STE1.5) &  69.9 &  73.8 \\
        \nameS (E5)     &  71.6 &  74.1 \\
        \nameS (SFR)    &  72.2 &  76.2 \\
        \nameS (GTE)    &  73.6 &  \textbf{76.2} \\
        \bottomrule
    \end{tabular}
    % \vspaceSQ{-1em}
\end{table}

\begin{table}[ht]
  \centering
  \setlength{\tabcolsep}{3pt}
  \caption{\textbf{Full \nameS results for the WikiBio benchmark.} PE stands for Pearson correlation coefficient and SP for Spearman's rank correlation coefficient.}
  \label{tab:wikibio_ce}
  \begin{tabular}{ccccccccccccc}
  \toprule
  \multirow{2}{*}{\textbf{\#Samples}} & \multicolumn{2}{c}{\textbf{SFR}} & \multicolumn{2}{c}{\textbf{STE400}} & \multicolumn{2}{c}{\textbf{STE1.5}} & \multicolumn{2}{c}{\textbf{GPT}} & \multicolumn{2}{c}{\textbf{E5}} & \multicolumn{2}{c}{\textbf{GTE}} \\
  \cmidrule(lr){2-3} \cmidrule(lr){4-5} \cmidrule(lr){6-7} \cmidrule(lr){8-9} \cmidrule(lr){10-11} \cmidrule(lr){12-13}
   & \textbf{PE} & \textbf{SP} & \textbf{PE} & \textbf{SP} & \textbf{PE} & \textbf{SP} & \textbf{PE} & \textbf{SP} & \textbf{PE} & \textbf{SP} & \textbf{PE} & \textbf{SP} \\
  \midrule
2   & 61.9 & 67.3 & 59.7 & 64.7 & 62.2 & 67.4 & 52.3 & 61.2 & 59.9 & 64.4 & 63.8 & 68.5 \\
4   & 67.9 & 72.3 & 64.4 & 68.9 & 66.3 & 70.3 & 63.1 & 68.9 & 68.8 & 72.0 & 67.8 & 70.3 \\
6   & 70.6 & 74.8 & 66.5 & 70.4 & 68.4 & 71.5 & 64.6 & 69.8 & 71.9 & 75.2 & 69.3 & 71.1 \\
8   & 71.0 & 75.4 & 67.4 & 72.1 & 68.9 & 72.5 & 65.0 & 71.0 & 72.4 & 75.3 & 70.0 & 72.4 \\
10  & 71.6 & 75.7 & 68.2 & 72.3 & 69.5 & 73.0 & 65.6 & 71.4 & 73.3 & 76.0 & 71.0 & 73.6 \\
12  & 71.2 & 75.8 & 67.7 & 72.5 & 69.2 & 73.4 & 66.0 & 71.8 & 72.9 & 75.9 & 71.2 & 73.8 \\
14  & 71.7 & 76.2 & 68.0 & \textbf{73.1} & 69.5 & \textbf{74.0} & 66.5 & 72.6 & 73.2 & 76.2 & 71.4 & 74.1 \\
16  & \textbf{72.2} & \textbf{76.2} & \textbf{68.5} & 72.9 & \textbf{69.9} & 73.8 & \textbf{66.8} & \textbf{72.6} & \textbf{73.6} & \textbf{76.2} & \textbf{71.6} & \textbf{74.1} \\
18  & 71.4 & 75.6 & 67.7 & 72.3 & 69.2 & 73.0 & 66.7 & 72.6 & 72.9 & 75.4 & 71.0 & 73.6 \\
20  & 71.5 & 75.3 & 68.0 & 72.4 & 69.6 & 73.1 & 66.7 & 72.2 & 72.9 & 75.2 & 71.3 & 73.8 \\
  \bottomrule
  \end{tabular}
\end{table}

\if 0
\begin{table}[ht]
  \centering
  \caption{\textbf{Baseline results for the WikiBio benchmark.}}
  \label{tab:wikibio_baselines}
  \begin{tabular}{lcc}
  \toprule
  \textbf{Baseline} & \textbf{Pearson} & \textbf{Spearman} \\
  \midrule
  BERTScore                & 67.7 & 67.9 \\
  SelfCheckGPT (BERTScore) & 57.4 & 54.6 \\
  SelfCheckGPT(NLI)        & \textbf{74.1} & \textbf{73.8} \\
  \bottomrule
  \end{tabular}
\end{table}
\fi

% \subsection{WikiBio}

% Old Comment by Lorenzo:
CheckEmbed demonstrates a robust performance compared to existing benchmarks,
particularly in Spearman's correlation, where our results are significantly
higher than those of competing methods. In Pearson's correlation, we closely
approach the leading scores achieved by SelfCheckGPT's NLI variant, indicating
our strong competitive standing. Our various embeddings show overall
effectiveness, with many outperforming comparable approaches, while only the
SelfCheckGPT (NLI) variant consistently surpasses our results. This versatility means
that we can continually improve our method; as new, more advanced embeddings
are developed, we can integrate them to enhance our accuracy and reliability,
demonstrating the adaptability of CheckEmbed. Although we could not evaluate
HalluDetect on this dataset due to its training requirements—given that the
238 data points are insufficient for effective training—our methodology
remains solid. Consistent with the SelfCheckGPT paper, we employed a passage scoring
system that aggregates sentence scores—assigning 0 for major inaccuracies, 0.5
for minor inaccuracies, and 1 for accurate sentences—before calculating the
average score. This allows us to utilize Pearson and Spearman correlation
metrics, reflecting the nuanced outputs of our methods. Unlike simplistic
black-and-white approaches that only categorize outputs as hallucinated or
not, our framework quantifies the extent of hallucination within passages,
revealing the depth of our analysis as evidenced by our results.

\subsection{Fine-Grained Hallucination Detection}

We also provide results for summarizing scientific documents in Figure~\ref{fig:hallucinate_gpt4}.

\begin{figure}[h]
    \centering
    \includegraphics[width=\textwidth]{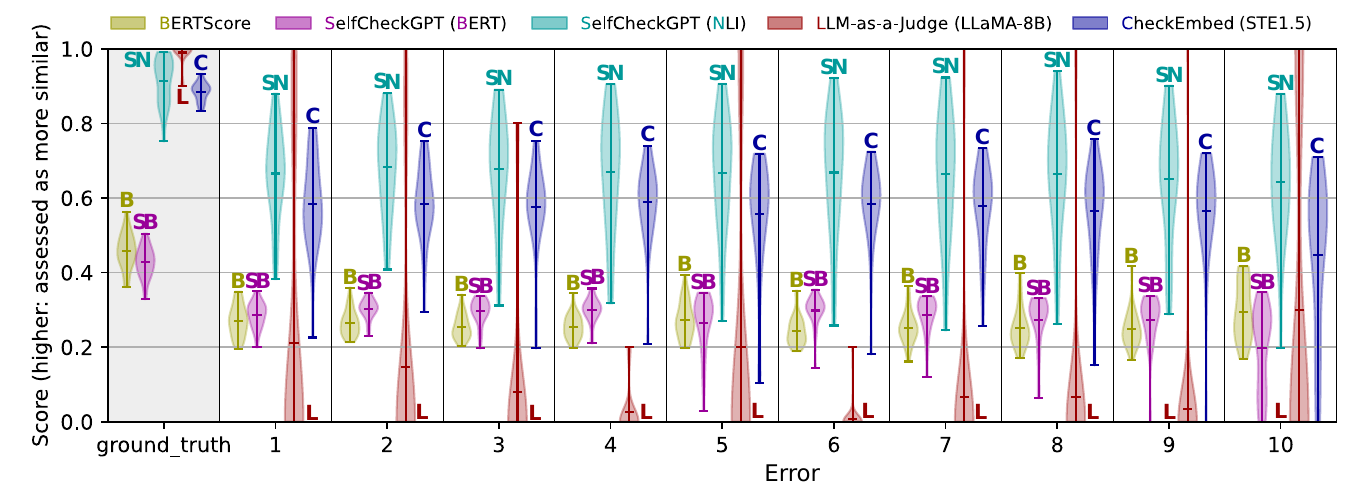}
    \caption{\textbf{Analysis of fine-grained hallucination verification of LLM answers (GPT-4o) when summarizing scientific documents}, details explained in Section~\ref{sec:eval-heatmaps}.}
    \label{fig:hallucinate_gpt4}
\end{figure}

\subsection{RAGTruth Dataset}
\label{sec:app:ragtruth}
%
% The RAGTruth dataset contains around 18,000 responses generated by models
% like GPT-3.5, GPT-4, Llama-2, and Mistral in retrieval-augmented generation
% tasks. The authors selected 150 responses per LLM for the test set. We
% generated additional samples for these responses and trained HalluDetect using 2.5k datapoints from the test set while testing on the
% remaining 150 responses per model.

% \lorenzo{Here's a more complete description of the dataset}

The RAGTruth dataset is a comprehensive benchmark designed to evaluate hallucinations of language models within retrieval-augmented generation (RAG) tasks. It consists of approximately 18,000 responses, which were generated by a diverse set of six LLMs, including OpenAI's GPT-3.5-turbo-0613 and GPT-4-0613, Mistral AI's Mistral-7b-Instruct, and Meta's Llama-2-7B-chat, Llama-2-13B-chat, and Llama-2-70B-chat (4bit quantized).
The approximately 3,000 unique source documents are categorized into three distinct tasks types: Question Answering (QA), Data-to-Text conversion, and Summarization. Each task type contributes around 1,000 responses to the overall dataset.
For the evaluation, the authors carefully selected 150 responses per LLM to create the test set, resulting in a total of 900 responses (150 responses $\times$ 6 LLM) per task. We augment these test responses by generating 10 additional samples for each one. We used 5,000 data points for each task to train the HalluDetect model, which also explains the very high scores of HalluDetect for this benchmark, as it is fine-tuned to these specific tasks whereas the other approaches were used out-of-the-box without any modifications.

\begin{table}[ht]
    \centering
    \caption{Precision, Recall and F1 score for all our baselines on the RAGTruth dataset.} % Hallu need specific training over this task with 90\% of the }
    \label{tab:hallucination_performance}
    % \scriptsize
    % \setlength{\tabcolsep}{1.5pt}
    \resizebox{\textwidth}{!}{
        \begin{tabular}{l ccc ccc ccc}
        \toprule
        \textbf{Method} & \multicolumn{3}{c}{\textbf{Summary}} & \multicolumn{3}{c}{\textbf{QA}} & \multicolumn{3}{c}{\textbf{Data-to-Text}} \\
        \cmidrule(lr){2-4} \cmidrule(lr){5-7} \cmidrule(lr){8-10} % \cmidrule(lr){11-13}
        & Precision & Recall & F1 & Precision & Recall & F1 & Precision & Recall & F1 \\
        \midrule
        \textbf{HalluDetection} \\
        - gpt2-large & 0.7734 & 0.9957 & 0.8706 & 0.8183 & 0.8095 & 0.8139 & 0.3000 & 0.0374 & 0.0665 \\
        - google/gemma-7b-it & \textbf{0.7773} & 0.9727 & 0.8641 & 0.8252 & 0.9122 & 0.8665 & 0.4286 & 0.0841 & 0.1406 \\
        - meta-llama/Llama-2-7b-chat-hf & 0.7749 & 0.9842 & 0.8671 & 0.8204 & 0.8459 & 0.8330 & 0.3810 & 0.0748 & 0.1250  \\
        - facebook/opt-6.7b & 0.7731 & \textbf{0.9986} & \textbf{0.8715} & 0.8270 & 0.8784 & 0.8519 & 0.3137 & 0.0498 & 0.0860  \\
        - EleutherAI/gpt-j-6B & 0.7734 & 0.9957 & 0.8706 & 0.8168 & 0.8554 & 0.8356 & 0.4286 & 0.0093 & 0.0183  \\
        - allenai/led-large-16384-arxiv & 0.7742 & 0.9411 & 0.8495 & 0.8297 & 0.7176 & 0.7696 & 0.3214 & 0.0561 & 0.0955 \\
        - facebook/bart-large-cnn & 0.7685 & 0.8635 & 0.8133 & 0.8266 & 0.7149 & 0.7667 & 0.3804 & 0.1090 & 0.1695  \\
        \midrule
        \textbf{SelfCheckGPT} \\
        - BertScore & 0.7633 & 0.3103 & 0.4413 & 0.9784 & 0.3054 & 0.4655 & 0.4128 & 0.3614 & 0.3854 \\
        - NLI & 0.7672 & 0.7529 & 0.7600 & 0.8759 & 0.6770 & 0.7637 & 0.3597 & 0.9907 & \textbf{0.5278} \\
        \midrule
        \textbf{LLM-as-a-Judge} \\
        - gpt-4o-mini & 0.7697 & 0.9124 & 0.8350 & 0.7273 & 0.0108 & 0.0213 & 0.0000 & 0.0000 & 0.0000  \\
        - gpt-4o & 0.7630 & 0.8649 & 0.8108 & 0.7500 & 0.0041 & 0.0081 & 0.0000 & 0.0000 & 0.0000  \\
        - Llama8B & 0.7650 & 0.9167 & 0.8340 & 0.7547 & 0.1081 & 0.1891 & \textbf{0.5000} & 0.0031 & 0.0062  \\
        - Llama70B & 0.7600 & 0.6825 & 0.7192 & \textbf{1.0000} & 0.0014 & 0.0027 & 0.0000 & 0.0000 & 0.0000  \\
        \midrule
        \textbf{CheckEmbed} \\
        - E5 & 0.7766 & 0.9440 & 0.8521 & 0.8233 & 0.9635 & 0.8879 & 0.3567 & \textbf{1.0000} & 0.5258 \\
        - GPT & 0.7734 & 0.9124 & 0.8372 & 0.8282 & 0.8662 & 0.8468 & 0.3575 & \textbf{1.0000} & 0.5267 \\
        - SFR & 0.7745 & 0.9526 & 0.8544 & 0.8209 & \textbf{0.9784} & \textbf{0.8927} & 0.3567 & \textbf{1.0000} & 0.5258 \\
        - GTE & 0.7737 & 0.9282 & 0.8439 & 0.8271 & 0.9500 & 0.8843 & 0.3583 & \textbf{1.0000} & 0.5275  \\
        - STE400 & 0.7742 & 0.9411 & 0.8495 & 0.8229 & 0.9608 & 0.8865 & 0.3567 & \textbf{1.0000} & 0.5258  \\
        - STE1.5 & 0.7750 & 0.9353 & 0.8477 & 0.8227 & 0.9595 & 0.8858 & 0.3567 & \textbf{1.0000} & 0.5258  \\
        \bottomrule
    \end{tabular}
    }
\end{table}

\clearpage
\section{Multimodal Results}
\label{sec:app:multimodal}

\subsection{Vision}
% 1. Short intro about hallucinations
% 2. Describe experiment in detail
% 3. Share images with corresponding nice plot
% 4. Evt share correllation plots
% 5. Conclude with embedding models are not very strong yet "**distance between the distributions of low‑level visual features** (e.g. pixel‑value histograms, local texture/edge descriptors, or early‑layer CNN activations). It does _not_ measure “number of objects” or any high‑level semantic concept like “how many apples."

Hallucinations in vision models occur when generated images contain elements that are implausible, inconsistent, or irrelevant to the input prompt, such as incorrect object counts, distorted anatomy, or extraneous artifacts. These issues highlight a gap between low-level visual fidelity and high-level semantic correctness, especially in models that lack explicit grounding. To investigate \nameS applicability in the vision domain, we induce controlled hallucinations by prompting models to generate images with increasing object counts, a setup that systematically challenges semantic precision.

\begin{figure}[b]
    \centering
    \includegraphics[width=0.95\linewidth]{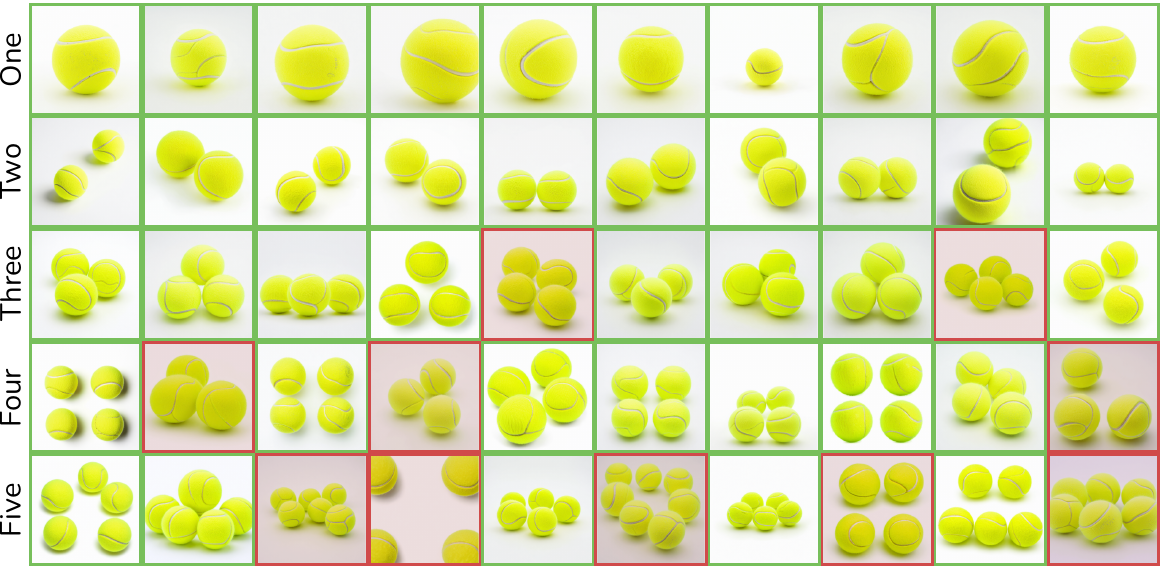}
    \caption{\textbf{Base images from one experimental run to detect hallucinations in generative image outputs.} Further details explained in Section~\ref{sec:app:multimodal}. All images were generated using the Stable Diffusion 3.5 Medium model~\cite{pmlr-v235-esser24a} with the prompt "\textit{\{NUM\}  yellow tennis ball\{s\} on a white background}", where \textit{NUM} is a number from one to five (written in words), and the plural ‘s’ in “balls” is used in all prompts except when requesting a single tennis ball. As the number of requested items increased, the accuracy of generating the exact count decreased, leading to more hallucinations.}
    \label{fig:tennis_balls_imgs}
\end{figure}

\begin{figure}[t]
    \centering
    \includegraphics[width=0.95\linewidth]{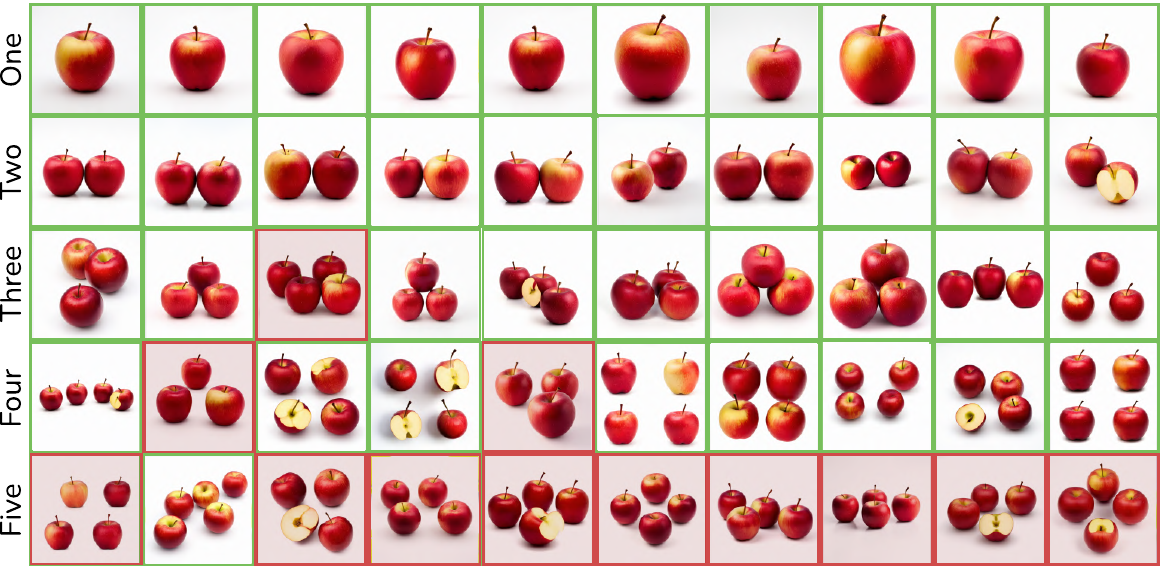}
    \caption{\textbf{Base images from one experimental run to detect hallucinations in generative image outputs.} Further details explained in Section~\ref{sec:app:multimodal}. All images were generated using the Stable Diffusion 3.5 Medium model~\cite{pmlr-v235-esser24a} with the prompt "\textit{\{NUM\}  apple\{s\} on a white background}", where \textit{NUM} is a number from one to five (written in words), and the plural ‘s’ in “apples” is used in all prompts except when requesting a single apple. As the number of requested items increased, the accuracy of generating the exact count decreased, leading to more hallucinations.}
    \label{fig:apples_imgs}
\end{figure}

\begin{figure}[tbp]
    \centering
    \includegraphics[width=0.75\linewidth]{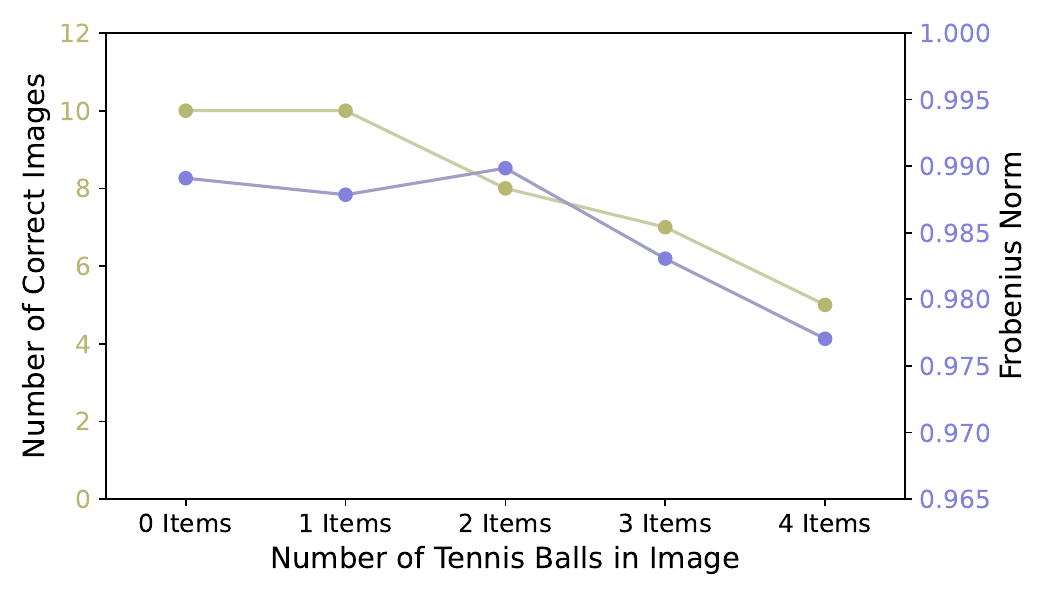}
    \caption{\textbf{\nameS results from the sample experiment using the images shown in Figure~\ref{fig:tennis_balls_imgs}.} \nameS tracks the trend in the number of correct images, effectively detecting hallucinations in the generated outputs. % Further details are explained in Section~\ref{sec:app:multimodal} with the average of all experiments shown in Figure~\ref{fig:hallucinate_images}.}
    We show the average of all experiments in Figure~\ref{fig:hallucinate_images}.}
    \label{fig:tennis_balls_frobnorm}
\end{figure}

In each experiment, we prompt the model to generate an image containing a specific object repeated $N$ times, where $N$ ranges from one to five. To ensure reliable evaluation with \nameS, each prompt is sampled 10 times, resulting in a set of generated images per object-count condition. These images are then embedded, and the resulting vectors are analyzed using \nameS to estimate the likelihood of hallucinations. The images of one such experiment are depicted in Figure~\ref{fig:tennis_balls_imgs}. We repeat this process across eight different objects to reduce variance in the results. The list of objects was created randomly without any prior screening or evaluation of the generated outcomes. Due to the need for manual inspection of each image, the total number of experiments is limited. All images are generated using the Stable Diffusion 3.5 Medium model~\cite{pmlr-v235-esser24a}, with embeddings computed using CLIP's Vision Transformer Large~\cite{icml/RadfordKHRGASAM21}. Both models are used with default hyperparameters.

To illustrate the experimental process in greater detail, we analyze a single run of the hallucination detection experiment presented in Section~\ref{sec:beyond_LLMs}. Figure~\ref{fig:tennis_balls_imgs} displays the generated images, with hallucinated outputs highlighted in red. Following this, Figure~\ref{fig:tennis_balls_frobnorm} plots the corresponding \nameS scores against the number of correctly rendered images, revealing a consistent trend: as the frequency of hallucinations increases, the \nameS score decreases. We provide an additional example in Figure~\ref{fig:apples_imgs}.

Analyzing our experiments also revealed notable limitations in vision embedding models such as CLIP. In contrast to their textual counterparts, the vision embedding models appear less semantically expressive and are largely focused on low-level visual features, such as pixel statistics, textures, or early-layer activations, rather than high-level semantic concepts like object count or identity. Consequently, while \nameS can flag deviations in visual patterns, it does not directly measure whether an image semantically matches its prompt, pointing to the need for more semantically grounded embeddings in vision tasks, a known problem.

%%%%%%%%%%%%%%%%%%%%%%%%%%%%%%%%%%%%%%%%%%%%%%%%%%%%%%%%%%%%

\end{document}